\documentclass[12pt]{article}
\usepackage{a4wide}

\usepackage{newproof}
\usepackage{latexsym}
\usepackage{amsfonts, amsmath, amssymb}
\usepackage{graphics}

\usepackage[hidelinks]{hyperref}

\usepackage{amsfonts,amsmath,amssymb,color,latexsym}
\newtheorem{theorem}{Theorem}
\newtheorem{lemma}[theorem]{Lemma}
\newtheorem{proposition}[theorem]{Proposition}

\newtheorem{example}[theorem]{Example}
\newtheorem{definition}[theorem]{Definition}

\def\N{\mathbb{N}}

\usepackage{url}

\newcommand{\eq}{\leftrightarrow}
\newcommand{\Eq}{\Leftrightarrow}
\newcommand{\imp}{\rightarrow}
\newcommand{\Imp}{\Rightarrow}

\newcommand{\Pmi}{\Leftarrow}
\newcommand{\et}{\wedge}
\newcommand{\vel}{\vee}
\newcommand{\Et}{\bigwedge}

\newcommand{\dia}[1]{\langle #1 \rangle}

\newcommand{\M}{\hat{B}}
\renewcommand{\phi}{\varphi}
\newcommand{\union}{\cup}

\newcommand{\weg}[1]{}

\newcommand{\Nat}{\mathbb N}
\newcommand{\Naturals}{\Nat}

\newcommand{\wordmodels}{\models^*}

\newcommand{\lang}{{\mathcal L}}
\newcommand{\Laa}{{\mathcal L}_{aa}}
\newcommand{\Lpal}{{\mathcal L}_{pal}}
\newcommand{\Lel}{{\mathcal L}_{ml}}
\newcommand{\Lml}{{\mathcal L}_{ml}}

\newtheorem{corollary}[theorem]{Corollary}

\newcommand{\proj}{{\upharpoonright}}
\newcommand{\tri}{\triangleright}

\usepackage{tikz}

\newcommand{\p}{p}
\newcommand{\q}{q}
\newcommand{\np}{\overline{p}}
\newcommand{\nq}{\overline{q}}

\newcommand{\powerset}{\mathcal P}

\newcommand{\Kw}{\mathit{Kw}}

\begin{document}
\title{Asynchronous Announcements}
\author{P. Balbiani\thanks{IRIT, CNRS, University of Toulouse, France} \and H. van Ditmarsch\thanks{LORIA, CNRS, University of Lorraine, France} \and S. Fern\'andez Gonz\'alez\thanks{IRIT, CNRS, University of Toulouse, France}}

\maketitle

\begin{abstract}
We propose a multi-agent epistemic logic of asynchronous announcements, where truthful announcements are publicly sent but individually received by agents, and in the order in which they were sent. Additional to epistemic modalities the logic contains dynamic modalities for making announcements and for receiving them. What an agent believes is a function of her initial uncertainty and of the announcements she has received. Beliefs need not be truthful, because announcements already made may not yet have been received. As announcements are true when sent, certain message sequences can be ruled out, just like inconsistent cuts in distributed computing.

We provide a complete axiomatization for this \emph{asynchronous announcement logic} ($AA$). It is a reduction system that also demonstrates that any formula in $AA$ is equivalent to one without dynamic modalities, just as for public announcement logic. 
A detailed example modelling message exchanging processes in distributed computing in $AA$ closes our investigation.
\end{abstract}
\section{Introduction} \label{sec.introduction}

What does an agent know and how does its knowledge change in a distributed system consisting of multiple agents that act independently from one another and wherein each agent may keep its own time? Agents' knowledge may change while they send and receive messages to each other and they may also receive messages from the environment, under conditions of temporal uncertainty. Such notions of \emph{asynchronous knowledge} and of asynchronous common knowledge have been investigated in depth in distributed computing \cite{Ben-ZviM14,GenestPS15,kshemkalyanietal:2008,lamport:1978,MukundS97} and in temporal epistemic logics to describe their behaviour \cite{chandymisra:85,halpernmoses:1990,panangadenetal:1992,Ramanujam96b}. Dynamic Epistemic Logic ($DEL$) \cite{baltagetal:1998,hvdetal.del:2007,baltagetal.hpi:2008, jfak.book:2011,hvdetal.handbook:2015} (or online references such as \url{https://en.wikipedia.org/wiki/Dynamic_epistemic_logic} and \url{https://plato.stanford.edu/entries/dynamic-epistemic/}) is a modal logic of knowledge and change of knowledge that models observation, i.e., receiving messages, and that has no notion of time, i.e., no temporal modalities. $DEL$ was thought to enforce synchrony \cite{jfaketal.JPL:2009,BenthemD08}. However, more recent studies revealed different ways for $DEL$ to accommodate asynchrony.

In the first place, agents may be uncertain about the number of actions that have already taken place: this is asynchrony due to partial observation, causing indistinguishable histories (sequences of messages) of different length. A framework for such asynchrony was convincingly presented in \cite{degremontetal:2011}, wherein they demonstrated that the supposed synchronicity of $DEL$ was a mere artifact of the \cite{jfaketal.JPL:2009} embedding of $DEL$ into synchronous temporal epistemic logics, namely caused by a non-standard interpretation of the Pnueli perfect recall axiom. Asynchrony of that kind is implicit in many $DEL$ scenarios. For example, in gossip protocols agents communicate by calling each other, so that $a$ may have called $b$ without another agent $c$ noticing that the call took place. Such fully distributed gossip is modelled in \cite{apt-tark}. For another example, in the `One hundred prisoners and a light bulb' riddle agents communicate asynchronously by individually toggling a light bulb out of sight and hearing of other agents \cite{hvdetal.KR:2010}. As a final example, the immediate snapshot algorithm, wherein agents are unaware of other agents possibly simultaneously accessing a shared memory location, has been modelled in $DEL$ by \cite{goubaultetal:2018}. 

A different kind of asynchrony results when the sending and receiving of messages are separate, so that the receiver is uncertain about the moment a received message was sent.
To our knowledge, the asynchronous reception of messages broadcast by the environment has only been modelled in $DEL$ by \cite{KnightMS15,KnightMS19,schwarzentruber:2019} --- from here on we refer to the reference journal version \cite{KnightMS19} only.
Our proposal further develops a preliminary version, see \url{https://arxiv.org/abs/1705.03392v2}, presented at the workshop \emph{Strategic Reasoning} (SR 2017) in Liverpool. It builds on the {\em protocol-generated forest} of \cite{jfaketal.JPL:2009,hvdetal.FAMAS:2013} and the {\em history-based structures} of \cite{parikhetal:2003}, as well as the \emph{asynchronous knowledge} and {\em concurrent common knowledge} of \cite{chandymisra:85,halpernmoses:1990,mosesetal:1986,panangadenetal:1992}.
Like \cite{KnightMS19} we assume that announcements are still broadcast to all agents, but individually received. Unlike \cite{KnightMS19} our epistemic notion is interpreted over past messages only, and we provide an axiomatization by way of a reduction to the modal fragment, just as for public announcement logic \cite{plaza:1989}. Before we delve further into technical particulars, let us first continue with a detailed example illustrating our approach.

Consider two agents Anne ($a$) and Bill ($b$), and two propositional variables ${p}$ and ${q}$. Anne knows the truth about ${p}$ and Bill knows the truth about ${q}$, and this is common knowledge between them. We can encode this uncertainty in a Kripke model (Figure \ref{fig}(i)). In {\em public announcement logic} ($PAL$) \cite{plaza:1989} we can formalize that after the announcement of $p \vel q$, Bill does not know that ${p}$ is true but Anne considers it possible that he knows, namely as $[p\vel{q}](\neg B_b {p} \et \M_a B_b {p})$. (The formula $\neg B_b {p} \et \M_a B_b {p}$ would be true if $p$ and $q$ were initially both true.) Operator $[{p}\vel{q}]$ is a dynamic modality interpreted by model restriction. The (knowledge or belief) modalities bound by it are interpreted in the restriction (Figure \ref{fig}({\sc{pal}})), not in the original model.

\begin{figure}
\scalebox{0.85}{
\begin{tikzpicture}
\node (at) at (-1,1) {\large $\stackrel {p \vel q} \Pmi$};
\node (abt) at (-1,.55) {\footnotesize (old)};
\node (a01tt) at (-4,2) {$\np\q$};
\node (a10tt) at (-2.5,.7) {$\p\nq$};
\node (a11tt) at (-2.5,2.7) {$\p\q$};
\draw (a01tt) -- node[fill=white,inner sep=1pt] {$b$} (a11tt);
\draw (a10tt) -- node[fill=white,inner sep=1pt] {$a$} (a11tt);
\node (00) at (0,0) {$\np\nq$};
\node (01) at (0,2) {$\np\q$};
\node (10) at (1.5,.7) {$\p\nq$};
\node (11) at (1.5,2.7) {$\p\q$};
\draw (00) -- node[fill=white,inner sep=1pt] {$b$} (10);
\draw (01) -- node[fill=white,inner sep=1pt] {$b$} (11);
\draw (00) -- node[fill=white,inner sep=1pt] {$a$} (01);
\draw (10) -- node[fill=white,inner sep=1pt] {$a$} (11);
\node (t) at (3,1) {\large $\stackrel {p \vel q} \Imp$};
\node (bt) at (3,.55) {\footnotesize (novel)};
\node (00t) at (4,0) {\color{gray} $\np\nq$};
\node (01t) at (4,2) {$\np\q$};
\node (10t) at (5.5,.7) {$\p\nq$};
\node (11t) at (5.5,2.7) {$\p\q$};
\draw (00t) -- node[fill=white,inner sep=1pt] {$b$} (10t);
\draw (01t) -- node[fill=white,inner sep=1pt] {$b$} (11t);
\draw (00t) -- node[fill=white,inner sep=1pt] {$a$} (01t);
\draw (10t) -- node[fill=white,inner sep=1pt] {$a$} (11t);
\node (t) at (7,1) {\large $\stackrel {a} \Imp$};
\node (00tt) at (8,0) {\color{gray} $\np\nq$};
\node (01tt) at (8,2) {$\np\q$};
\node (10tt) at (9.5,.7) {$\p\nq$};
\node (11tt) at (9.5,2.7) {$\p\q$};
\draw (00tt) -- node[fill=white,inner sep=1pt] {$b$} (10tt);
\draw (01tt) -- node[fill=white,inner sep=1pt] {$b$} (11tt);
\draw[color=gray] (00tt) -- node[color=gray,fill=white,inner sep=1pt] {\color{gray} $a$} (01tt);
\draw (10tt) -- node[fill=white,inner sep=1pt] {$a$} (11tt);
\node (tt) at (11,1) {\large $\stackrel {b} \Imp$};
\node (00uu) at (12,0) {\color{gray} $\np\nq$};
\node (01uu) at (12,2) {$\np\q$};
\node (10uu) at (13.5,.7) {$\p\nq$};
\node (11uu) at (13.5,2.7) {$\p\q$};
\draw[color=gray] (00uu) -- node[color=gray,fill=white,inner sep=1pt] {\color{gray} $b$} (10uu);
\draw (01uu) -- node[fill=white,inner sep=1pt] {$b$} (11uu);
\draw[color=gray] (00uu) -- node[color=gray,fill=white,inner sep=1pt] {\color{gray} $a$} (01uu);
\draw (10uu) -- node[fill=white,inner sep=1pt] {$a$} (11uu);
\node (pal) at (-3.35,-.5) {$(\text{\sc{pal}})$};
\node (i) at (.75,-.5) {$(i)$};
\node (ii) at (4.75,-.5) {$(ii)$};
\node (iii) at (8.75,-.5) {$(iii)$};
\node (iv) at (12.75,-.5) {$(iv)$};
\end{tikzpicture}
}
\caption{Left, public announcement, and right, asynchronous announcement. Right, the announcement $p\vel q$ is sent, after which first Anne and then Bill receives it. What Anne and Bill know is a function of the initial \emph{model} encoding their knowledge and ignorance and the actual \emph{state} in this model $(i)$, and this \emph{history} $(p\vel{q})ab$ of three events. States are labelled with the valuations of $p$ and $q$, where $\np$ stands for $\neg p$ and $\nq$ stands for $\neg q$. States that are indistinguishable for an agent are linked with a label for that agent. The greying of states and links is merely for expository purposes. Cf.\ to Figure~\ref{fig.history}, later.}
\label{fig}
\end{figure}

Let us now assume that announcements are still publicly sent, but individually received. Then, after the announcement $p \vel q$ is made (Figure \ref{fig}(ii)), Anne may have received that information ${p}\vel{q}$ but Bill not yet (Figure \ref{fig}(iii)), after which Bill receives it too (Figure \ref{fig}(iv)). Unlike in Figure \ref{fig}({\sc{pal}}), in (iv) they do not know that the other knows; there is no common knowledge between them of $p \vel q$.

Separating sending from receiving messages permits a notion of asynchronous knowledge in $DEL$, that is a function of the usual modal accessibility but also of uncertainty over the announcements received by other agents. In (iii), after receiving announcement $p \vel q$, Anne considers it possible that Bill knows $p$, namely if the state is $p\nq$ and if Bill has also received the announcement, as in (iv). For different reasons she also considers it possible that Bill does not know $p$. Firstly, if the state is $\p\q$ (or $\np\q$) and the announcement has been received by Bill (iv)). But, secondly, also if the actual state is $\p\nq$ and the announcement has not yet been received by Bill (iii), in which case Bill still considers it possible that the state is $\np\nq$. And what Anne knows in (iii) should be the same as what she knows in (iv).

What does Bill know? According to usage in distributed computing \cite{halpernmoses:1990}, even when Bill has not received the announcement $p\vel q$, he can imagine that such a message has been sent and that Anne has received it. Therefore, although he is uncertain about $p$ (i), he should consider it possible that announcement $p \vel q$ was made (ii), and that Anne has received it (iii), that she therefore considers it possible that he has received it too (iv), and that he therefore now knows that $p$. This notion of knowledge does not seem to fit well a setting wherein the messages are announcements, whose role is to reduce uncertainty of the value of unchanging facts. It well fits the setting of distributed computing wherein messages that are broadcast, i.e., announcements, contain novel facts. In $PAL$ the future is predictable: all facts may become known. In our setting, this only allows for weak forms of higher-order knowledge: an agent cannot know that another agent remains ignorant.

We therefore focus on what agents know based on the announcements they have received so far, ignoring possible future announcements. That means that in situations (i), (ii), (iii) above, Bill `knows' that Anne knows that he is uncertain about $p$, as he has not received the announcement $p \vel q$. In (i) and (ii) this is true, but in (iii) this is no longer true. Bill's knowledge is then incorrect belief. Indeed, the asychronous epistemic notion that we propose is one of asynchronous belief (however, as we will see, of the special kind that many such beliefs will eventually become knowledge). Other defining assumptions of our asynchronous semantics are that agents receive the announcements in the order in which they are made, as is not uncommon in distributed computing \cite{halpernmoses:1990,panangadenetal:1992}; and that announcements are true when sent, as in $PAL$.

The assumption that announcements are true when sent, results in {\em partial synchronization}. Let us suppose that Anne and Bill are both uncertain about $p$. Then, announcement $p$ is followed by announcement $\neg B_b p$. If Anne received $p$ and $\neg B_b p$, she should not consider it possible that Bill has received $p$ before the second announcement was made. In other words, the histories of sending and receiving events that she considers possible include $pa(\neg B_bp)ba$ and $p(\neg B_bp)aab$, but exclude $pb(\neg B_bp)aa$ and $pab(\neg B_bp)ba$ where the first $a$ in the sequence stands for Anne receiving the first announcement $p$, the second $a$ stands for her receiving the second announcement $\neg B_b p$, and similarly for $b$.
In terms of \cite{panangadenetal:1992}, $pb(\neg B_bp)aa$ and $pab(\neg B_bp)ba$ would be called {\em inconsistent cuts}. In order for second announcement $\neg B_b p$ to be truthful, Bill must still be uncertain about $p$, and for Bill to remain uncertain about $p$ he must not yet have received the first announcement $p$. We will introduce a so-called `agreement' relation between states and histories, and we then say that a state $s$ in the model does not \emph{agree} with history $pb(\neg B_bp)aa$. Agents only consider histories possible that agree with the states they consider possible. This requires to define a satisfaction relation and such an agreement relation by simultaneous induction.

Intuitively, summing up, in our approach an agent knows/believes $\phi$ iff $\phi$ is true: (1) in all states that it considers possible, (2) for all prefixes of announcement sequences that other agents may have received, (3) taking into account that the announcements it received were true when sent, (4) while ignoring that other agents may have received more announcements than itself.

\medskip

We now present an outline of our contribution. Section \ref{sec.logic} defines the syntax and Section \ref{sec.semantics} defines the semantics of \emph{asynchronous announcement logic} $AA$. In particular, Section \ref{sec.ml} relates $AA$ to basic modal logic. Section \ref{sec.ax} discusses the axiomatization: Section \ref{sec.elim} provides an axiomatization of $AA$ on the class of models with empty histories, and Section \ref{sec.star} provides rewrite rules on the class of models with arbitrary histories. Section \ref{sec.comparison} obtains results for the model class $\mathcal S5$, elaborates on the difference between knowledge and belief, and compares our proposal with various dynamic and temporal epistemic logics and with distributed computing.
Section \ref{sec.dc} models a typical scenario of message sending agents in distributed computing.

\section{Syntax} \label{sec.logic}

\subsection{Language of asynchronous announcement logic}

\begin{definition}[Language of AA]
Let $P$ be a countable set of atoms (denoted $p$, $q$, etc.) and $A$ be a finite set of agents (denoted $a$, $b$, etc.).
The language {$\Laa$} of asynchronous announcement logic is defined as follows:
\[ \varphi,\psi:=p\mid\bot\mid\neg\varphi\mid(\varphi\vee\psi)\mid B_a\varphi \mid\lbrack\varphi\rbrack\psi\mid\lbrack a\rbrack\varphi\]
\end{definition}
We will follow the standard rules for omission of the parentheses.
Without the constructs $[a]\cdot$ we get the language $\Lpal$ of public announcement logic and without the construct $[\phi]\cdot$ as well we get the language $\Lel$ of multi-agent modal logic. The \emph{positive} fragment $\Lel^+$ of $\Lel$ is defined as follows:
\[ \varphi,\psi:=p\mid\neg p \mid\bot\mid\top\mid (\varphi\vee\psi)\mid (\varphi\et\psi)\mid B_a\varphi \]
We will use the standard abbreviations for the Boolean constructs.
We will also use the following constructs: $\langle a\rangle\varphi:=\neg\lbrack a\rbrack\neg\varphi$, $\langle\varphi\rangle\psi:=\neg\lbrack\varphi\rbrack\neg\psi$ and $\M_a\varphi:=\neg B_a\neg\varphi$.
For $B_a \phi$ read ``agent $a$ believes/knows $\phi$,'' for $[\phi]\psi$ read ``after public announcement $\phi$ has been sent/made, $\psi$,'' and for $[a]\phi$ read ``after agent $a$ receives/reads the next announcement, $\phi$.''
For all formulas $\eta,\psi$ and for all atoms $p$, we denote by $\eta(p/\psi)$ the uniform substitution of the occurrences of $p$ in $\eta$ by $\psi$.

Consider $A\cup \Laa$ as an alphabet, with agents and formulas as letters.
Variables for words in this language are $\alpha,\beta,\dots$.
The empty word is denoted $\epsilon$.
Given a word $\alpha$ over $A\cup\Laa$, ${\mid}{\alpha}{\mid}$ is its length, ${\mid}{\alpha}{\mid}_{a}$ is the number of its $a$'s for each $a\in A$, ${\mid}{\alpha}{\mid}_!$ is the number of its formula occurrences, $\alpha\proj_!$ is the projection of $\alpha$ to $\Laa$, and $\alpha\proj_{!a}$ is the restriction of $\alpha\proj_!$ to the first $|\alpha|_a$ formulas. (We use the symbol `!' here because in $PAL$ expression $!\phi$ often represents ``the announcement is $\phi$''.) 
These notions have obvious inductive definitions.
We say that a word $\beta$ is a \emph{prefix} of a word $\alpha$ (in symbols $\alpha\subseteq\beta$) if $\beta$ is an initial sequence of $\alpha$.
Obviously, for all words $\alpha,\beta$, $\alpha\subseteq\alpha$ and if $\beta\subseteq\alpha$, then for all $a \in A$ and $\psi\in\Laa$, $\beta\subseteq\alpha{a}$ and $\beta\subseteq\alpha\psi$.
Given a word $\alpha$ and $n \in \Naturals$, $\alpha^n$ denotes a concatenation of $n$ copies of $\alpha$.

Let $\alpha$ be a word over $A\cup\Laa$.
In the single-agent case, when $A=\{a\}$, it is clear that ${\mid}{\alpha}{\mid}={\mid}{\alpha}{\mid}_{a}+{\mid}{\alpha}{\mid}_!$.
Otherwise, in the multi-agent case, when $|A|\geq 2$, considering an enumeration $(a_{1},\ldots,a_{n})$ of $A$ without repetition, ${\mid}{\alpha}{\mid}={\mid}{\alpha}{\mid}_{a_{1}}+\ldots+{\mid}{\alpha}{\mid}_{a_{n}}+{\mid}{\alpha}{\mid}_!$.
\begin{definition}[History]
A word $\alpha$ in the language $A\cup \Laa$ is a {\em history} if for all prefixes  $\beta\subseteq\alpha$ and for all $a \in A$, ${\mid}\beta{\mid}_!\geq{\mid}\beta{\mid}_{a}$.
\end{definition}
Obviously, if $\beta$ is a prefix of a history $\alpha$, then $\beta$ is a history too.
In the definition of histories, the requirement ``for all $a \in A$, ${\mid}\beta{\mid}_!\geq{\mid}\beta{\mid}_{a}$''
means that, for all $n \in \Naturals$, if there is an $n$-th occurrence of agent $a$ in $\alpha$, then there is a prior $n$-th formula in $\alpha$ that will be the announcement then received by agent $a$.
This match will be used in the semantics.
\begin{definition}[View relation]\label{definition:view:relation}
Let $\alpha,\beta$ be histories and $a\in A$.
We define: $\alpha\triangleright_a\beta$ iff  $\beta\proj_{!}=\beta\proj_{!a}=\alpha\proj_{!a}$. \
The set $\mathsf{view}_a(\alpha) := \{ \beta \mid \alpha \triangleright_a \beta \}$ is the {\em view} of $a$ given $\alpha$. 
\end{definition}
Observe that, for all $a\in A$, if $\alpha\triangleright_{a}\beta$ then ${\beta}{\upharpoonright}_!$ is a prefix of ${\alpha}{\upharpoonright}_!$. Informally, the view of agent $a$ given history $\alpha$ consists of all the different ways in which $a$ can receive the announcements in $\alpha$.
In other words, the view of $a$ given $\alpha$ consists of the histories $a$ considers possible but without taking the meaning of the announcements in the history into account, which, as we will see, results in a further restriction.
In Section \ref{sec.comparison} we will present an alternative for the view relation, without the requirement that ${\mid}{\beta}{\mid}_!={\mid}{\alpha}{\mid}_{a}$.
%
%
\begin{example} \label{exa.history}
Let us have two agents, $A=\{a,b\}$, and let the history be $\alpha = (p\vel\q)a$.
Then $\mathsf{view}_a(\alpha)$, the set of all histories $\beta$ such that $\alpha \tri_a \beta$, is $\{ (p\vel\q)ab, (p\vel\q)ba, (p\vel\q)a \}$, whereas $\mathsf{view}_b(\alpha) = \{ \epsilon \}$.
Let now $\alpha' = (p\vel\q)ab$.
Then $\mathsf{view}_a(\alpha')= \{ (p\vel\q)ab, (p\vel\q)ba, (p\vel\q)a \}$ and $\mathsf{view}_{b}(\alpha')= \{ (p\vel\q)ab, (p\vel\q)ba, (p\vel\q)b \}$.
\end{example}
The following alternative characterizations of the $\triangleright_a$ relation will be useful. The proof is left to the reader.
\begin{lemma}\label{lemma:simple:properties:about:histories}
Let $\alpha,\beta$ be histories and $a\in A$. The following conditions are equivalent:
\begin{enumerate}
\item ${\mid}{\beta}{\mid}_{a}={\mid}{\alpha}{\mid}_{a}$, ${\beta}\proj_{!a} = {\alpha}\proj_{!a}$ and ${\mid}{\beta}{\mid}_!={\mid}{\alpha}{\mid}_{a}$;
\item ${\mid}\beta{\mid}_{a}={\mid}\alpha{\mid}_{a}$ and $\beta\proj_{!}=\alpha\proj_{!a}$;
\item $\beta\proj_{!}=\beta\proj_{!a}=\alpha\proj_{!a}$.
\end{enumerate}
\end{lemma}
%
%
%
%
We can introduce modalities for histories by abbreviation, using reception and announcement modalities of the form $[a]$ and $[\psi]$.
For all words $\alpha$ over $A\cup\Laa$, the modality $\lbrack\alpha\rbrack$ is inductively defined as: $\lbrack\epsilon\rbrack\varphi:=\varphi$, $\lbrack\alpha a\rbrack\varphi:=\lbrack\alpha\rbrack\lbrack a\rbrack\varphi$, and $\lbrack\alpha\psi\rbrack\varphi:=\lbrack\alpha\rbrack\lbrack\psi\rbrack\varphi$; whereas its dual $\langle\alpha\rangle\phi$ is defined by abbreviation as $\neg\lbrack\alpha\rbrack\neg\varphi$. We will read $[\alpha]\phi$ as ``if the sequence $\alpha$ of events can be executed then $\phi$ holds after its execution,'' whereas we will read $\langle\alpha\rangle\phi$ as ``the sequence $\alpha$ of events can be executed and $\phi$ holds after its execution''. Clearly, for all words $\alpha$ over $A\cup\Laa$, for all $a\in A$ and for all $\varphi,\psi\in\Laa$, $\lbrack a\alpha\rbrack\varphi$ is an abbreviation of $\lbrack a\rbrack\lbrack\alpha\rbrack\varphi$, $\langle a\alpha\rangle\varphi$ of $\langle a\rangle\langle\alpha\rangle\varphi$, $\lbrack\psi\alpha\rbrack\varphi$ to $\lbrack\psi\rbrack\lbrack\alpha\rbrack\varphi$, and $\langle\psi\alpha\rangle\varphi$ of $\langle\psi\rangle\langle\alpha\rangle\varphi$.
%
%

\weg{

the modalities $\lbrack\alpha\rbrack$ and $\langle\alpha\rangle$ are inductively defined as follows:
\begin{itemize}
\item $\lbrack\epsilon\rbrack\varphi:=\varphi$,
\item $\langle\epsilon\rangle\varphi:=\varphi$,
\item $\lbrack\alpha a\rbrack\varphi:=\lbrack\alpha\rbrack\lbrack a\rbrack\varphi$,
\item $\langle\alpha a\rangle\varphi:=\langle\alpha\rangle\langle a\rangle\varphi$,
\item $\lbrack\alpha\psi\rbrack\varphi:=\lbrack\alpha\rbrack\lbrack\psi\rbrack\varphi$,
\item $\langle\alpha\psi\rangle\varphi:=\langle\alpha\rangle\langle\psi\rangle\varphi$.
\end{itemize}
We will read $[\alpha]\phi$ as ``if the sequence $\alpha$ of events can be executed then $\phi$ holds after its execution'' whereas we will read $\langle\alpha\rangle\phi$ as ``the sequence $\alpha$ of events can be executed and $\phi$ holds after its execution''.
\begin{lemma}\label{useful:lemma:about:boxes:and:diamonds}
For all words $\alpha$ over $A\cup\Laa$, for all $a\in A$ and for all $\varphi,\psi\in\Laa$, $\lbrack a\alpha\rbrack\varphi=\lbrack a\rbrack\lbrack\alpha\rbrack\varphi$, $\langle a\alpha\rangle\varphi=\langle a\rangle\langle\alpha\rangle\varphi$, $\lbrack\psi\alpha\rbrack\varphi=\lbrack\psi\rbrack\lbrack\alpha\rbrack\varphi$ and $\langle\psi\alpha\rangle\varphi=\langle\psi\rangle\langle\alpha\rangle\varphi$.
\end{lemma}
\begin{proof}
The proof is by $<$-induction on ${\mid}\alpha{\mid}$.
\end{proof}
Lemma \ref{useful:lemma:about:boxes:and:diamonds} will be implicitly used several times later in the paper.
}

\subsection{Results for histories} \label{sec.resultsforhistories}
We continue with some basic results for histories that will be used later.
\weg{
\begin{lemma}\label{lemma:histories:triangle:right:prefix}
%
%
For all $a\in A$, if $\alpha\triangleright_{a}\beta$ then ${\beta}{\upharpoonright}_!$ is a prefix of ${\alpha}{\upharpoonright}_!$.
\end{lemma}
\begin{proof}
Let $a\in A$ be such that $\alpha\triangleright_{a}\beta$.
Hence, ${\mid}\beta{\mid}_{a}={\mid}\alpha{\mid}_{a}$, ${\beta}{\upharpoonright}_{!a}={\alpha}{\upharpoonright}_{!a}$ and ${\mid}\beta{\mid}_{!}={\mid}\alpha{\mid}_{a}$.
Thus, $\beta$ contains the same number of $a$'s as $\alpha$, the restriction of the projection of $\beta$ to $\Laa$ to the first ${\mid}\beta{\mid}_{a}$ formulas is equal to the restriction of the projection of $\alpha$ to $\Laa$ to the first ${\mid}\alpha{\mid}_{a}$ formulas and the number of formula occurrences in $\beta$ is equal to the number of $a$'s in $\alpha$.
Consequently, ${\beta}{\upharpoonright}_!$ is a prefix of ${\alpha}{\upharpoonright}_!$.
\end{proof}
}
\begin{lemma}\label{lemma:5:bis}
Let $\alpha,\beta$ be histories.
For all $a,b\in A$, if $\alpha\triangleright_a\beta$ then ${\mid}\beta{\mid}_{b}\leq{\mid}\alpha{\mid}_{a}$.
\end{lemma}
\begin{proof}
Let $a,b\in A$ be such that ${\alpha}{\triangleright_a}{\beta}$.
Hence, ${\mid}\beta{\mid}_{!}={\mid}\alpha{\mid}_{a}$.
Since $\beta$ is a history, ${\mid}\beta{\mid}_!\geq{\mid}\beta{\mid}_{b}$.
Since ${\mid}\beta{\mid}_{!}={\mid}\alpha{\mid}_{a}$,  ${\mid}\beta{\mid}_{b}\leq{\mid}\alpha{\mid}_{a}$.
\end{proof}
\begin{lemma} \label{lemma.epsilon}
Let $\alpha,\beta$ be histories, and $a \in A$. \begin{enumerate} \item If $\epsilon\triangleright_{a}\alpha$ then $\alpha=\epsilon$,  \item if $\alpha\triangleright_{a}\beta$ then $\beta\triangleright_{a}\beta$. \end{enumerate}
\end{lemma}
\begin{proof} \  \vspace{-.2cm}
\begin{enumerate}
\item Suppose $\epsilon\triangleright_{a}\alpha$.
Hence, ${\mid}\alpha{\mid}_{!}={\mid}\epsilon{\mid}_{a} = 0$.
Since $\alpha$ is a history,  for all $b\in A$, ${\mid}\alpha{\mid}_{!}\geq{\mid}\alpha{\mid}_{b}$, and thus $|\alpha|_b = 0$.
Consequently,  $\alpha=\epsilon$.
\item Suppose $\alpha\triangleright_{a}\beta$.
Hence, ${\mid}\beta{\mid}_{a}={\mid}\alpha{\mid}_{a}$ and ${\mid}\beta{\mid}_{!}={\mid}\alpha{\mid}_{a}$.
Thus, ${\mid}\beta{\mid}_{!}={\mid}\beta{\mid}_{a}$.
Consequently, $\beta\triangleright_{a}\beta$.
\end{enumerate}
 \vspace{-.6cm}
\end{proof}
Given history $\alpha$ and agent $a$, we recursively define a word $\alpha_{a}$ as follows:
\begin{itemize}
\item $\epsilon_a=\epsilon$;
\item $(\alpha\phi)_a=\alpha_a$;
\item $(\alpha b)_a=\alpha_a$ for each $b\in A\setminus\{a\}$;
\item for all $n>0$, $(\alpha b a^n)_a=(\alpha a^n)_a$ for each $b\in A\setminus\{a\}$;
\item for all $n>0$, if $|\alpha\phi a^n|_!=|\alpha\phi a^n|_a$ then $(\alpha \phi a^n)_a=\alpha\phi a^n$ else $(\alpha \phi a^n)_a=(\alpha a^n)_a$.
\end{itemize}
Informally, $\alpha_a$ is the concatenation of the prefix $\gamma$ of $\alpha$ until the $|\alpha|_a$-th occurrence of a formula in $\alpha$ with $|\alpha|_a - |\gamma|_a$ times the letter $a$.
\begin{lemma}\label{lemma:about:alpha:a:definition:serial:tri}
For all histories $\alpha$ and for all agents $a$, $\alpha_{a}$ is a history such that $\alpha\triangleright_a\alpha_a$
\end{lemma}
\begin{proof}
The proof is by $<$-induction on ${\mid}\alpha{\mid}$.
\end{proof}
\begin{proposition}[The view relation is serial, transitive, and Euclidean] \label{lemma.tri}
Let $\alpha$, $\beta$ and $\gamma$ be histories.
For all agents $a$,
\begin{enumerate}
\item \label{one} there is a history $\delta$ such that $\alpha \tri_a \delta$, namely $\alpha_{a}$,
\item if $\alpha\triangleright_{a}\beta$ and $\beta\triangleright_{a}\gamma$ then $\alpha\triangleright_{a}\gamma$,
\item if $\alpha\triangleright_{a}\beta$ and $\alpha\triangleright_{a}\gamma$ then $\beta\triangleright_{a}\gamma$.
\end{enumerate}
\end{proposition}
\begin{proof}
\  \vspace{-.2cm}
\begin{enumerate}
\item By Lemma \ref{lemma:about:alpha:a:definition:serial:tri}.
\item Suppose $\alpha\triangleright_{a}\beta$ and $\beta\triangleright_{a}\gamma$.
Hence, ${\mid}\beta{\mid}_{a}={\mid}\alpha{\mid}_{a}$, $\beta{\upharpoonright}_{!a}=\alpha{\upharpoonright}_{!a}$, ${\mid}\gamma{\mid}_{a}={\mid}\beta{\mid}_{a}$, $\gamma{\upharpoonright}_{!a}=\beta{\upharpoonright}_{!a}$ and ${\mid}\gamma{\mid}_{!}={\mid}\beta{\mid}_{a}$.
Thus, ${\mid}\gamma{\mid}_{a}={\mid}\alpha{\mid}_{a}$, $\gamma{\upharpoonright}_{!a}=\alpha{\upharpoonright}_{!a}$ and ${\mid}\gamma{\mid}_{!}={\mid}\alpha{\mid}_{a}$.
Consequently, $\alpha\triangleright_{a}\gamma$.
\item Suppose $\alpha\triangleright_{a}\beta$ and $\alpha\triangleright_{a}\gamma$.
Hence, ${\mid}\beta{\mid}_{a}={\mid}\alpha{\mid}_{a}$, $\beta{\upharpoonright}_{!a}=\alpha{\upharpoonright}_{!a}$, ${\mid}\gamma{\mid}_{a}={\mid}\alpha{\mid}_{a}$, $\gamma{\upharpoonright}_{!a}=\alpha{\upharpoonright}_{!a}$ and ${\mid}\gamma{\mid}_{!}={\mid}\alpha{\mid}_{a}$.
Thus, ${\mid}\gamma{\mid}_{a}={\mid}\beta{\mid}_{a}$, $\gamma{\upharpoonright}_{!a}=\beta{\upharpoonright}_{!a}$ and ${\mid}\gamma{\mid}_{!}={\mid}\beta{\mid}_{a}$.
Consequently, $\beta\triangleright_{a}\gamma$.
\end{enumerate}
 \vspace{-.2cm}
\end{proof}
From Proposition \ref{lemma.tri}, we obtain:
\begin{corollary} \label{corollary.proja}
For all histories $\alpha,\beta$ and for all agents $a$, if $\alpha \tri_a \beta$ then $\alpha_a \tri_a \beta$ and $\beta\tri_a\alpha_a$.
\end{corollary}
%

\begin{lemma}\label{lemma:for:all:words:if:then:A}
Let $\alpha,\beta$ be histories.
In the single-agent case, if ${\alpha}{\triangleright_a}{\beta}$ then ${\mid}{\beta}{\mid}=2{\mid}{\alpha}{\mid}_{a}$.
Otherwise, in the multi-agent case, if ${\alpha}{\triangleright_a}{\beta}$ then $2{\mid}{\alpha}{\mid}_{a}\leq{\mid}{\beta}{\mid}\leq(|A|+1){\mid}{\alpha}{\mid}_{a}$.
%
%
\end{lemma}
\begin{proof}
In the single-agent case, suppose $\alpha\triangleright_a\beta$.
Hence, ${\mid}{\beta}{\mid}_{a}={\mid}{\alpha}{\mid}_{a}$ and ${\mid}{\beta}{\mid}_!={\mid}{\alpha}{\mid}_{a}$.
Thus, ${\mid}{\beta}{\mid}=2{\mid}{\alpha}{\mid}_{a}$.
In the multi-agent case, suppose ${\alpha}{\triangleright_a}{\beta}$.
Hence, by Lemma~\ref{lemma:5:bis}, for all $b\in A\setminus\{a\}$, ${\mid}\beta{\mid}_{b}\leq{\mid}\alpha{\mid}_{a}$.
Moreover, ${\mid}\beta{\mid}_{a}={\mid}\alpha{\mid}_{a}$ and ${\mid}\beta{\mid}_{!}={\mid}\alpha{\mid}_{a}$.
Thus, $2\cdot{\mid}\alpha{\mid}_{a}\leq{\mid}\beta{\mid}\leq(|A|+1)\cdot{\mid}\alpha{\mid}_{a}$.
\end{proof}
Since $A$ is finite, by Lemma~\ref{lemma:for:all:words:if:then:A}, the relation ${\triangleright_a}$ is image-finite.
But we can do better.
Let $X$ and $Y$ be distinct symbols.
A {\em Dyck word} is a string consisting, for some $n\in\N$, of $n$ $X$'s and $n $ $Y$'s such that no prefix of the string has more $Y$'s than $X$'s. This matches exactly our histories of announcements ($X$) and read actions ($Y$).
The number of {\em Dyck words} of length $2n$ is $C_n$ where $C_n$ is the $n$-th {\em Catalan number}, defined as $C_n := \frac{1}{n+1}\binom{2n}{n}$. This generates the sequence $1,1,2,5,14,42,\dots$, see \url{https://oeis.org/A000108}.
\begin{proposition}
In the single-agent case, for all histories $\alpha$, $|\mathsf{view}_a(\alpha)| = C_{{\mid}\alpha{\mid}_{a}}$.
%
%
\end{proposition}
\begin{proof}
%
%
If there is a single agent, then histories can be transformed into Dyck words over the symbols $X$ and $Y$ when one replaces announcements by the symbol $X$ and read actions by the symbol $Y$.
Moreover, $\mathsf{view}_{a}(\alpha)$ is the set of all histories $\beta$ such that $\beta\proj_{!}=\alpha\proj_{!a}$ and $\beta\proj_{!a}=\alpha\proj_{!a}$.
Hence, ${\mid}\mathsf{view}_{a}(\alpha){\mid}=C_{{\mid}\alpha{\mid}_{a}}$.
%
\weg{
Let us now consider the case of $m$ agents receive $n$ announcements.
For each of the $m$ agents, there are $C_n$ ways to receive them. This makes $m C_n$ altogether. Interleaving for read modalities for different agents can be done in any way, without restriction, i.e., one of the $m$ agents can interleave in $\binom{mn}{n}$ ways with the other agents receiving messages. 
For $m$ agents this is $m\binom{mn}{n}$. The total is therefore $m C_n + m \binom{mn}{n}$, i.e., \[ m \Big( \frac{1}{n+1}\binom{2n}{n} + \binom{mn}{n} \Big) \]
}
%
%
%
\end{proof}
However, in the multi-agent case, an agent can receive $n$ announcements in many more than $C_n$ ways. Example \ref{exa.history} showed that if there are two agents, an agent can receive one announcement in three different ways instead of one way for one agent.

\section{Semantics} \label{sec.semantics}

\subsection{A well-founded order for the semantics} \label{sec.aux}
A well-founded order $\ll$ between (history, formula) pairs will be the basis of our semantics. It uses an auxiliary function ${\parallel}\cdot{\parallel}$ on formulas and on histories, and an auxiliary function $deg(\cdot)$ on (history, formula) pairs.

For all $\varphi\in\Laa$, let ${\parallel}{\varphi}{\parallel}$ be the positive integer inductively defined as follows:
\[\begin{array}{lllllllll}
{\parallel}{p}{\parallel} &:= & 2 & {\parallel}{\varphi\vee\psi}{\parallel}&:=&{\parallel}{\varphi}{\parallel}+{\parallel}{\psi}{\parallel} & {\parallel}{\lbrack\varphi\rbrack\psi}{\parallel}&:=&2{\parallel}{\varphi}{\parallel}+{\parallel}{\psi}{\parallel} \quad\quad \\
{\parallel}{\bot}{\parallel} &:=& 1 & {\parallel}{B_a\varphi}{\parallel} &:=&{\parallel}{\varphi}{\parallel}+1 \quad\quad\quad & {\parallel}{\lbrack a\rbrack\varphi}{\parallel} &:=&{\parallel}{\varphi}{\parallel}+2 \\
{\parallel}{\neg\varphi}{\parallel} &:= & {\parallel}{\varphi}{\parallel}+1  \quad\quad
\end{array}\]
and for all words $\alpha$ over $A\cup\Laa$, let ${\parallel}{\alpha}{\parallel}$ be the nonnegative integer inductively defined as:
\[\begin{array}{lllllllll}
{\parallel}{\epsilon}{\parallel}&:=&0 \quad\quad\quad & {\parallel}{\alpha a}{\parallel}&:=&{\parallel}{\alpha}{\parallel}+1 \quad\quad\quad& {\parallel}{\alpha\psi}{\parallel}&:=&{\parallel}{\alpha}{\parallel}+{\parallel}{\psi}{\parallel}
\end{array}\]
Then, for all formulas $\varphi$, let $\deg(\varphi)$ be the nonnegative integer inductively defined as follows (this is often known as the \emph{modal depth} of a formula, the maximum stack of epistemic modalities potentially occurring in it):
\[\begin{array}{llllll}
\deg(p)&=&0\quad\quad\quad & \deg(B_{a}\varphi)&=&\deg(\varphi)+1 \\
\deg(\bot)&=&0 \quad\quad\quad & \deg(\lbrack\varphi\rbrack\psi)&=&\deg(\varphi)+\deg(\psi) \\
\deg(\neg\varphi)&=&\deg(\varphi) \quad\quad\quad & \deg(\lbrack a\rbrack\varphi)&=&\deg(\varphi) \\
\deg(\varphi\vee\psi)&=&\max\{\deg(\varphi),\deg(\psi)\}\quad\quad
\end{array}\]
Also, given a pair of the form $(\alpha,\varphi)$ where $\alpha$ is a history and $\varphi\in\Laa$, \[\begin{array}{lll} \deg(\alpha,\varphi) & = & \deg(\lbrack\alpha\rbrack\varphi) \end{array}\]
Finally, let $\ll$ be the well-founded order between (history, formula) pairs defined as follows:
\[ \begin{array}{lllll}
(\alpha,\varphi)\ll(\beta,\psi) & \text{iff} & \text{either} & \deg(\alpha,\varphi)<\deg(\beta,\psi), \\ && \text{or} & \deg(\alpha,\varphi)=\deg(\beta,\psi) \text{ and } {\parallel}{\alpha}{\parallel}+{\parallel}{\varphi}{\parallel}<{\parallel}{\beta}{\parallel}+{\parallel}{\psi}{\parallel}. \end{array}\]
Various results for these orders are found in the Appendix on page \pageref{appendix}.

\subsection{Semantics of asynchronous announcement logic}
A {\em model} is a triple $(W,R,V)$, where $W$ is a nonempty set of states, $R: A \imp \powerset(W \times W)$ maps each $a\in A$ to a binary accessibility relation $R_a$ on $W$, and $V: P \imp \powerset(W)$ maps each atom $p$ to the set $V(p)$ of states in $W$ where $p$ is true. 
\begin{definition}[Semantics]
Given a model $(W,R,V)$, we simultaneously define the {\em agreement relation} $\bowtie$ between states and histories and the {\em satisfaction relation} $\models$ between pairs of states and histories, and formulas. The model is left implicit in these relations. 
\[ \begin{array}{lcl}
s\bowtie\epsilon \\
s\bowtie\alpha a & \text{iff} & s\bowtie\alpha \text{ and } {\mid}{\alpha}{\mid}_{a}<{\mid}{\alpha}{\mid}_! \\
s\bowtie\alpha\psi & \text{iff} & s\bowtie\alpha \text{ and } s,\alpha\models\psi \\
\ \\
s,\alpha\models p & \text{iff} & s\in V(p) \\ s,\alpha\not\models\bot \\
s,\alpha\models\neg\varphi & \text{iff} & s,\alpha\not\models\varphi \\
s,\alpha\models\varphi\vee\psi &\text{iff} & s,\alpha\models\varphi \text{ or } s,\alpha\models\psi \\
s,\alpha\models B_a\varphi & \text{iff} & t,\beta\models\varphi \text{ for all } t\in W\ \text{and for all histories}\ \beta \\ && \text{such that } sR_{a}t, \alpha\triangleright_a\beta, \text{ and } t\bowtie\beta \\
s,\alpha\models\lbrack\varphi\rbrack\psi & \text{iff} & s,\alpha\models\varphi \text{ implies } s,\alpha\varphi\models\psi \\
s,\alpha\models\lbrack a\rbrack\varphi & \text{iff} & {\mid}{\alpha}{\mid}_{a}<{\mid}{\alpha}{\mid}_! \text{ implies } s,\alpha a\models\varphi
\end{array} \]
A formula $\phi$ is {\em $\epsilon$-valid} (or {\em valid}), notation $\models^\epsilon \phi$ (or $\models \phi$), iff for all models $(W,R,V)$ and for all $s \in W$, $s,\epsilon \models \phi$. The set of validities is called $AA^\epsilon$ (or $AA$), for {\em asynchronous announcement logic}.
A formula $\phi$ is {\em $*$-valid} (or {\em always valid}), notation $\wordmodels \phi$, iff for all histories $\alpha$, $\models [\alpha]\phi$; further, $\phi$ is {\em $\epsilon$-satisfiable} (or {\em satisfiable}) iff there are $(W,R,V)$ and $s \in S$ such that $s,\epsilon \models \phi$, and $\phi$ is {\em $*$-satisfiable} (or {\em sometimes satisfiable}) if there is a history $\alpha$ such that $\dia{\alpha}\phi$ is $\epsilon$-satisfiable. The set of $*$-validities is called $AA^*$.
\end{definition}
Thanks to the items~$1$--$7$ of Lemma~\ref{lemma:let:be:the:well:founded} in the Appendix, the reader may verify that the definitions of the relations ``agrees with'' and ``satisfies'' are well-founded.

Importantly, the meaning of $[\phi]\psi$ in $AA$ is different from the meaning of $[\phi]\psi$ in $PAL$.

As dynamic epistemic logics go, $AA$ is unusual because dynamic modalities do not result in model transformations. 
Such transformations are implicit in the history. Given a model $M = (W,R,V)$, we can easily see the clause for announcement as a model transformer: as the truth of $[\phi]\psi$ is conditional on the truth of $\phi$, the states in the domain $W$ that survive this operation are exactly the $\phi$-restriction, as in $PAL$.
However, to interpret the $\psi$ bound by announcement $\phi$, we may have to access the model prior to that announcement. In that respect our models are rather like the protocol-generated forests of~\cite{jfaketal.JPL:2009}, however with the additional complication of uncertainty of reception of announcements by other agents, which is made precise in the $[a]\phi$ and $B_a\phi$ semantics.
The relation of $AA$ to history-based semantics is addressed in Section \ref{sec.history}.

\subsection{Examples} \label{sec.examples}

We continue with examples of validities and non-validities.
\begin{example} \label{ex.pakap}
We have $\models [p][a] B_a p$.
The formula $[p][a] B_a p$ stands for `after announcement of factual information $p$, and subsequent reception by agent $a$, agent $a$ knows that $p$'. To show that it is valid, is elementary.
Take any $(W,R,V)$, $s \in W$. Then the following conditions are equivalent:
\begin{itemize}
\item $s,\epsilon \models [p][a] B_a p$,
\item if $s,\epsilon \models p \text{ then } s,p \models [a] B_a p$,
\item if $s,\epsilon \models p \text{ then } s,pa \models B_a p$,
\item if $s,\epsilon \models p \text{ then } t,\beta \models p \text{ for all } t\in W\ \text{and for all histories}\ \beta \text{ such that } sR_{a}t, t\bowtie\beta \text{ and } pa\triangleright_a \beta$.
\end{itemize}
If $b$ is the only other agent, the possible histories $\beta$ such that $pa\triangleright_{a}\beta$ are: $pa,pba,pab$.
As they all contain the announcement $p$, the above conditions are true.
\end{example}
\begin{example}\label{ex.notvalid}
On the other hand, $\not\models [p] B_a p$. Let $p$ be true but not known to agent $a$, as in the model $p(s)$---$a$---$\overline{p}(t)$. Then $s,\epsilon \models p$, and therefore $s \bowtie p$. But we do not have $s,p \models B_a p$: from $p \tri_a \epsilon$, $R_ast$, $t \bowtie \epsilon$, and $t,\epsilon \models \neg p$ it follows that $s, p \models \M_a \neg p$. In fact, because $p \tri_a \beta$ iff $\epsilon \tri_a \beta$, we have that $[p] B_a p$ is equivalent to $p \imp B_a p$.
\end{example}
\begin{example}
Next, $\not\wordmodels [p][a] B_a p$.
Consider a model wherein agent $a$ initially is uncertain about the truth of $p$ and wherein in actual state $s$ variables $p$ and $q$ are true with $p\not=q$, e.g.\ $pq$---$a$---$\overline{p}q$.
The left state is $s$, let the right state be $t$.
Then $s,q \models \dia{p}\dia{a}\M_a \neg p$, seeing that $s,\epsilon \models q$, $s,q \models p$, $|qp|_a < |qp|_!$ and $s,qpa \models \M_a \neg p$.
Differently said, given the history $qpa$, in the event wherein $a$ receives ``the next announcement,'' it receives the information that $q$ contained in the first announcement, not the information that $p$ contained in the second announcement which agent $a$ will receive next if she reads again.
\end{example}
\begin{example} We also have $\wordmodels [a]\bot \imp [p][a] B_a p$. For any state $s$ and history $\alpha$, $[a]\bot$ is only true in $(s,\alpha)$ if agent $a$ has received all announcements in the history $\alpha$ (i.e., $|\alpha|_! = |\alpha|_a$). This means that if a further announcement is made, such as $p$, and $a$ then receives `the next announcement', that must be the annoucement of $p$ just made. After that, $B_a p$ is true.
Similarly, $\wordmodels [p][a] ([a]\bot \imp B_a p)$. It will also be clear that $\models [a]\bot$, but $\not\wordmodels [a]\bot$.
\end{example}
\begin{example}
In Figure \ref{fig}(iv), after Anne and Bill have both received the announcement $p \vel q$, they both know $p \vel q$: $B_a (p \vel q) \et B_b (p \vel q)$ is now true. For this we can, as usual, write $E_{ab} (p \vel q)$ (everybody knows $p \vel q$). But they do not know that the other knows $p \vel q$. However, after the announcement of $E_{ab} (p \vel q)$ and both receiving it we obtain $E^2_{ab} (p \vel q)$: everybody knows that everybody knows $p \vel q$. And so on. Anne and Bill can achieve any finite approximation of common knowledge, but they cannot get common knowledge of $p \vel q$.

With individually received messages no growth of common knowledge will ever occur, unlike in $PAL$ where reception is synchronous \cite{halpernmoses:1990,mosesetal:1986}. But we can gradually construct so-called {\em concurrent common knowledge} \cite{halpernmoses:1990,panangadenetal:1992}, as above.
\end{example}
Section \ref{sec.dc} contains an extended example relating the semantics of $AA$ to modelling processes sending and receiving events in distributed computing.

\subsection{Validities and other results for the semantics} \label{sec.resultssemantics}

We continue with results relating the satisfaction relation and the agreement relation, and with some fairly general always-validities.
In the following result, $p$ being a propositional variable and $\varphi$ being a formula, for all formulas $\chi$ possibly containing a specific occurrence of $p$, the expression $\chi\lbrack p/\varphi\rbrack$ will denote the formula obtained from $\chi$ by replacing this specific occurrence by $\varphi$ and for all histories $\gamma$ possibly containing a specific occurrence of $p$, the expression $\gamma\lbrack p/\varphi\rbrack$ will denote the history obtained from $\gamma$ by replacing this specific occurrence by $\varphi$.
\begin{lemma}\label{lemma:simple:properties:about:replacing:by:equivalent:formulas}
Let $(W,R,V)$ be a model.
Let $p$ be a propositional variable.
Let $\varphi,\psi$ be formulas such that for all $s\in W$ and for all histories $\alpha$, $s,\alpha\models\varphi$ iff $s,\alpha\models\psi$.
Let $\chi$ be a formula possibly containing a specific occurrence of $p$ and $\gamma$ be a history possibly containing a specific occurrence of $p$.
For all $s\in W$, the following conditions hold:
\begin{itemize}
\item $s\bowtie\gamma\lbrack p/\varphi\rbrack$ iff $s\bowtie\gamma\lbrack p/\psi\rbrack$,
\item $s,\gamma\lbrack p/\varphi\rbrack\models\chi$ iff $s,\gamma\lbrack p/\psi\rbrack\models\chi$,
\item $s,\gamma\models\chi\lbrack p/\varphi\rbrack$ iff $s,\gamma\models\chi\lbrack p/\psi\rbrack$.
\end{itemize}
\end{lemma}
\begin{proof}
The proof is by $\ll$-induction on $(\gamma,\chi)$.
\end{proof}
\begin{lemma} \label{lemma.prefixes}
Let $(W,R,V)$ be a model.
Let $s$ be a state and let $\alpha$ be a history.
If $s\bowtie\alpha$ and $\beta\subseteq\alpha$, then $s\bowtie\beta$.
\end{lemma}
\begin{proof}
The proof is by induction on ${\mid}\alpha{\mid}$.
\end{proof}
\begin{lemma} \label{lemma:concatenationdiamond}
Let $(W,R,V)$ be a model.
Let $\alpha$ be a history and $\beta$ be a word.
For every formula $\chi$ and for every world $s$ such that $s\bowtie\alpha$, $s,\alpha\models\langle\beta\rangle\chi$ if and only if (i) the concatenation $\alpha\beta$ is a history, (ii) $s\bowtie\alpha\beta$, and (iii) $s,\alpha\beta\models\chi$.
\end{lemma}
\begin{proof}
The proof is by induction on ${\mid}\beta{\mid}$.
\\
Case ``$\beta=\epsilon$''.
Left to the reader.
\\
Case ``$\beta=a\beta'$''.
Suppose $s,\alpha\models\langle a \rangle\langle\beta'\rangle\chi$.
Hence, $|\alpha|_a<|\alpha|_!$.
Thus, $\alpha a$ is a history and $s\bowtie \alpha a$.
Moreover, $s,\alpha a\models \langle\beta'\rangle\chi$.
Consequently, by induction hypothesis, $\alpha a \beta'$ is a history, $s\bowtie \alpha a\beta'$ and $s, \alpha a\beta'\models\chi$.
Conversely, suppose $\alpha a \beta'$ is a history, $s\bowtie \alpha a\beta'$ and $s, \alpha a\beta'\models\chi$.
Hence, by induction hypothesiss $s,\alpha a\models \langle\beta'\rangle\chi$.
Thus $s,\alpha\models\langle a \rangle\langle\beta'\rangle\chi$.
\\
Case ``$\beta=\psi\beta'$''.
Suppose $s,\alpha\models\langle \psi \rangle\langle\beta'\rangle\chi$.
Hence, $s,\alpha\models \psi$ and $s,\alpha\psi\models\langle\beta'\rangle \chi$.
Thus, by induction hypothesis, $\alpha \psi \beta'$ is a history, $s\bowtie \alpha \psi\beta'$ and $s, \alpha \psi\beta'\models\chi$. Conversely, suppose $\alpha \psi \beta'$ is a history, $s\bowtie \alpha \psi\beta'$ and $s, \alpha \psi\beta'\models\chi$.
Consequently, $s\bowtie \alpha \psi$ and $s,\alpha\models\psi$.
Moreover, by induction hypothesis, $s, \alpha \psi\models\langle\beta'\rangle\chi$.
Hence, $s,\alpha\models\langle \psi \rangle\langle\beta'\rangle\chi$.
\end{proof}
\begin{lemma}\label{lemma:about:boxes:and:diamonds}
Let $(W,R,V)$ be a model, $\alpha$ be a history, $\beta$ be a word over $A\cup\Laa$ and $\varphi$ be a formula.
For all $s\in W$,
\begin{itemize}
\item $s,\alpha\models\langle\beta\rangle\varphi$ iff $s,\alpha\not\models\lbrack\beta\rbrack\neg\varphi$,
\item $s,\alpha\models\lbrack\beta\rbrack\varphi$ iff $s,\alpha\not\models\langle\beta\rangle\neg\varphi$.
\end{itemize}
\end{lemma}
\begin{proof}
The proof is by $<$-induction on ${\mid}\beta{\mid}$.
\end{proof}
\begin{lemma}\label{lemma:let:be:a:model:and}
Let $\beta$ be a word over $A\cup\Laa$.
For all models $(W,R,V)$, for all $s\in W$ and for all formulas $\varphi$,
\begin{enumerate}
\item $s,\epsilon\models\langle\beta\rangle\varphi$ iff $\beta$ is a history, $s\bowtie\beta$ and $s,\beta\models\varphi$,
\item $s,\epsilon\models\lbrack\beta\rbrack\varphi$ iff, if $\beta$ is a history and $s\bowtie\beta$, then $s,\beta\models\varphi$.
\end{enumerate}
\end{lemma}
\begin{proof}
By Lemma \ref{lemma:concatenationdiamond} and Lemma \ref{lemma:about:boxes:and:diamonds}.
\end{proof}
\begin{corollary}\label{corollary:models:histories:bowtie:aa}
For all $\varphi\in\Laa$, $\wordmodels \phi$ iff for all models $(W,R,V)$, for all $s \in W$ and for all histories $\alpha$, if $s \bowtie \alpha$ then $s,\alpha \models \phi$.
\end{corollary}
\begin{proof}
By Lemma~\ref{lemma:let:be:a:model:and}.
\end{proof}
We note that the formulation of Corollary~\ref{corollary:models:histories:bowtie:aa} could well have served as an alternative \emph{definition} of $*$-validity, instead of ``for all histories $\alpha$, $\models [\alpha]\phi$.''
\begin{lemma}\label{premier:lemme:pour:bowtie}
Let $(W,R,V)$ be a model.
For all histories $\alpha$ and for all states $s$, the following conditions are equivalent:
\begin{enumerate}
\item $s\bowtie\alpha$,
\item for all histories $\beta$, for all words $\gamma$ and for all formulas $\varphi$, if $\alpha=\beta\varphi\gamma$ then $s,\beta\models\varphi$.
\end{enumerate}
\end{lemma}
\begin{proof}
Let $\alpha$ be a history and $s$ be a state.
\\
$(1\Rightarrow2).$
Suppose $s \bowtie \alpha$.
Let $\beta$ be a history, $\gamma$ be a word and $\varphi$ be a formula such that $\alpha=\beta\phi\gamma$.
Since $s \bowtie \alpha$, $s\bowtie\beta\phi$.
Hence, $s\bowtie\beta$ and $s,\beta\models\phi$.
\\
$(2\Rightarrow1).$
Suppose for all histories $\beta$, for all words $\gamma$ and for all formulas $\varphi$, if $\alpha=\beta\varphi\gamma$ then $s,\beta\models\varphi$.
We prove that $s \bowtie \alpha$ by $<$-induction on ${\mid}\alpha{\mid}$.
\\
Case ``$\alpha=\epsilon$''.
Then $s \bowtie \alpha$.
\\
Case ``$\alpha=\alpha^{\prime}a$''.
Since $\alpha$ is a history, $\alpha^{\prime}$ is a history such that ${\mid}\alpha^{\prime}{\mid}_!\geq{\mid}\alpha^{\prime}{\mid}_{a}$.
Moreover, for all histories $\beta$, for all words $\gamma$ and for all formulas $\varphi$, if $\alpha^{\prime}=\beta\varphi\gamma$ then $\alpha=\beta\varphi\gamma a$ and, by our hypothesis, $s,\beta\models\varphi$.
Thus, by induction hypothesis, $s\bowtie\alpha^{\prime}$.
Since ${\mid}\alpha^{\prime}{\mid}_!\geq{\mid}\alpha^{\prime}{\mid}_{a}$,  $s\bowtie\alpha$.
\\
Case ``$\alpha=\alpha^{\prime}\psi$''.
Since $\alpha$ is a history,  $\alpha^{\prime}$ is a history.
Moreover, by our hypothesis, $s,\alpha^{\prime}\models\psi$.
Consequently, $s\bowtie\alpha$.
\end{proof}
We continue with some results for always-validity $\models^*$.
\begin{proposition} \label{prop.starepsilon}
Let $\phi \in \Laa$.
If $\models^* \phi$ then $\models \phi$.
\end{proposition}
\begin{proof}
Suppose $\models^* \phi$.
Hence, $\models\lbrack\epsilon\rbrack\varphi$.
Thus, $\models\varphi$.
\end{proof}
\begin{proposition} \label{lemma.K}
Let $\phi,\psi \in \Laa$ and $a\in A$.
Then:
\begin{enumerate}
\item $\wordmodels \phi$ implies $\wordmodels B_a \phi$,
\item $\wordmodels B_a (\phi \imp \psi) \imp (B_a \phi \imp B_a \psi)$.
\end{enumerate}
\end{proposition}
\begin{proof} \ \vspace{-.2cm}
\begin{enumerate}
\item Suppose $\not\wordmodels B_{a}\varphi$.
Hence, by Corollary \ref{corollary:models:histories:bowtie:aa}, let $(W,R,V)$ be a model, $\alpha$ be a history and $s\in W$ be such that $s\bowtie\alpha$ and $s,\alpha\not\models B_{a}\varphi$.
Let $t\in W$ and $\beta$ be a history such that $sR_{a}t$, $\alpha\triangleright_{a}\beta$, $t\bowtie\beta$ and $t,\beta\not\models\varphi$.
Thus, by Lemma~\ref{lemma:let:be:a:model:and}, $t,\epsilon\not\models\lbrack\beta\rbrack\varphi$.
Consequently, $\not\wordmodels\varphi$.
\item Suppose $\not\wordmodels B_{a}(\varphi\rightarrow\psi)\rightarrow(B_{a}\varphi\rightarrow B_{a}\psi)$.
Hence, by Corollary \ref{corollary:models:histories:bowtie:aa}, let $(W,R,V)$ be a model, $\alpha$ be a history and $s\in W$ be such that $s\bowtie\alpha$ and $s,\epsilon\not\models B_{a}(\varphi\rightarrow\psi)\rightarrow(B_{a}\varphi\rightarrow B_{a}\psi)$.
Thus, $s,\alpha\models B_{a}(\varphi\rightarrow\psi)$, $s,\alpha\models B_{a}\varphi$ and $s,\alpha\not\models B_{a}\psi$.
Let $t\in W$ and $\beta$ be a history such that $sR_{a}t$, $\alpha\triangleright_{a}\beta$, $t\bowtie\beta$ and $t,\beta\not\models\psi$.
Since $s,\alpha\models B_{a}(\varphi\rightarrow\psi)$ and $s,\alpha\models B_{a}\varphi$, we obtain that $t,\beta\models\varphi\rightarrow\psi$ and $t,\beta\models\varphi$.
Thus, $t,\beta\models\psi$: a contradiction.
\end{enumerate}
\end{proof}
\begin{lemma} \label{lemma:equivalent:formulas::psi:eta:words:become:equivalent:too}
Let $\phi,\psi$ be formulas such that $\models^*\phi\leftrightarrow\psi$.
Let $M = (W,R,V)$ be a model.
Let $\alpha$ be a history.
Let $\beta$ be a word such that $\alpha\phi\beta$ and $\alpha\psi\beta$ are histories and let $\chi$ be a formula.
For all states $s \in W$,  $s\bowtie \alpha\phi\beta$ iff $s\bowtie \alpha\psi\beta$, and $s,\alpha\phi\beta\models\chi$ iff $s,\alpha\psi\beta\models\chi$.
\end{lemma}
The proof of this lemma is found in the Appendix.

\begin{proposition}[Substitution of equivalents] \label{prop.substeq}
Let $\varphi,\psi$ be formulas such that $\wordmodels \phi \leftrightarrow \psi$.
For all formulas $\chi$ and for all atoms $p$, $\wordmodels \chi(p/\phi) \eq \chi(p/\psi)$.
\end{proposition}
\begin{proof}
The proof is by induction on $\chi$.
\\
Cases ``$\chi$ is an atom'', ``$\chi=\bot$'', ``$\chi=\neg\chi^{\prime}$'' and ``$\chi=\chi_{1}\vee\chi_{2}$''.
Left to the reader.
\\
Case ``$\chi=\lbrack\eta\rbrack\chi^{\prime}$''.
Let $(W,R,V)$ be a model, $s$ be a state and $\alpha$ be a history such that $s\bowtie\alpha$.
We have: $s,\alpha\models[\eta(p/\phi)]\chi^{\prime}(p/\phi)$ iff $s,\alpha\models\eta(p/\phi)$ implies $s,\alpha\eta(p/\phi)\models\chi^{\prime}(p/\phi)$ iff, by induction hypothesis and using Lemma \ref{lemma:equivalent:formulas::psi:eta:words:become:equivalent:too}, $s,\alpha\models\eta(p/\psi)$ implies $s,\alpha\eta(p/\psi)\models\chi^{\prime}(p/\psi)$ iff $s,\alpha\models[\eta(p/\psi)]\chi^{\prime}(p/\psi)$.
Since $(W,R,V)$, $s$ and $\alpha$ were arbitrary, $\wordmodels \chi(p/\phi) \eq \chi(p/\psi)$.
\\
Case ``$\chi=B_{a}\chi^{\prime}$''.
Let $(W,R,V)$ be a model, $s$ be a state and $\alpha$ be a history such that $s\bowtie\alpha$.
We have: $s,\alpha\models B_{a}\chi^{\prime}(p/\phi)$ iff for all states $t$ and for all histories $\beta$, if $sR_{a}t$, $\alpha\tri_{a}\beta$ and $t\bowtie\beta$ then $t,\beta\models\chi^{\prime}(p/\phi)$ iff, by induction hypothesis, for all states $t$ and for all histories $\beta$, if $sR_{a}t$, $\alpha\tri_{a}\beta$ and $t\bowtie\beta$ then $t,\beta\models\chi^{\prime}(p/\psi)$ iff $s,\alpha\models B_{a}\chi^{\prime}(p/\psi)$.
Since $(W,R,V)$, $s$ and $\alpha$ were arbitrary, $\wordmodels \chi(p/\phi) \eq \chi(p/\psi)$.
\end{proof}
\begin{lemma} \label{1234}
Let $(W,R,V)$ be a model, $s$ be a state and $a$ be an agent.
For all histories $\alpha,\beta$, if ${\alpha} \tri_a {\beta}$, $s\bowtie\alpha$ and $s\bowtie\beta$ then $s,\alpha \models B_a \phi$ iff $s, \beta \models B_a \phi$.
\end{lemma}
\begin{proof}
By Proposition \ref{lemma.tri}.
\end{proof}
In particular, it follows that $s,\alpha\models B_a\phi$ iff $s,\alpha_a\models B_a\phi$.
Let us now see some results concerning the positive fragment $\Lml^+$.
For this we will define a preorder on histories as follows: $\alpha\preceq\beta$ if and only if:
\begin{itemize}
\item $\alpha \proj_! \subseteq \beta\proj_!$;
\item for all $a\in A$, $|\alpha|_a\leq|\beta|_a$;
\item for every model $(W,R,V)$ and every state $s$, $s\bowtie\beta$ implies $s\bowtie\alpha$.
\end{itemize}
It is easy to see that $\preceq$ is a reflexive and transitive relation between histories.
\begin{lemma} \label{lemma.fiets}
Let histories $\alpha,\beta$ and $a \in A$ be given.
\begin{enumerate}
\item $\alpha\subseteq\beta$ implies $\alpha\preceq\beta$,
\item $\alpha_a\preceq \alpha$. \label{fiets.two}
\end{enumerate}
\end{lemma}
\begin{proof}
$(1).$
Suppose $\alpha\subseteq\beta$.
Hence, $\alpha \proj_! \subseteq \beta\proj_!$ and for all $a\in A$, $|\alpha|_a\leq|\beta|_a$.
Now, let $(W,R,V)$ be a model and $s$ be a state such that $s\bowtie\beta$.
Thus, by Lemma \ref{lemma.prefixes}, $s \bowtie \alpha$.
Since $(W,R,V)$ and $s$ were arbitrary,  $\alpha\preceq\beta$.
\\
$(2).$
From the construction of $\alpha_a$ (see  Section~\ref{sec.resultsforhistories}) it follows that $\alpha_a=\gamma a^n$ for some history $\gamma$ such that $\gamma \subseteq \alpha$ and $n = |\gamma|_! - |\gamma|_a$.
Moreover, $\alpha\proj_! = \alpha_a\proj_!$,  $|\alpha|_a = |\alpha_a|_a$.
Now, let $(W,R,V)$ be a model and $s$ be a state such that $s\bowtie\alpha$.
Thus, by Lemma \ref{lemma.prefixes}, $s \bowtie\gamma$.
Since $\alpha_a=\gamma a^n$ and $n = |\gamma|_! - |\gamma|_a$,  $s \bowtie \alpha_a$.
Since $\alpha\proj_! = \alpha_a\proj_!$ and $|\alpha|_a = |\alpha_a|_a$,  $\alpha_a \preceq \alpha$.
\end{proof}
\begin{lemma} \label{5678}
Let $\alpha$, $\beta$ and $\gamma$ be histories.
If $\gamma\preceq\alpha$ and $\alpha \tri_a \beta$ then there exists a history $\delta$ such that $\gamma \tri_a \delta$ and $\delta\preceq\beta$.
\end{lemma}
\begin{proof}
Suppose $\gamma\preceq\alpha$ and $\alpha \tri_a \beta$.
Hence, $\gamma \proj_! \subseteq \alpha\proj_!$ and $|\gamma|_a\leq |\alpha|_a$.
Moreover, ${\mid}{\beta}{\mid}_{a}={\mid}{\alpha}{\mid}_{a}$, ${\beta}\proj_{!a} = {\alpha}\proj_{!a}$ and ${\mid}{\beta}{\mid}_!={\mid}{\alpha}{\mid}_{a}$.
Thus, $|\gamma|_a\leq|\beta|_a$.
Let $\beta^{\prime}$ be the initial segment of $\beta$ up until the $|\gamma|_a$-th occurrence of $a$.
Consequently, ${\mid}{\beta^{\prime}}{\mid}_{a}={\mid}{\gamma}{\mid}_{a}$.
Moreover, $\beta^{\prime}\subseteq\beta$.
Hence, by Lemma \ref{lemma.fiets}, $\beta^{\prime}\preceq\beta$.
Let $\delta=\beta^{\prime}_{a}$.
Thus, by Lemma \ref{lemma:about:alpha:a:definition:serial:tri}, $\beta^{\prime} {\upharpoonright_{!a}} =  \delta {\upharpoonright_{!a}} =  \delta {\upharpoonright_!}$ and ${\mid}{\delta}{\mid}_{a}={\mid}{\beta^{\prime}}{\mid}_{a}$.
Moreover, by Lemma \ref{lemma.fiets}, $\delta\preceq\beta^{\prime}$.
Since $\beta^{\prime}\preceq\beta$,  $\delta\preceq\beta$.
Since ${\mid}{\beta^{\prime}}{\mid}_{a}={\mid}{\gamma}{\mid}_{a}$ and ${\mid}{\delta}{\mid}_{a}={\mid}{\beta^{\prime}}{\mid}_{a}$, we obtain that ${\mid}{\delta}{\mid}_{a}={\mid}{\gamma}{\mid}_{a}$.
From all this, it follows that $\gamma \tri_a \delta$ and $\delta\preceq\beta$.
\end{proof}
\begin{lemma}\label{lem:alphaleqbeta}
Let $(W,R,V)$ be a model.
Let $\phi\in\Lml^+$.
For all states $s$ and for all histories $\alpha',\alpha$, if $\alpha'\preceq\alpha$, $s\bowtie\alpha$ and $s,\alpha'\models\phi$ then $s,\alpha\models\phi$.
\end{lemma}
\begin{proof}
The proof is by induction on $\varphi$.
\\
Cases ``$\varphi=p$'', ``$\varphi=\neg p$'', ``$\varphi=\bot$'', ``$\varphi=\top$'', ``$\varphi=\psi\vee\chi$'' and ``$\varphi=\psi\wedge\chi$''.
Left to the reader.
\\
Case ``$\varphi=B_{a}\psi$''.
Let $s$ be a state and $\alpha',\alpha$ be histories such that $\alpha'\preceq\alpha$, $s\bowtie\alpha$, $s,\alpha'\models B_{a}\psi$ and $s,\alpha\not\models B_{a}\psi$.
Let $t$ be a state and $\beta$ be a history such that $sR_{a}t$, $\alpha\tri_a\beta$, $t\bowtie\beta$ and $t,\beta\not\models\psi$.
Since $\alpha'\preceq\alpha$, by Lemma \ref{5678}, let $\beta'$ be a history such that $\alpha'\tri_a\beta'$ and $\beta^{\prime}\preceq\beta$.
Since $t \bowtie \beta$,  $t\bowtie\beta'$.
Since $s,\alpha'\models B_a\psi$, $sR_{a}t$ and $\alpha'\tri_a\beta'$, we obtain that $t,\beta'\models \psi$.
Since $\beta^{\prime}\preceq\beta$ and $t\bowtie\beta$, by induction hypothesis, $t,\beta\models \psi$: a contradiction.
\end{proof}
With this lemma in hand, we can now easily demonstrate that:
\begin{proposition}[Positive is preserved] \label{9012}
For all $\phi \in \Lml^+$ and words $\alpha$, $\models^* \phi \imp [\alpha]\phi$.
\end{proposition}
\begin{proof}
Let $\phi \in \Lml^+$ and $\alpha$ be a word such that $\not\models^* \phi \imp [\alpha]\phi$.
Hence, by Corollary \ref{corollary:models:histories:bowtie:aa}, let $(W,R,V)$ be a model, $s$ be a state and $\beta$ be a history such that $s\bowtie\beta$ and $s,\beta\not\models\phi \imp [\alpha]\phi$.
Thus, $s,\beta\models \phi$ and $s,\beta\not\models\lbrack\alpha\rbrack\varphi$.
Consequently, by Lemma \ref{lemma:concatenationdiamond} and Lemma \ref{lemma:about:boxes:and:diamonds}, the concatenation $\beta\alpha$ is a history, $s\bowtie\beta\alpha$, and $s,\beta\alpha\not\models\varphi$.
Hence, $\beta\subseteq\beta\alpha$.
Thus, by Lemma \ref{lemma.fiets}, $\beta\preceq\beta\alpha$.
Since $s\bowtie\beta\alpha$ and $s,\beta\models \phi$, by Lemma \ref{lem:alphaleqbeta}, $s,\beta\alpha\models\varphi$: a contradiction.
\end{proof}

\subsection{Relation to standard modal logic} \label{sec.ml}

If we restrict the language $\Laa$ to $\Lel$, the fragment without dynamic modalities, there is an interesting relation between `valid' and `always valid', and the standard modal logic {\bf K}.
Let $\models^K$ be the standard satisfaction relation (`Kripke semantics') for the language $\Lel$ on models $(W,R,V)$; so that, in particular, $M,s \models^K B_a \phi$ iff $M,t \models^K \phi$ for all $t$ such that $sR_a t$. To properly describe the relation between $\models^K$ and (asynchronous announcement semantics) $\models$ we introduce the notion of a \emph{flat model}.
\begin{definition}[Flat model] \label{def.flatmodel}
Given a model $M = (W,R,V)$, a state $s \in W$, and a history $\alpha$ such that $s \bowtie \alpha$. Let $\triangleright$ be the reflexive transitive closure of the union of all $\triangleright_a$. The \emph{flat model} $\overline{M^{s\alpha}} = (W^{s\alpha},R^{s\alpha},V^{s\alpha})$ is defined as follows.\footnote{The `overline' in $\overline{M^{s\alpha}}$ is used to suggest weight `flattening' the epistemic and temporal aspects of $M$ and $(s,\alpha)$ into a model with merely epistemic aspects. In the old days of manual proof correction similar notation was used to `push down' an upper case letter into the lower case correction.}
\[\begin{array}{lcl}
W^{s\alpha} & := & \{ (t,\beta) \mid t \in W\ \text{and}\ \alpha \triangleright \beta\ \text{and}\ t \bowtie \beta \}, \\
(t,\beta)R^{s\alpha}_{a}(t',\beta') & \text{:iff} & tR_at' \text{ and } \beta \triangleright_a \beta', \\
(t,\beta) \in V^{s\alpha}(p) & \text{:iff} & t \in V(p).
\end{array} \]
\end{definition}
\begin{proposition} \label{lemma.eqeq}
For all $\phi \in \Lel$, (i) $\models^K \phi$ iff $\models \phi$, and (ii) $\models \phi$ iff $\wordmodels \phi$.
\end{proposition}
\begin{proof}
The first item follows from the fact that for all histories $\beta$, if $\epsilon\triangleright_{a}\beta$ then $\beta=\epsilon$.
As for the second item, remark that if $\wordmodels \phi$ then $\models\lbrack\epsilon\rbrack\phi$ and, therefore, $\models\phi$. (We also recall Proposition \ref{prop.starepsilon}, for arbitrary $\phi\in\Laa$, of which the previous is a special case.)  Reciprocally, to prove that $\models \phi$ implies $\wordmodels \phi$, we now first prove that $\models^K \phi$ implies $\wordmodels \phi$, by way of proving the contrapositive: if $\phi$ is sometimes satisfiable ($*$-satisfiable), then $\phi$ is satisfiable in ordinary Kripke semantics.

Let $M = (W,R,V)$, $s \in W$, and history $\alpha$ with $s \bowtie \alpha$ be given. Consider the flat model $\overline{M^{s\alpha}}$ for $M$, $s$, and $\alpha$ (Def.~\ref{def.flatmodel}). We now prove by induction on $\phi\in\Lel$ that for all $(t,\beta) \in W^{s\alpha}$, $M, t, \beta \models \phi$ iff $\overline{M^{s\alpha}}, (t, \beta) \models^K \phi$ (where for clarity we write $M$ explicitly in expressions like $M,t,\beta\models\phi$). The only case of interest is that for the modality $B_{a}$.

\bigskip

\noindent $
M, t, \beta \models B_a \phi \\
\Eq \\
M, t^{\prime}, \beta^{\prime} \models \phi \text{ for all } t^{\prime}, \beta^{\prime} \text{ such that } tR_{a}t^{\prime}, \beta \triangleright_a \beta^{\prime} \text{ and } t^{\prime} \bowtie \beta^{\prime} \\
\Eq \quad\quad \text{induction} \\
\overline{M^{s\alpha}}, (t^{\prime}, \beta^{\prime}) \models^K \phi \text{ for all } t^{\prime}, \beta^{\prime} \text{ such that } tR_{a}t^{\prime}, \beta \triangleright_a \beta^{\prime} \text{ and } t^{\prime} \bowtie \beta^{\prime} \\
\Eq \\
\overline{M^{s\alpha}}, (t^{\prime}, \beta^{\prime}) \models^K \phi \text{ for all } (t^{\prime}, \beta^{\prime}) \text{ such that } (t,\beta)R^{s\alpha}_a(t^{\prime},\beta^{\prime}) \\
\Eq \\
\overline{M^{s\alpha}}, (t,\beta) \models^K B_a\phi
$

\bigskip

\noindent Having shown this, in particular it holds that $M, s, \alpha \models \phi$ iff $\overline{M^{s\alpha}}, (s, \alpha) \models^K \phi$. As $s$ and $\alpha$ were arbitrary, it follows that $\models^K \phi$ implies $\models^* \phi$, as required. From $\models \phi$ iff $\models^K \phi$, and $\models^K \phi$ implies $\wordmodels \phi$, it now follows that $\models \phi$ implies $\wordmodels \phi$.
\end{proof}
From Lemma \ref{lemma.eqeq} it follows that for all three semantics the logic (with respect to the language $\Lel$ only) is the minimal modal logic {\bf K}.

In particular, this means that the standard axiom $B_a (\phi \imp \psi) \imp (B_a \phi \imp B_a \psi)$ (K axiom), and the standard rules `$\phi$ implies $B_a \phi$' (Necessitation) and `$\phi$ iff $\phi(p/\psi)$' (Uniform Substition) of the modal logic {\bf K} are valid, resp., validity preserving. For example, for $\phi,\psi\in\lang_{ml}$ we have that $\wordmodels \phi$ iff (Prop.~\ref{lemma.eqeq}) $\models^K \phi$ iff (in {\bf K}) $\models^K \phi(p/\psi)$ iff (Prop.~\ref{lemma.eqeq}) $\wordmodels \phi(p/\psi)$.

The axiom $K$ and the rule Necessitation for $B_a$ were already established in Proposition \ref{lemma.K}, for all formulas $\phi\in\Laa$. It is therefore relevant to observe that Uniform Substitution is not conservative for this extension $\Laa$.

\begin{example} There exist $\varphi,\psi\in\Laa$ such that $\wordmodels\varphi$ and $\not\wordmodels\varphi(p/\psi)$, for instance $\varphi=\lbrack p\rbrack\lbrack a\rbrack(\lbrack a\rbrack\bot\rightarrow B_{a}p)$ and $\psi=p\wedge\neg B_{a}p$. The thing is that $\wordmodels\lbrack p\rbrack\lbrack a\rbrack(\lbrack a\rbrack\bot\rightarrow B_{a}p)$, but $s,\epsilon\not\models\lbrack p\wedge\neg B_{a}p\rbrack\lbrack a\rbrack(\lbrack a\rbrack\bot\rightarrow B_{a}(p\wedge\neg B_{a} p))$ where $s$ is a state in the model $(W,R,V)$ defined by $W=\{s,t\}$, $R_a=W\times W$ and $V(p)=\{s\}$.
\end{example}


\section{Axiomatization} \label{sec.ax}

In Section~\ref{sec.elim} we axiomatize $AA^\epsilon$, the $\epsilon$-validities. This is a reduction system eliminating reception and announcement modalities. In Section~\ref{sec.star} we determine $*$-validities that are reduction axioms. However, we do not axiomatize $AA^*$. Section~\ref{sec.palaa} compares  $PAL$ reductions, $AA^\epsilon$ reductions, and $AA^*$ reductions.

\subsection{Axiomatization of AA} \label{sec.elim}
In this section, we present an axiomatization of $AA$ on the class of all models with empty histories.
We prove its completeness by showing that for all formulas $\varphi \in \Laa$, there exists a formula $\psi \in \Lel$ such that $\varphi\leftrightarrow\psi$ is valid in the class of all models with empty histories.
In other words, the dynamic modalities $[a]$ and $[\phi]$ can be eliminated from the language, as far as one is concerned with $\epsilon$-validity.
We will do this by using an truth preserving transformation $tr$.
The completeness proof therefore consists in showing that $\Laa$ is equally expressive as $\Lel$ on the class of all models with empty histories.
Similar results are well-known for $PAL$, but we consider them remarkable for its asynchronous version.
\begin{definition} \label{def.tr}
For all words $\alpha$ over $A\cup\Laa$ and for all $\Laa$-formulas $\varphi$, we inductively define the $\Lel$-formula $tr(\alpha,\varphi)$ as follows:
\begin{itemize}
\item $tr(\epsilon,\bot)=\bot$,
\item $tr(\alpha a,\bot)=tr(\alpha,\bot)$ if ${\mid}\alpha{\mid}_{a}<{\mid}\alpha{\mid}_{!}$,
\item $tr(\alpha a,\bot)=\top$ if ${\mid}\alpha{\mid}_{a}\geq{\mid}\alpha{\mid}_{!}$,
\item $tr(\alpha\varphi,\bot)=tr(\alpha,\varphi)\rightarrow tr(\alpha,\bot)$,
\item $tr(\alpha,p)=tr(\alpha,\bot)\vee p$,
\item $tr(\alpha,\neg\varphi)=tr(\alpha,\varphi)\rightarrow tr(\alpha,\bot)$,
\item $tr(\alpha,\varphi\vee\psi)=tr(\alpha,\varphi)\vee tr(\alpha,\psi)$,
\item $tr(\alpha,B_{a}\varphi)=tr(\alpha,\bot)\vee\bigwedge\{B_{a} tr(\beta,\varphi)\mid \alpha\triangleright_{a}\beta\}$,
\item $tr(\alpha,\lbrack a\rbrack\varphi)=tr(\alpha a,\varphi)$,
\item $tr(\alpha,\lbrack\varphi\rbrack\psi)=tr(\alpha\varphi,\psi)$.
\end{itemize}
\end{definition}
Let us remark that the above definition of the truth preserving translation $tr$ is indeed inductive, namely with respect to the well-founded order $\ll$ between (history, formula) pairs defined in Section~\ref{sec.aux} (Lemma~\ref{lemma:let:be:the:well:founded} in the Appendix).
\begin{lemma}\label{lemma:soundness:of:the:function:tr}
Let $(W,R,V)$ be a model and $s\in W$.
For all words $\alpha$ over $A\cup\Laa$ and for all formulas $\varphi$, $s,\epsilon\models tr(\alpha,\varphi)$ iff $s,\epsilon\models\lbrack\alpha\rbrack\varphi$.
\end{lemma}
The proof is found in the Appendix. In particular, for $\alpha=\epsilon$ and given that $[\epsilon]\phi=\phi$, we obtain that for all models, states and formulas: $s,\epsilon\models tr(\epsilon,\varphi)$ iff $s,\epsilon\models\varphi$, so that $\phi \mapsto tr(\epsilon,\phi)$ therefore defines a truth (value) preserving translation from $\Laa$ to $\Lel$.
\begin{corollary}[Elimination of dynamic modalities] \label{cor.elim} \ \\
For all $\phi\in \Laa$ there is a $\psi\in\Lml$ (namely $\psi=tr(\epsilon,\phi)$) such that $\models \phi \eq \psi$.
\end{corollary}

With these results in hand we will now present the axiomatization $\mathbf{AA}$.
The axioms of $\mathbf{AA}$ exactly follow the pattern of the translation function $tr$ of Def.~\ref{def.tr}.
\begin{definition}[Axiomatization $\mathbf{AA}$] \label{definition:axiomatisation:Laa}
Let $\mathbf{AA}$ be the axiomatization given by the following axioms and inference rules:
\begin{itemize}
\item the tautologies in the language $\Laa$,
\item the theorems of the least normal modal logic in the language $\Lml$,
\item the following axioms:
\begin{description}
\item[$(A1)$:] $\lbrack\alpha\rbrack p\leftrightarrow\lbrack\alpha\rbrack\bot\vee p$,
\item[$(A2)$:] $\lbrack\alpha a\rbrack\bot\leftrightarrow\lbrack\alpha\rbrack\bot$ \quad if ${\mid}\alpha{\mid}_{a}<{\mid}\alpha{\mid}_{!}$,
\item[$(A3)$:] $\lbrack\alpha a\rbrack\bot$ \quad if ${\mid}\alpha{\mid}_{a}\geq{\mid}\alpha{\mid}_{!}$,
\item[$(A4)$:] $\lbrack\alpha\varphi\rbrack\bot\leftrightarrow\lbrack\alpha\rbrack\bot\vee\neg\lbrack\alpha\rbrack\varphi$,
\item[$(A5)$:] $\lbrack\alpha\rbrack\neg\varphi\leftrightarrow\lbrack\alpha\rbrack\bot\vee\neg\lbrack\alpha\rbrack\varphi$,
\item[$(A6)$:] $\lbrack\alpha\rbrack(\varphi\vee\psi)\leftrightarrow\lbrack\alpha\rbrack\varphi\vee\lbrack\alpha\rbrack\psi$,
\item[$(A7)$:] $\lbrack\alpha\rbrack B_{a}\varphi\leftrightarrow\lbrack\alpha\rbrack\bot\vee\bigwedge\{B_{a}\lbrack\beta\rbrack\varphi \mid \alpha\triangleright_{a}\beta\}$.
\end{description}
\item the following inference rules:
\begin{description}
\item[$(MP)$:] from $\phi$ and $\phi\imp\psi$ infer $\psi$,
\item[$(R2)$:] from $\varphi\leftrightarrow\psi$ infer $B_{a}\varphi\leftrightarrow B_{a}\psi$,
\end{description}
\end{itemize}
The notion of $\mathbf{AA}$-proof being defined as usual, we will say that a formula $\varphi$ is $\mathbf{AA}$-derivable (denoted $\vdash\varphi$) iff there exists a proof of $\varphi$ from the above axiomatization.
%
%
\end{definition}
\begin{lemma}\label{lemma:axiomatisation:and:translation}
For all words $\alpha$ over $A\cup\Laa$ and for all formulas $\varphi$, $\vdash\lbrack\alpha\rbrack\varphi\leftrightarrow tr(\alpha,\varphi)$.
\end{lemma}
\begin{proof}
The proof is by $\ll$-induction on $(\alpha,\varphi)$.
\end{proof}
\begin{theorem}[Axiomatization AA is sound and complete]\label{proposition:soundness:completeness}  \ \\
For all $\varphi\in\Laa$, $\vdash\varphi$ iff $\models\varphi$.
%
%
%
%
%
%
%
%
%
%
\end{theorem}
\begin{proof}
The soundness of {\bf AA} ($\vdash\varphi$ implies $\models\varphi$) follows from Lemma~\ref{lemma:soundness:of:the:function:tr}, wherein it is shown that the translation $tr$ is truth preserving, and thus that all the axioms are sound.

We now show the completeness ($\models\varphi$ implies $\vdash\varphi$). Suppose $\not\vdash\varphi$.
Let $\psi=tr(\epsilon,\varphi)$.
Since $\not\vdash\varphi$, by Lemma~\ref{lemma:axiomatisation:and:translation}, $\not\vdash\psi$.
Since $\psi$ is a formula in $\Lml$, by the standard completeness of the least normal modal logic in the language $\Lml$, $\not\models^{K}\psi$.
Hence, by Lemma~\ref{lemma.eqeq}, $\not\models\psi$.
Thus, by Lemma~\ref{lemma:soundness:of:the:function:tr}, $\not\models\varphi$.
\end{proof}
Let us remark that, as for $\Lpal$, we now have for $\Laa$ an effective way to determine whether a given $\phi$ is $\epsilon$-valid (for the class of models with arbitrary relations): if $tr(\epsilon,\phi)$ is a theorem in the minimal modal logic {\bf K}, $\phi$ is $\epsilon$-valid; otherwise, $\phi$ is not $\epsilon$-valid. This makes it fairly easy to prove the decidability of $AA$.

	\begin{proposition}
		$AA$ has the finite model property.
	\end{proposition}
	\begin{proof}
		Suppose $\phi$ is satisfiable. Let $M = (W,R,V)$ and state $s \in W$ be such that $s,\epsilon \models \phi$. By Lemma \ref{lemma:axiomatisation:and:translation}, this means that $s,\epsilon\models tr(\epsilon,\phi)$. Since $tr(\epsilon,\phi)\in\mathcal{L}_{ml}$, by Proposition~\ref{lemma.eqeq} (stating that $\models\psi$ iff $\models^K\psi$ for $\psi\in\Lml$), this gives $M,s\models^K tr(\epsilon,\phi)$ in the usual Kripke semantics.  As the minimal modal logic {\bf K} has the finite model property, there exists a finite model $M^f$ and a world $v$ in $M^f$ such that $M^f,v\models^K tr(\epsilon,\phi)$. By the same reasoning, this means that in $M^f$ we have $v,\epsilon\models tr(\epsilon,\phi)$ according to the $AA$ semantics, and thus, again by Lemma \ref{lemma:axiomatisation:and:translation}, $v,\epsilon\models \phi$.
	\end{proof}
	Since $AA$ has a finitary axiomatization and the finite model property we directly obtain decidability.
	\begin{corollary} \label{cor.aadecidable}
		$AA$ is decidable.
	\end{corollary}


\subsection{Reduction axioms for AA$^*$} \label{sec.star}

In this section we determine always-validities ($*$-validities) that have the shape of reduction axioms for announcement. This is instructive, because they resemble the reduction axioms of $PAL$. However, these reductions cannot provide an complete axiomatization as in the previous section. Although we can eliminate the dynamic modalities from $\epsilon$-validities, as formulated in Corollary~\ref{cor.elim}, we cannot eliminate dynamic modalities from $*$-validities.

\begin{proposition}[Failure of elimination of dynamic modalities] \label{prop.elim} \ \\
There are $\phi\in \Laa$ such that for all $\psi\in\Lml$ $\not\wordmodels \phi \eq \psi$.
\end{proposition}
\begin{proof}
Consider the formula $[a]\bot\in\Laa$.
Suppose towards a contradiction that there is a $\psi\in\Lel$ such that $\wordmodels [a]\bot \eq \psi$.
Then in particular we have that $\models [a]\bot \eq \psi$.
As $\models [a]\bot$, it follows that $\models \psi$: $\psi$ is $\epsilon$-valid. As $\psi\in\Lml$, with Proposition~\ref{lemma.eqeq} (stating that $\models\psi$ iff $\wordmodels\psi$) we obtain $\wordmodels \psi$: $\psi$ is $*$-valid.

Now take any $M = (W,R,V)$ and $s \in W$.
From $\models \psi$ then follows $s,\epsilon\models\psi$.
Also, obviously, $s \bowtie \top$ and $s,\top\models\neg[a]\bot$. From that and $\wordmodels [a]\bot \eq \psi$ then follows that $s,\top\models\neg\psi$. However, from $\wordmodels \psi$ we obtain $s,\top\models\psi$. We have the required contradiction.
\end{proof}
We now continue by listing reductions for atoms, conjunction, and negation after announcements $[\phi]$, and some other reductions for formulas occurring after read modalities $[a]$.

\begin{proposition} \label{prop.annk}
Let $\phi,\psi \in \Laa$. Then \
\begin{enumerate}
\item $\wordmodels [\phi] \bot \eq \neg\phi$
\item $\wordmodels [\phi] p \eq (\phi \imp p)$
\item $\wordmodels [\phi] \neg\psi \eq (\phi \imp \neg [\phi] \psi)$
\item $\wordmodels [\phi] (\psi\vee\chi) \eq ([\phi] \psi \vee [\phi] \chi)$
\item $\wordmodels [\phi]B_a\psi \eq (\phi \imp B_a\psi)$
\end{enumerate}
\end{proposition}
The proof is found in the Appendix.

The possibly better known reduction schema $[\phi] (\psi\et\chi) \eq ([\phi] \psi \et [\phi] \chi)$ is also $*$-valid. It can be obtained from $[\phi] \neg\psi \eq (\phi \imp \neg [\phi] \psi)$ and $[\phi] (\psi\vee\chi) \eq ([\phi] \psi \vee [\phi] \chi)$, and some propositional manipulations.

We emphasize that $[\phi]B_a\psi \eq (\phi \imp B_a\psi)$ is invalid for $PAL$. This is obvious, as in $AA$ an agent does not observe the announcement (yet). Dually, the axiom $[\phi]B_a\psi \eq (\phi \imp B_a[\phi]\psi)$ of $PAL$ is invalid for $AA$.
That is equally obvious (take $\varphi=p$ and $\psi=p$), as in $PAL$ the sending and receiving of the announcement are instantaneous (synchronous).

\begin{proposition} \label{prop.ann}
For all $\phi \in \Lpal$ there is a $\phi' \in \Lml$ such that $\wordmodels \phi \eq \phi'$.
\end{proposition}
\begin{proof}
This is proved by truth preserving rewriting, as in $PAL$, where it makes no difference that the reduction for belief after announcement is different from the one in $PAL$: it is still an equivalence, with a lower complexity on the right.

This proof is by (natural) induction on the number of announcements occurring in $\phi$. If $\phi$ contains no announcements, we are done. Otherwise, take an innermost announcement, i.e., a subformula $[\psi]\eta$ of $\phi$ such that $\eta$ does not contain an announcement modality. Then, show that $\wordmodels[\psi]\eta \eq \psi'$ for some $\psi' \in \Lel$ by repeated use of the cases distinguished in Proposition \ref{prop.annk}. Finally, using Proposition~\ref{prop.substeq}, apply induction on the formula $\phi([\psi]\eta/\psi')$, as that contains one less announcement modality than $\phi$.
\end{proof}
By the same rewriting procedure we obtain
\begin{corollary} \label{cor.corrrrr}
For all $\phi,\psi \in \Lpal$, $\wordmodels [\phi]\psi \eq (\phi \imp \psi)$.
\end{corollary}
And the last may be used to prove the following proposition, that may be of interest as it is an axiom in a well-known axiomatization of $PAL$.
\begin{proposition} \label{propseq} For all $\phi,\psi,\eta \in \Lpal$, $\wordmodels [\phi][\psi]\eta \eq [\phi \et [\phi]\psi]\eta$.
\end{proposition}
\begin{proof}
Using the result in Corollary \ref{cor.corrrrr}, observe that $\models^*[\phi][\psi]\eta \eq (\phi \imp [\psi]\eta)$ and $\models^*(\phi \imp [\psi]\eta) \eq (\phi \imp (\psi \imp \eta))$, so that $\models^*[\phi][\psi]\eta \eq (\phi\et\psi \imp \eta)$. Somewhat similarly, $\models^*[\phi \et [\phi]\psi]\eta \eq (\phi \et [\phi]\psi \imp \eta)$ and $\models^*(\phi \et [\phi]\psi \imp \eta) \eq (\phi \et (\phi \imp \psi)\imp \eta)$, i.e., $\models^*(\phi \et [\phi]\psi \imp \eta) \eq (\phi \et \psi \imp \eta)$. Therefore, $\models^*[\phi][\psi]\eta \eq [\phi \et [\phi]\psi]\eta$.
\end{proof}
Clearly, $[\phi][\psi]\eta \eq [\phi \et [\phi]\psi]\eta$ is invalid for asynchronous announcement logic. Merging two announcements into one announcement upsets the order of their reception in formulas bound by such announcements. For a simple counterexample, \begin{itemize} \item $[a]\bot \imp [p][q][a][a]\bot$ is not $*$-valid;
\item whereas $[a]\bot \imp [p \et [p]q][a][a]\bot$ is $*$-valid;
\item therefore, $[p][q][a][a]\bot \eq [p \et [p]q][a][a]\bot$ is not $*$-valid. \end{itemize}
We continue with various $*$-validities involving reception modalities $[a]$. We first show that the order in which agents receive the announcements is irrelevant as long as no announcement is sent in between (similar to the validity $\bigcirc_a\bigcirc_b\phi \eq \bigcirc_b\bigcirc_a\phi$ of \cite{KnightMS19}).

\begin{proposition} \label{proprecep}
$\wordmodels [a] [b] \phi \eq [b] [a] \phi$
\end{proposition}
The proof is found in the Appendix.
\begin{proposition}
Let $\phi\in\Laa$.
\begin{enumerate}
\item $\wordmodels [\phi]\neg [a] \bot$,
\item $\wordmodels B_a [a]\bot$.
\end{enumerate}
\end{proposition}
\begin{proof} Let $(W,R,V)$, $s \in W$, and $\alpha$ such that $s\bowtie\alpha$ be given.
\begin{enumerate}
\item Assume $s,\alpha\models\phi$. From that and $s \bowtie \alpha$ follows $s \bowtie \alpha\phi$.
From $s \bowtie \alpha$ it follows that $|\alpha|_a \leq |\alpha|_!$, so that $|\alpha\phi|_a < |\alpha\phi|_!$. Therefore $s \bowtie \alpha\phi{a}$, so that $s,\alpha\phi\models\dia{a}\top$, in other words, $s,\alpha\phi\models\neg [a]\bot$.
By definition, $s,\alpha\models[\phi]\neg [a]\bot$.
\item Suppose $s,\alpha\not\models B_{a}\lbrack a\rbrack\bot$.
Let $t\in W$ and $\beta$ be a word over $A\cup\Laa$ such that $sR_{a}t$, $\alpha\triangleright_{a}\beta$, $t\bowtie\beta$ and $t,\beta\not\models\lbrack a\rbrack\bot$.
Hence, ${\mid}\beta{\mid}_{a}<{\mid}\beta{\mid}_{!}$.
Since $\alpha\triangleright_{a}\beta$,  ${\mid}\beta{\mid}_{a}={\mid}\beta{\mid}_{!}$: a contradiction.
\end{enumerate} \vspace{-.6cm}
\end{proof}
The result that $B_a [a]\bot$ may puzzle the reader. It formalizes that agents reason about the past and not about the future. Even when more announcements have already been sent than have been received by agent $a$, the beliefs of agent $a$ are only based on the received announcements, not on all announcements. See also Section \ref{sec.korb}.

\begin{proposition} \label{prop.nexttrue}
Let $\phi\in\Laa$. Then $\wordmodels [a] \bot \imp ([a] \phi \eq \top)$.
\end{proposition}
\begin{proof} Let $(W,R,V)$, $s \in W$, and $\alpha$ such that $s\bowtie\alpha$ be given.
$s,\alpha \models [a]\bot$ implies $|\alpha|_a \geq |\alpha|_!$. We now get that $s,\alpha \models [a]\phi$, i.e., $s,\alpha \models [a]\phi \eq \top$.
\end{proof}
The shape $[a] \bot \imp ([a] \phi \eq \top)$ of this validity is to emphasize the difference with the next validity $\neg [a]\bot \imp ([a]\phi\eq\phi)$. Of course we also have that $\wordmodels [a] \bot \imp [a] \phi$.

\begin{proposition} \label{prop.nextcomm}
Let $\phi \in \Laa$ and such that $B_a$ and $[a]$ do not occur in $\phi$. Then $\wordmodels \neg [a]\bot \imp ([a]\phi\eq\phi)$.
\end{proposition}
The proof is found in the Appendix.
We can contrast Proposition \ref{prop.nextcomm} and Proposition \ref{prop.nexttrue} as follows: if agent $a$ has not yet received all announcements ($\neg[a]\bot$ is true), that agent receiving the next announcement does not influence the beliefs of other agents or the truth of any proposition not involving $a$, so we can delete or add the reception modality while preserving truth ($[a]\phi\eq\phi$ is true); whereas if the agent has received all announcements ($[a]\bot$ is true), anything goes after (the impossible event of) receiving the next announcement, i.e., any proposition is true after the reception modality for ($[a]\phi$ is true).

However, if $a$ may occur in $\phi$, then $\not\wordmodels \neg [a]\bot \imp ([a]\phi\eq\phi)$. In particular, neither $\wordmodels \neg [a]\bot \imp ([a]B_a\phi\eq B_a\phi)$ nor $\wordmodels \neg [a]\bot \imp ([a][a]\phi\eq[a]\phi)$.
For a counterexample of the former, consider a model wherein the agent initially does not believe $p$, where $p$ is true in state $s$, and after the announcement that $p$.
Then $s,p \models \neg [a]\bot$ and $s,p \models [a]B_ap$ but $s,p \not\models B_ap$: the agent had to receive the announcement $p$ in order to believe that $p$. For a counterexample of the latter consider the same model. On the one hand, $s,p \models \neg [a]\bot$, i.e., $s,p \not\models [a]\bot$. But on the other hand, $s,p\models [a][a]\bot$. Therefore, $s,p\not\models [a]\bot \eq [a][a]\bot$. For any pair $(s,\alpha)$, whenever $\alpha$ contains one unreceived announcement, then $(s,\alpha)$ satisfies $[a][a]\bot$ but falsifies $[a]\bot$.

\paragraph*{Reduction of belief after receiving announcement?}

We did not find a reduction axiom for belief after receiving announcement, that one could expect to have a shape $[\phi][a]B_a\psi \eq \eta$, where $\eta$ contains a subformula $B_a \eta'$ such that $\eta'$ contains a subformula $[\phi]\eta''$ (or, more specifically, $[\phi][a]\eta''$). Or else, some reduction that might have a slightly more general shape $[\alpha]B_a\psi \eq \eta$, where $\alpha$ is a history containing $a$ after $\phi$ and possibly satisfying even further constraints. The special case of axiom $(A7)$ of $AA^\epsilon$ for $\alpha = \phi {a}$, that in the single-agent case has form $[\phi][a]B_a\psi \eq (\phi \imp B_a [\phi][a]\psi)$ (see also the next subsection), is clearly not valid in $AA^*$, as the $a$ in question may read another announcement than $\phi$, namely the next unread announcement in a history $\alpha$ of a pair $(s,\alpha)$ in which we interpret that form. We also have not proved that no reduction exists for belief after receiving announcement. 

\begin{example}
Consider the model $p\nq{r}(s)$---$a$---$pq\overline{r}(t)$. We will now evaluate a formula of shape $\dia{\phi}\dia{a}\M_a\psi$ (the dual of shape $[\phi][a]B_a\psi$) in state $s$ of this model, however, not for the empty history but assuming history $pa$. Let $\phi$ be $\dia{a}\top\vel r$ and let $\psi$ be $q$. Then $s,pa \models \dia{\phi}\dia{a}\M_a q$. Indeed $s,pa \models \dia{a}\top\vel r$ (because $s,pa \models r$), and also $s, pa\phi{a} \models \M_a q$, because $R_ast$, $pa\phi{a} \tri_a p\phi{a}a$, $t \bowtie p\phi{a}a$ (because $t,p \models \dia{a}\top$), and $t, p\phi{a}a \models q$.

In order to interpret $\M_a$ in $\dia{\phi}\dia{a}\M_a\psi$, we need to swap elements of the history $\alpha$ in which we interpret this formula, with elements of the history $\phi{a}$ preceding $\M_a$ in the formula. It is unclear how to formalize this interaction in general.
\end{example}

The reductions for $AA^*$ given in this section do not constitute a system to rewrite formulas into some standard, simpler, form. We recall that not all dynamic modalities can be eliminated from the language in a $*$-valid equivalent way, as $[a]\bot$ is irreducible. It would be of interest to determine whether all public announcement modalities can be eliminated, i.e., whether for each $*$-validity there is an equivalent formula without public announcement modalities.

\subsection{Public announcements versus asynchronous announcements} \label{sec.palaa}

Table \ref{table.palaa} compares the axioms of $PAL$, and the axioms of the two asynchronous announcement logics $AA^\epsilon$ and $AA^*$. Additional reductions only involving the interaction of read modalities $[a]$ have not been taken into account, as they are irrelevant for $PAL$.

To see the correspondence with the axioms of {\bf AA} in Proposition \ref{prop.annk}, replace the arbitrary history in those formulations by the appropriate announcement. For example, in $(A1) \ [\alpha]p \eq ([\alpha]\bot \vel p)$ we take $\alpha=\phi$ which results in $[\phi]p \eq ([\phi]\bot \vel p)$, in other words, $[\phi]p \eq (\neg\phi \vel p)$, i.e., $[\phi]p \eq (\phi \imp p)$. For another example, in $(A4) \ [\alpha\phi]\bot \eq [\alpha]\bot \vel \neg [\alpha]\phi$ we take $\alpha = \epsilon$ which results in $[\phi]\bot \eq [\epsilon]\bot \vel \neg [\epsilon]\phi$, i.e., $[\phi]\bot \eq \neg \phi$.

In particular it should be noted that if in \[ (A7) \ \lbrack\alpha\rbrack B_{a}\varphi\leftrightarrow\lbrack\alpha\rbrack\bot\vee\bigwedge\{B_{a}\lbrack\beta\rbrack\varphi \mid \alpha\triangleright_{a}\beta\} \] we (simultaneously) take $\phi=\psi$ and $\alpha =\phi{a}$ then we get \[ {[}\phi{a}]B_a \psi \eq [\phi{a}]\bot \vel \Et \{ B_a [\beta]\psi \mid \phi{a} \tri_a \beta\}, \] i.e., also using that $\wordmodels [ab]\psi \eq [ba]\psi$ (Proposition \ref{proprecep}), \[ {[}\phi{a}]B_a \psi \eq [\phi{a}]\bot \vel \Et \{ B_a [\phi{a}{B}]\psi \mid B \subseteq A\setminus{a} \}, \] and therefore in the single-agent case, also using that $\wordmodels\neg[\phi{a}]\bot \eq \phi$,
\[ {[}\phi][a]B_a \psi \eq (\phi \imp B_a [\phi][a]\psi). \]

\begin{table}[h]
{\small
\[ \begin{array}{l|l|l|l}
\text{formula} & PAL & AA^\epsilon & AA^* \\
\hline
{[}\phi]\bot \eq \neg\phi & \checkmark & \checkmark  \text{ Def. } \ref{definition:axiomatisation:Laa}.A4 & \checkmark \text{ Prop. } \ref{prop.annk}.1. \\
{[}\phi]p \eq (\phi \imp p) & \checkmark &\checkmark \text{ Def. } \ref{definition:axiomatisation:Laa}.A1 &  \checkmark \text{ Prop. } \ref{prop.annk}.2. \\
{[}\phi](\psi\vel\eta) \eq ([\phi]\psi \vel [\phi]\eta) &\checkmark & \checkmark \text{ Def. } \ref{definition:axiomatisation:Laa}.A6 & \checkmark \text{ Prop. } \ref{prop.annk}.4. \\
{[}\phi]\neg\psi \eq (\phi \imp \neg [\phi]\psi) &\checkmark & \checkmark \text{ Def. } \ref{definition:axiomatisation:Laa}.A5 & \checkmark \text{ Prop. } \ref{prop.annk}.3. \\
{[}\phi][\psi]\eta \eq [\phi \et [\phi]\psi]\eta &\checkmark & \times & \times  \ (\checkmark \text{ in } \Lpal, \text{Prop. } \ref{propseq}) \\
{[}\phi]B_a \psi \eq (\phi \imp B_a [\phi]\psi) &\checkmark & \times & \times \\
{[}\phi]B_a \psi \eq (\phi \imp B_a \psi) & \times & \checkmark & \checkmark \text{ Prop. } \ref{prop.annk}.5. \\
{[}\phi][a]B_a \psi \eq (\phi \imp B_a [\phi][a]\psi)  \ \ \ \text{(1-agent)} & \text{N/A} & \checkmark \text{ Def. } \ref{definition:axiomatisation:Laa}.A7 & \times \\
{[}\phi][a]B_a \psi \eq (\phi \imp \Et_{B\subseteq A\setminus{a}}B_a [\phi][a][B]\psi) & \text{N/A} & \checkmark \text{ Def. } \ref{definition:axiomatisation:Laa}.A7 & \times \\
\end{array} \]
}
\caption{Comparing public announcement logic and asynchronous announcement logic}
\label{table.palaa}
\end{table}

\section{Comparison to other semantics} \label{sec.comparison}

In this section we give results for the class $\mathcal{S}5$ of models where all accessibility relations are equivalence relations, we consider an alternative for the `view'-relation resulting in asynchronous \emph{knowledge} instead of asynchronous belief, we motivate the belief semantics by a detailed example involving belief as acknowledgement, we relate the $AA$ semantics to history-based semantics, we present the results of the related asynchronous broadcast logic, and we compare our histories containing announcements and receptions to the cuts of distributed computing.

\subsection{Asynchronous announcement logic on the class S5} \label{sec.s5}

In this section we restrict the models $(W,R,V)$ to those where all accessibility relations $R_a$ are equivalence relations. Such models are known as {$\mathcal{S}5$ models}, and in that case $B_a \phi$ stands for `the agent knows $\phi$', and, in standard Kripke semantics $\models^K$, the operator then satisfies the so-called {\em properties of knowledge} $B_a \phi \imp \phi$ (T, factivity), $B_a \phi \imp B_a B_a \phi$ (4, positive introspection), and $\neg B_a \phi \imp B_a \neg B_a \phi$ (5, negative introspection).
These properties correspond to, respectively, the facts that the accessibility relation $R_a$ is reflexive, transitive and Euclidean.

The \emph{properties of belief} (also known as \emph{introspective belief}) are as the properties of knowledge, except that $B_a \phi \imp \phi$ is replaced by $B_a \phi \imp \M_a \phi$ (D, consistency), which corresponds to seriality of underlying frames.
The models with serial, transitive and Euclidean relations are known as {$\mathcal{KD}45$ models}.

We very straightforwardly have that $B_a$ satisfies the properties of knowledge:
\begin{proposition} \label{prop.s5epsilon}
Let $\phi\in\Laa$. Then
\begin{itemize}
\item $\mathcal{S}5\models B_a \phi \imp \phi$
\item $\mathcal{S}5\models B_a \phi \imp B_a B_a \phi$
\item $\mathcal{S}5\models \neg B_a \phi \imp B_a \neg B_a \phi$
\end{itemize} \vspace{-.4cm}
\end{proposition}
\begin{proof}
Let $(W,R,V)$ and $s \in W$ be given. Then $s,\epsilon \models B_a \phi$ iff $t,\beta \models \phi$ for all $t,\beta$ such that $sR_at$, $\epsilon\tri_a \beta$, and $t \bowtie \beta$. As $\mathsf{view}_a(\epsilon) = \{\epsilon\}$, and $t \bowtie \epsilon$ holds by definition, we get that: $s,\epsilon \models B_a \phi$ iff $t,\epsilon \models \phi$ for all $t$ such that $sR_at$. As $R_a$ is an equivalence relation, $B_a$ therefore satisfies the three properties of knowledge.
\end{proof}

In asynchronous announcement logic interpreted on $\mathcal{S}5$ models with history, the $B_a$ operator does not satisfy all the properties of knowledge (and therefore we write $B_a$ and not $K_a$ for this modality). For example, if agents $a,b$ are initially both uncertain about $p$ and this is common knowledge, as in the $\mathcal{S}5$ model $p$---$ab$---$\overline{p}$ (where the names of the states reflect the value of $p$ there), and the announcement of $p$ is made and received by $a$ but not yet by $b$, then $p,pa \models B_a p \et B_b \neg B_a p$: the beliefs of agent $b$ are incorrect. In general, whenever $|\alpha|_! > |\alpha|_a$, then agent $a$ has not yet received all announcements and may therefore hold incorrect beliefs. If $|\alpha|_! > |\alpha|_a$ then it is not the case that $\alpha \tri_a \alpha$: the view relation $\tri_a$ is not reflexive.

However, all the other properties of introspective belief hold for asynchronous announcement logic interpreted on $\mathcal{S}5$ models.

\begin{proposition}
Let $\phi\in\Laa$. Then:
\begin{itemize}
\item $\mathcal{S}5\wordmodels B_a \phi \imp \neg B_a \neg \phi$
\item $\mathcal{S}5\wordmodels B_a \phi \imp B_a B_a \phi$
\item $\mathcal{S}5\wordmodels \neg B_a \phi \imp B_a \neg B_a \phi$
\end{itemize}
\end{proposition}
\begin{proof}
We recall that $s,\alpha \models B_a \phi$ iff $t,\beta \models \phi$ for all $t,\beta$ such that $sR_at$, $\alpha \tri_a \beta$, and $t \bowtie \beta$, and that the view relation is defined as $\alpha\tri_a\beta$ iff $|\beta|_a = |\alpha|_a$, $\beta\proj_{!a} = \alpha\proj_{!a}$, and $|\beta|_! = |\alpha|_a$.

We show that the relation ${\mathbf R}_a$ defined on the set of pairs $(s,\alpha)$ with $s\bowtie\alpha$ as follows: \[ (s,\alpha)\mathbf{R}_a(t,\beta) \text{ \ \ iff \ \ } sR_at, \alpha \tri_a \beta, \text{ and } t \bowtie \beta \] is introspective (i.e., transitive, Euclidean, and serial), so that $B_a$ satisfies the three properties of belief. In the proof we will use that, as $R_a$ is an equivalence relation and $\tri_a$ is transitive and Euclidean, their product is also transitive and Euclidean.

Transitivity of ${\mathbf R}_a$ follows from the transitivity of $R_a$ and $\tri_a$ (the parts involving $\bowtie$ merely result in a restriction).
\[ \begin{array}{llll}
\text{if} & sR_at, & \alpha \tri_a \beta, & t \bowtie \beta, \\
\text{and} & tR_au, & \beta \tri_a \gamma, & u \bowtie \gamma, \\
\text{then} & sR_au, & \alpha \tri_a \gamma, & u \bowtie \gamma
\end{array} \]

Since $R_a$ and $\tri_a$ are Euclidean, ${\mathbf R}_a$ is Euclidean.
\[ \begin{array}{llll}
\text{if} & sR_at, & \alpha \tri_a \beta, & t \bowtie \beta, \\
\text{and} & sR_au, & \alpha \tri_a \gamma, & u \bowtie \gamma, \\
\text{then} & tR_au, & \beta \tri_a \gamma, & u \bowtie \gamma
\end{array} \]

Seriality of ${\mathbf R}_a$ follows from the reflexivity of $R_a$ and the seriality of $\tri_a$. For the latter, Proposition \ref{lemma.tri}.\ref{one} showed that for any history $\alpha$, $\alpha \tri_a \alpha_a$. The proof of Lemma \ref{lemma.fiets}.\ref{fiets.two} that $\alpha_a \preceq \alpha$ demonstrated that for any state $s$ with $s \bowtie \alpha$ we also have $s \bowtie \alpha_a$. From $sR_as$, $\alpha \tri_a \alpha_a$, and $s \bowtie \alpha_a$ we get that $(s,\alpha){\mathbf R}_a(s,\alpha_a)$.
\end{proof}


Let us now restrict the formulas $\phi$ from the language $\Laa$ to the language $\Lml$, just as in Section \ref{sec.ml}. We there obtained in Lemma \ref{lemma.eqeq} that $\models^K \phi$ iff $\models \phi$, and that $\models \phi$ iff $\wordmodels \phi$. We are now in for a small surprise. Although we still have that \begin{corollary} $\mathcal{S}5\models^K \phi$ iff $\mathcal{S}5\models \phi$, \end{corollary} we do not have in general that $\mathcal{S}5\models \phi$ iff $\mathcal{S}5\wordmodels \phi$. As a simple corollary of Proposition~\ref{prop.starepsilon} we still have that $\mathcal{S}5\models^* \phi$ implies $\mathcal{S}5\models \phi$. However, recalling the proof of Lemma \ref{lemma.eqeq}, we now do not have in general that $\mathcal{S}5\models^K \phi$ implies $\mathcal{S}5\wordmodels \phi$ (which by analogy we would need to prove that $\mathcal{S}5\models \phi$ iff $\mathcal{S}5\wordmodels \phi$). As $\mathcal{S}5\not\wordmodels B_a \phi \imp \phi$, $*$-satisfiability in $\mathcal{S}5$ does not entail Kripke satisfiability in $\mathcal{S}5$. The typical counterexample is the one at the beginning of this section: $p,pa \models B_a p \et B_b \neg B_a p$. But $B_a p \et B_b \neg B_a p$ is not satisfiable in $\mathcal{S}5$. We note that, If $M$ is a $\mathcal{S}5$ model and $s$ a state in $M$, then the flat model $\overline{M^{s\alpha}}$ need not be a $\mathcal{S}5$ model.

Despite this disappointment, in the $\mathcal{S}5 \wordmodels$ semantics some beliefs are, after all, correct, and thus knowledge. This may also come as a surprise, but in this case a pleasant one.

\begin{proposition}[Positive beliefs are correct] \label{prop.positive}
Let $\phi \in \Lml^+$. Then $\mathcal{S}5\models^* B_a \phi \imp \phi$.
\end{proposition}
\begin{proof}
Let $s$ and $\alpha$ be a world and a history such that $s \bowtie \alpha$ and $s,\alpha \models B_a \phi$. From that and Lemma \ref{1234} it follows that $s,\alpha_a \models B_a \phi$. From $s \bowtie \alpha$ and Lemma \ref{lemma.fiets}.\ref{fiets.two} it follows $s \bowtie \alpha_a$. Then, from $s,\alpha_a \models B_a \phi$, $s R_a s$, $\alpha_a \tri_a \alpha_a$, and $s \bowtie \alpha_a$ it follows that $s,\alpha_a \models \phi$.
As $\phi \in \Lml^+$ and $\alpha_a \preceq \alpha$, from Lemma \ref{lem:alphaleqbeta} it follows that $s,\alpha \models \phi$.
\end{proof}

A different way of seeing this result is that eventually all beliefs become correct and therefore knowledge, because eventually all messages will be received (and, as we know, all messages were truthful when sent) and eventually all uncertainty may be resolved. In other words, initially or at some intermediate stage an agent may well incorrectly believe that another agent is ignorant, namely when the other agent has already received more announcements, but eventually the first agent will also receive those messages and then change his incorrect beliefs into correct and stable beliefs: knowledge of positive formulas. We consider this an important observation.

\medskip

Let now {\bf AA}$_{\mathbf{S5}}$ be the axiomatization formed by extending the axiomatization {\bf AA} of $AA$ with the {\bf S5} axioms {\bf T}, {\bf 4}, and {\bf 5}.
Recalling the soundness and completeness of {\bf AA} (Theorem~\ref{proposition:soundness:completeness}), in view of Proposition \ref{prop.s5epsilon} we immediately obtain:

\begin{corollary}[Axiomatization {\bf AA}$_{\mathbf{S5}}$ is sound and complete]\label{cor.s5ax} \ \\
For all $\varphi\in\Laa$, {\bf AA}$_{\mathbf{S5}}\vdash\varphi$ iff $\mathcal{S}5\models\varphi$.
\end{corollary}

We conclude this section with yet another observation on the relation between knowledge and belief. Although $\mathcal{S}5 \wordmodels$ satisfies the properties of belief, $\mathcal{KD}45 \wordmodels$ does {\bf not} satisfy the properties of belief. For a simple counterexample, consider the single-agent two-state $\mathcal{KD}45$ model with $R_a = \{(s,t),(t,t)\}$ and where $p$ is only true in $s$, visualized as $p\xrightarrow{a}\overline{p}$. Then $s, \epsilon \models B_a \neg p$. After the truthful announcement that $p$ and $a$ receiving it, the beliefs of agent $a$ are inconsistent, so that $s,pa \not \models B_a p \imp \M_a p$. This is a well-known problem of $\mathcal{KD}45$ updates in $\mathcal{KD}45$ models \cite{balbianietal:2012}.

\subsection{Knowledge or belief?} \label{sec.korb}

We recall the definition of the view relation as \begin{quote} $\alpha\triangleright_a\beta$ iff ${\alpha}\proj_{!a} = {\beta}\proj_{!a} = {\beta}\proj_!$.\end{quote} The restriction ${\beta}\proj_{!a} = {\beta}\proj_!$ rules out that the agent considers other agents having received more announcements than herself. If we remove that constraint, we get \begin{quote} $\alpha\equiv_a\beta$ iff ${\alpha}\proj_{!a} = {\beta}\proj_{!a}$.\end{quote} The relation $\equiv_a$ is an equivalence relation.

The interpretation of $B_a$ is defined as $s,\alpha\models B_a\varphi$ iff $t,\beta\models\varphi$ for all $t,\beta$ such that $R_ast$, $\alpha\triangleright_a\beta$, $t\bowtie\beta$. If $R_a$ is an equivalence relation (the $\mathcal{S}5$ models), and if we replace $\alpha\triangleright_a\beta$ by $\alpha\equiv_a\beta$, it is no longer clear that the agreement relation $\bowtie$ is well-founded, for example one such $\beta$ would be $\alpha(B_a\phi)$, as $\alpha \equiv_a \alpha(B_a \phi)$. More precisely, in order to determine whether  $s,\alpha\models B_a\varphi$, given that $R_ass$ and $\alpha \equiv_a \alpha(B_a \phi)$, we have to determine whether $s \bowtie \alpha(B_a\varphi)$, for which we have to determine whether $s,\alpha\models B_a\varphi$: a vicious circle. Or at least vicious on first sight, without alternative modelling solutions such as fixpoints.

We may need a novel way to give a semantics to the epistemic modality. However, any such modality will clearly be interpreted by an {\em equivalence} relation. Instead of $B_a$ having the properties of belief, it would then have the properties of \emph{knowledge}; and one might as well write $K_a$ for it, as we will do from here on. In the temporal epistemic logics for interpreted systems the epistemic modality is indeed such a knowledge modality, and the view relation in such works always is an equivalence relation \cite{halpernmoses:1990,ram:1999,MukundS97}. This is also the approach followed in \cite{KnightMS19}.

\medskip

Given the history of asychronous knowledge in distributed computing, one would by all means preferably have such a notion of knowledge also in a dynamic epistemic logic. This we cannot offer at this stage. Clearly, the generalization of our framework to epistemic notions that are closer to the common epistemic notions in distributed computing are obligatory further research, and we hope that our readers and those of \cite{ KnightMS19} will be encouraged to develop such logics. However, we do not think that therefore the belief semantics is somehow a second choice. Both the knowledge and belief semantics have their advantages, and ideally one would have a logic wherein both knowledge and belief appear, and that can be tailored according to the need of the modeller. In the remainder of this subsection, let us more precisely focus on the differences between asynchronous knowledge and asynchronous belief, and on possible modelling advantages of asynchronous belief.

\paragraph*{Knowledge of novel propositions}
In dynamic epistemic logics, the messages sent do not contain novel relevant propositions but are {\em updates} on the uncertainty about the currently relevant propositions, that are a given and that have a fixed unchangeable value. The goal of such sequences of updates is to finally determine their value, and the interesting phenomena are those wherein some agents reveal their uncertainty about the beliefs of other agents and thus acquire hard information about such facts.

If facts also change value, for example if messages sent and received are recorded by making fresh variables (atoms) true, even knowledge of atoms can change and $K_a p$ may be true now but $K_a \neg p$ may be true later. This is the common scenario in distributed computing.

\paragraph*{Belief in positive formulas is correct}
As shown in Proposition~\ref{9012}, beliefs in positive formulas are stable. And, as shown in Proposition~\ref{prop.positive}, for the class $\mathcal{S}5$ of initial models, such beliefs are correct, and thus knowledge. As explained there, this can be interpreted as all belief eventually becoming knowledge.

\paragraph*{Decidability}
The knowledge semantics reasons over all possible future updates of the current model, and therefore over all possible model restrictions. In other words, it quantifies over all announcements. Arbitrary public announcement logic is a logic with a modality for quantifying over announcements and this logic is known to be undecidable \cite{frenchetal:2008}. One might therefore expect a logic of announcements with asynchronous knowledge to be also undecidable (but we emphasize that we do not know this). However, the logic $AA$ with the belief semantics is decidable (Corollary~\ref{cor.aadecidable}). 

\paragraph*{Should knowledge of ignorance be unsatisfiable?}
We now continue to explore somewhat informally the above knowledge semantics with $\equiv_a$. In this semantics, it seems that an agent can never know that another agent is ignorant.

Let us first see why $K_a \neg (K_b p \vel K_b \neg p)$ is unsatisfiable for an atom $p$. Let some model $M$ be given, as well as a state $s$ and a history $\alpha$ that is executable in $s$ (these assumptions of course remain somewhat informal, we would like to say that we assume a pair $(s,\alpha)$ such that $s \bowtie \alpha$). Atom $p$ is necessarily either true or false in $s$, and therefore in $(s,\alpha)$, such that either $p$ or $\neg p$ can be announced, and following this announcement $b$ may have received it but not (yet) $a$. In the first case, $b$ no longer considers any state possible wherein $p$ is false: $s,\alpha{p}{b}\not\models \hat{K}_b \neg p$, because there is no $t$ with $s R_b t$ and (for some unspecified $\beta$) $t, \beta \models \neg p$. In the second case, similarly, $b$ no longer considers any state possible wherein $p$ is true: $s,\alpha{\neg p}{b}\not\models \hat{K}_b \neg p$. (This is of course merely an intuitive argument explaining that $[pb]K_b p$ and $[\neg pb]K_b \neg p$ are also valid for the knowledge semantics.) From  $s,\alpha{p}{b}\not\models \hat{K}_b p$ or $s,\alpha{\neg p}{b}\not\models \hat{K}_b \neg p$ and the semantics of knowledge we conclude that $s,\alpha\not\models K_a (\hat{K}_b p \et \hat{K}_b \neg p)$, i.e., that $s,\alpha\not\models K_a \neg (K_b p \vel K_b \neg p)$. As $M$, $s$ and $\alpha$ were arbitrary, it follows that $K_a \neg (K_b p \vel K_b \neg p)$ is unsatisfiable.

Similarly, this argument holds for any Boolean formula instead of an atom: for any Boolean formula $\phi$, $K_a \neg (K_b \phi \vel K_b \neg \phi)$ is unsatisfiable.

We conjecture that it is also impossible to know that other agents are ignorant for arbitrary formulas $\phi$, but this is even harder to make precise given that we have only informally considered the knowledge semantics. Given a finite model, an {\em a priori} argument is that we can always announce the characteristic formula of the current state and have this announcement be received by agent $b$, after which any formula $\phi$ is either true or false, and, as we conjecture, even known by or knowable to $b$. From $K_b \phi \vel K_b \neg\phi$ we then obtain $\hat{K}_a (K_b \phi \vel K_b \neg\phi)$, negating the above.

Now consider the belief semantics. Here, it is obvious that formulas of shape $B_a \neg (B_b \phi \vel B_b \neg \phi)$ are satisfiable --- and such beliefs may even be correct. For a very basic example, consider an initial model consisting of a $p$-state $s$ and a $\neg p$-state $t$ that are indistinguishable for two agents $a,b$. Obviously, $s,\epsilon \models B_a \neg (B_b p \vel B_b \neg p)$.  Of course, also $s,pb \models B_a \neg (B_b p \vel B_b \neg p)$ even though $s,pb \models B_b p$. This is trivial. Let us proceed with the non-trivial: belief as acknowledgement.

\paragraph*{Belief as acknowledgement}
Continuing the analysis of this basic example, it is however non-trivial that $a$ may signal to $b$ that she has not yet received novel information, by announcing $B_a \neg (B_b p \vel B_b \neg p)$ --- we recall that a truthful announcement by $a$ of $\phi$ to $b$ is a truthful public announcement of $B_a \phi$. Then, e.g., history $pb(B_a \neg (B_b p \vel B_b \neg p))b$ reveals to $b$ that he received the first announcement $p$ before $a$. Such announcements $B_a \neg (B_b p \vel B_b \neg p)$ are more \emph{acknowledgements} by $a$ than \emph{beliefs} of $a$. We can play this acknowledgement game to the full in the subsequent asynchronous analysis of the muddy children puzzle, wherein agents gain factual knowledge by \emph{acknowledging} the ignorance of others, as usual. Such an analysis is evidently impossible in the $\equiv_a$ semantics wherein knowledge of ignorance of others is unsatisfiable.

\subsection{Solving muddy children with asynchronous announcements} \label{sec.muddy}

Consider the Muddy Children problem \cite{mosesetal:1986,hvdetal.puzzle:2015} for the case of three children $a,b,c$ that are all muddy. We present the story as one about knowledge $K_a$, for sake of the exposition, but we always assume the $AA$ semantics of belief $B_a$ (possibly mistaken beliefs play no role in the analysis). Father first announces that at least one of the children is muddy. This is a public announcement of $m_a \vel m_b \vel m_b$. Father then repeatedly requests all children who know whether they are muddy to step forward. After the first two requests, no child steps forward, which corresponds to the public announcement of formula $\neg (\Kw_a m_a \vel \Kw_b m_b \vel \Kw_c m_c)$ --- wherein we abbreviate $K_i \phi \vel K_i \neg \phi$ as $\Kw_i \phi$, for ```agent $i$ knows whether $\phi$.'' This formalizes the statement ``no child knows whether it is muddy.'' Then, at the third request, all three children step forward. The second time they do not step forward is a typical example of an {\em unsuccessful update} (a formula that becomes false after being announced), because the final action can be seen as the public announcement of $\Kw_a m_a \vel \Kw_b m_b \vel \Kw_c m_c$. Note that in the given model this (inclusive) disjunction has the same informative effect as the conjunction of the three knowing-whether terms.

This standard solution no longer works with asynchronous announcements, for different reasons.
In the first place, the action of no child stepping forward can no longer be represented by a single formula, as this \emph{public} announcement formula is implicitly a synchronization of the children's individual decisions not to step forward. In the second place, if after receiving father's announcement that $m_a \vel m_b \vel m_c$ a child $i$ merely  announces $\neg \Kw_i m_i$, it is unclear to the other children receiving both announcements in that order that child $i$ had received father's announcement when it made its own announcement. Receiving this information is then no longer informative. The other children cannot rule out that child $i$ had announced $\neg \Kw_i m_i$ before receiving father's announcement (but still after father's announcement was made), in which case it merely describes a commonly known property of the initial model. As the other children cannot distinguish this uninformative history from the informative history, they cannot draw an informative conclusion.

Our solution emphasizes the use of our epistemic belief/knowledge notion as one of acknowledgement. Following father's initial announcement of $m_a \vel m_b \vel m_c$, let each child $i$ announce \[ \neg \Kw_i m_i \et K_i (m_a \vel m_b \vel m_c) \] This means that $i$ still does not know whether she is muddy after having received father's announcement that at least one child is muddy. She announces her ignorance while \emph{acknowledging the reception} of father's first announcement to the other children. Let all three children do this.

After receiving all these, each child $i$ now announces her continued ignorance while acknowledging reception of these three ignorance announcements including her own:
\[ \begin{array}{lll} \neg \Kw_i m_i & \et & K_i (\neg \Kw_a \et K_a (m_a \vel m_b \vel m_c) \et \\ && K_i (\neg \Kw_b \et K_b (m_a \vel m_b \vel m_c) \et \\ && K_i (\neg \Kw_c \et K_c (m_a \vel m_b \vel m_c) \end{array} \]
This formalizes that even after having received the information that no child knows whether it is muddy, $i$ still does not know whether she is muddy.

After all three children have sent and subsequently received this information, they will all know that they are muddy. (And they may finally all three step forward, although slightly out of step as they do this asynchronously.) Problem solved!

It is straightforward to generalize this to any number of $n \geq 3$ children of which $k \leq n$ are muddy by further iterations of acknowledgement of ignorance in the previous round.

In the knowledge semantics with $\equiv_i$ instead of $\tri_i$, announcement $\neg \Kw_i m_i \et K_i (m_a \vel m_b \vel m_c)$ can still be made and received, but, crucially, not the other more complex announcements acknowledging ignorance (or subsequent iterations). We recall that knowledge of other agents' ignorance of Booleans (such as $m_i$ and $\neg m_i$) is not satisfiable in that semantics.

\subsection{History-based structures} \label{sec.history}

In dynamic epistemic logic, the dynamic modalities induce model transformations and are then interpreted as relations between pointed models. They are not interpreted as relations between states in a given model. Various frameworks are known to enforce the interpretation of dynamic modalities by way of relations in a given model. This is typically in the setting of translating dynamic epistemic logics to temporal epistemic logics, in which case the dynamic transformations correspond to steps of a temporal next-time operator. If we wish to interpret dynamic modalities with an accessibility relation in some model, that model is then much larger than the `initial' model used for model transformations. For example, if the dynamic modalities are for public announcements, the `supermodel' should somehow contain all modally definable submodels of that initial model.
To our knowledge, such constructions have first been proposed by Venema in~\cite{jfaketal:2001}. Well-known are the {\em protocol generated forests} of \cite{jfaketal.JPL:2009}, see also \cite{hvdetal.FAMAS:2013}, and \cite{degremontetal:2011} for an asynchronous interpretation. A maybe less well-known, but equally elegant approach are the {\em extended models} of \cite{WC13}. The {\em asynchronous pre-models} of \cite{KnightMS19} are also history-based structures.

In this section we define a history-based structure called {\em asynchronous model} that can be defined given a model and history, and show the obvious required correspondence. 

\begin{definition}[Asynchronous model]
Let model $M = (W,R,V)$ and history $\alpha$ be given. The {\em asynchronous model} $M^{\alpha\proj_!} = (W',R',V')$ is defined as:
\[ \begin{array}{lll}
W' & = & \{ (s,\beta) \mid s \in W, \beta \text{ a history such that } \beta\proj_! \subseteq \alpha\proj_!, \text{and}\ s \bowtie \beta \} \\
(s,\beta) R'_a (t,\gamma)  & \text{iff} & sR_at \ \text{and}\ \beta=\gamma \\
(s,\beta) R'_\phi (t,\gamma) & \text{iff} & s=t \ \text{and}\ \beta\phi=\gamma \\
(s,\beta) R'_{[a]} (t,\gamma) & \text{iff} & s=t \ \text{and}\ \beta{a}=\gamma \\
(s,\beta) R'_{\tri_a} (t,\gamma) & \text{iff} & s=t \ \text{and}\ \beta \tri_a \gamma \\
V'(p) & = & \{ (s,\beta) \in W' \mid s \in V(p) \}
\end{array} \]
\end{definition}
A member of the domain of an asynchronous model is an \emph{asynchronous state}.
\begin{definition}
Let model $M = (W,R,V)$ and history $\alpha$ be given. The interpretation of formulas $\phi \in \Laa$ on asynchronous model $M^{\alpha\proj_!} = (W',R',V')$ is defined by induction. We omit all the obvious clauses.
\[\begin{array}{lll}
M^{\alpha\proj_!}, s,\beta \models [\phi]\psi & \text{iff} & \text{if} \ \beta\phi\proj_! \subseteq \alpha\proj_! \ \text{then} \\ && M^{\alpha\proj_!}, t,\gamma \models \psi \text{ for all } (t,\gamma) \text{ such that } (s,\beta) R'_\phi (t,\gamma), \\
 & &  \text{else} \\ && M^{\beta\phi\proj_!}, s,\beta \models [\phi]\psi \\
M^{\alpha\proj_!}, s,\beta \models B_a \psi & \text{iff} & M^{\alpha\proj_!}, t,\gamma \models \psi \text{ for all } (t,\gamma) \ \text{such that}\ (s, \beta) R'_a (t,\gamma) \ \text{and}\ (s, \beta) R'_{\tri_a} (t,\gamma) \\
M^{\alpha\proj_!}, s,\beta \models [a]\psi & \text{iff} & M^{\alpha\proj_!}, t,\gamma \models \psi \text{ for all } (t,\gamma) \ \text{such that}\ (s,\beta) R'_{[a]} (t,\gamma)
\end{array} \]
\end{definition}
As usual, in the clause for $[\phi]\psi$, it need not be the case that $M^{\alpha\proj_!}, s,\alpha\models \phi$. Therefore, we do not assume that $s \bowtie \beta\phi$. The two conditions of that clause are for the case where $\phi$ is the next announcement in the asynchronous model for $\alpha$, and where $\phi$ is `novel', so to speak, in which case we have to construct another asynchronous model incorporating announcement $\phi$, in order to proceed.

\begin{proposition} \label{prop.historiesforever}
Let $\phi\in\Laa$ be given. For all $M$, and $s$ and $\alpha$ with $s \bowtie \alpha$, $s, \alpha \models \phi$ iff $M^{\alpha\proj_!}, s,\alpha \models \phi$.
\end{proposition}
The proof is found in the Appendix.

\begin{example}
We can now finally validate the depictions used for the introductory example involving Anne who knows about $p$ and Bill who knows about $q$, and the asynchronous announcement of $p \vel q$ followed by first Anne and then Bill receiving it.

Figure \ref{fig.history} depicts the asynchronous history model for this sequence of events. The relations $\tri_a$ and $\tri_b$ have only been visualized for some typical cases. Instead of naming the states we show their valuation of $p$ and $q$, as common in depictions of multi-agent models. For example, if in the initial model (i) the topleft state $\np{q}$ is called $s$ and the topright state $pq$ is called $t$, then these correspond to the asynchronous states $(s,\epsilon)$ and $(t,\epsilon)$ respectively, and the topright state $pq$ in the figure would then be $(t,(p\vel{q})ab)$, and so on.

The structures $i,ii,iii,iv$ of Figure \ref{fig} in the introduction should be seen as the, respectively $\epsilon$, $p\vel q$, $(p\vel q)a$, and $(p \vel q)ab$ projections of the asynchronous model.

\begin{figure}[h]
\begin{tikzpicture}
\node (00t) at (2,0) {$\np\nq$};
\node (01t) at (2,2) {$\np\q$};
\node (10t) at (3.5,.7) {$\p\nq$};
\node (11t) at (3.5,2.7) {$\p\q$};
\draw (00t) -- node[fill=white,inner sep=1pt] {\footnotesize $b$} (10t);
\draw (01t) -- node[fill=white,inner sep=1pt] {\footnotesize $b$} (11t);
\draw (00t) -- node[fill=white,inner sep=1pt] {\footnotesize $a$} (01t);
\draw (10t) -- node[fill=white,inner sep=1pt,near start] {\footnotesize $a$} (11t);
\node (01tt) at (5,2) {$\np\q$};
\node (10tt) at (6.5,.7) {$\p\nq$};
\node (11tt) at (6.5,2.7) {$\p\q$};
\draw (01tt) -- node[fill=white,inner sep=1pt] {\footnotesize $b$} (11tt);
\draw (10tt) -- node[fill=white,inner sep=1pt] {\footnotesize $a$} (11tt);
\draw[->,thick,dotted] (01t) -- node[fill=white,inner sep=1pt] {\footnotesize $p\vel q$} (01tt);
\draw[->,thick,dotted] (11t) -- node[fill=white,inner sep=1pt] {\footnotesize $p\vel q$}  (11tt);
\draw[->,thick,dotted] (10t) -- node[fill=white,inner sep=1pt] {\footnotesize $p\vel q$}  (10tt);
\node (01ttt) at (8,2) {$\np\q$};
\node (10ttt) at (9.5,.7) {$\p\nq$};
\node (11ttt) at (9.5,2.7) {$\p\q$};
\draw (01ttt) -- node[fill=white,inner sep=1pt] {\footnotesize $b$} (11ttt);
\draw (10ttt) -- node[fill=white,inner sep=1pt] {\footnotesize $a$} (11ttt);
\draw[->,thick,dotted] (01tt) -- node[near start,fill=white,inner sep=1pt] {\footnotesize $[a]$}  (01ttt);
\draw[->,thick,dotted] (11tt) -- node[fill=white,inner sep=1pt] {\footnotesize $[a]$} (11ttt);
\draw[->,thick,dotted] (10tt) -- node[fill=white,inner sep=1pt] {\footnotesize $[a]$} (10ttt);
\node (01tttt) at (11,2) {$\np\q$};
\node (10tttt) at (12.5,.7) {$\p\nq$};
\node (11tttt) at (12.5,2.7) {$\p\q$};
\draw (01tttt) -- node[fill=white,inner sep=1pt] {\footnotesize $b$} (11tttt);
\draw (10tttt) -- node[fill=white,inner sep=1pt] {\footnotesize $a$} (11tttt);
\draw[->,thick,dotted] (01ttt) -- node[fill=white,inner sep=1pt,near end] {\footnotesize $[b]$} (01tttt);
\draw[->,thick,dotted] (11ttt) -- node[fill=white,inner sep=1pt] {\footnotesize $[b]$} (11tttt);
\draw[->,thick,dotted] (10ttt) -- node[fill=white,inner sep=1pt] {\footnotesize $[b]$} (10tttt);
\node (01b) at (3,-.35) {$(i)$};
\node (01bb) at (6,-.35) {$(ii)$};
\node (01bbb) at (9,-.35) {$(iii)$};
\node (01bbbb) at (12,-.35) {$(iv)$};
\node (01x) at (8,5) {$\np\q$};
\node (10x) at (9.5,3.7) {$\p\nq$};
\node (11x) at (9.5,5.7) {$\p\q$};
\draw (01x) -- node[fill=white,inner sep=1pt] {\footnotesize $b$} (11x);
\draw (10x) -- node[fill=white,inner sep=1pt] {\footnotesize $a$} (11x);
\draw[->,thick,dotted] (01tt) to node[fill=white,inner sep=1pt] {\footnotesize $[b]$}  (01x);
\draw[->,thick,dotted] (11tt) to node[fill=white,inner sep=1pt] {\footnotesize $[b]$} (11x);
\draw[->,bend left = 30,thick,dotted] (10tt) to node[fill=white,inner sep=1pt,near end] {\footnotesize $[b]$} (10x);
\node (01xx) at (11,5) {$\np\q$};
\node (10xx) at (12.5,3.7) {$\p\nq$};
\node (11xx) at (12.5,5.7) {$\p\q$};
\draw (01xx) -- node[fill=white,inner sep=1pt] {\footnotesize $b$} (11xx);
\draw (10xx) -- node[fill=white,inner sep=1pt] {\footnotesize $a$} (11xx);
\draw[->,thick,dotted] (01x) -- node[fill=white,inner sep=1pt, near end] {\footnotesize $[a]$} (01xx);
\draw[->,thick,dotted] (11x) -- node[fill=white,inner sep=1pt] {\footnotesize $[a]$} (11xx);
\draw[->,thick,dotted] (10x) -- node[fill=white,inner sep=1pt] {\footnotesize $[a]$} (10xx);
\draw[->,bend left = 40,dashed] (11xx) to node[fill=white,inner sep=1pt] {\footnotesize $\tri_a$} (11tttt);
\draw[->,bend right = 30,dashed] (11xx) to node[fill=white,inner sep=1pt] {\footnotesize $\tri_b$} (11x);
\draw[->,bend left = 30,dashed] (10tt) to node[fill=white,inner sep=1pt] {\footnotesize $\tri_b$} (10t);
\draw[->,bend right = 20,dashed] (11x) to node[fill=white,inner sep=1pt] {\footnotesize $\tri_a$} (11t);
\end{tikzpicture}
\caption{Asynchronous model for the history $(p\vel q)ab$, given initial uncertainty for $a$ about $q$ and for $b$ about $p$. Cf.\ to Figure~\ref{fig}.}
\label{fig.history}
\end{figure}
\end{example}

\subsection{Asynchronous broadcast logic} \label{sec.kms}

In \cite{KnightMS19} and the already cited related works the authors develop a logical semantics for sending and receiving messages that are announcements and where the epistemic notion is one of knowledge and not one of belief, unlike ours. The modelling justifications and consequences of these different semantics were discussed at length in Section~\ref{sec.korb}. The logical language is the same as ours, except that they write $\bigcirc_a$ instead of (the diamond form) $\dia{a}$ for the reception modality, and $K_a$ instead of $B_a$ for the epistemic modality. They interpret their language on structures called asynchronous pre-models, with domain elements that are triples $(s,\sigma,c)$ where $s$ is an abstract state (from the domain of some initial Kripke model), $\sigma$ is a sequence of formulas taken from a protocol of allowed sequences of formulas, and $c$ is a \emph{cut}, listing for each agent which announcements in the sequence $\sigma$ that agent has already received. The relation between such cuts and the histories in $AA$ will be explored in the next Section~\ref{sec.cuts}.

Announcements must be true when made and are individually received by the agents in the order in which they were sent. The knowledge modality $K_a$ is interpreted over histories of arbitrary length, thus guaranteeing that knowledge is correct: $K_a \phi \imp \phi$ is valid. When interpreting knowledge on pre-models, not all triples $(s,\sigma,c)$ of the pre-model are taken into account but only those where all formulas of $\sigma$ could have been truthfully announced. They call this the requirement of \emph{consistency} of $\sigma$ with $s$ (where we called this \emph{agreement}, the relation $\bowtie$). Their semantics is then based on mutual recursion of `truth' and `consistency', similar to ours.

As their epistemic notion is one of knowledge, also taking into account announcements that have not yet been received, they face the already mentioned issue of circularity and of a well-founded semantics. They provide two well-founded solutions to this circularity problem. Their first solution is to restrict the structures to (initial) models that are finite point-generated trees, and where the model transformations are relative to the root of the model. This solution is reminiscent of \cite{lomuscioetal:1998b}. Their second solution is to restrict the language to the so-called existential fragment wherein negations are only allowed of atoms, and wherein modalities are only allowed in `diamond' form $\hat{K}_a\phi$, $\bigcirc_a$, and $\dia{\psi}\phi$. They present some validities for their logical semantics, such as $\bigcirc_a\bigcirc_b\phi \eq \bigcirc_b\bigcirc_a\phi$: without intervening announcements, the order of reception does not matter. They do not provide an axiomatization. They obtain various complexity results.  An interesting special case for complexity is when all announced formulas are  Booleans. They show that the complexity of model checking for that case is PSPACE-complete, and that  the complexity of satisfiability is NExptime-hard. They also show that the complexity of model checking on finite trees is in PSPACE and that the complexity of model checking for the existential fragment is in Exptime.

\subsection{Cuts and distributed computing} \label{sec.cuts}

In distributed computing, the activity of each agent $a \in A$ consists of events that are either messages sent to other agents or messages received from other agents. These events are temporally ordered. A \emph{cut} is a selection of one of these events for each agent. For each agent we can then distinguish the events before and up to the cut (i.e., weakly before) from those that come after (i.e., strictly after) the cut. The cut is inconsistent if there are agents $a,b$ and a message $m$ such that the event of $a$ sending $m$ is after the cut for $a$, and the event of $b$ receiving $m$ is (weakly) before the cut for $b$. Otherwise the cut is consistent. An inconsistent cut violates that messages must have been sent (and therefore `happened') before they are received. For the `happened before' relation see the foundational \cite{lamport:1978} (that does not use the term `cut'), for `cuts' see e.g.\ \cite{panangadenetal:1992} and \cite[p.\ 45]{kshemkalyanietal:2008}.

In the logic $AA$, the agents only receive messages and the anonymous environment (the `announcer') only sends messages. The requirement of consistency is satisfied by words $\alpha \in (A \union \Laa)$ that are histories: a word $\alpha$ is a history if for all prefixes $\beta$ of $\alpha$ and for all agents $a$ we have that $|\beta|_a \leq |\beta|_!$, i.e., if at any stage an agent $a$ can only receive an announcement that has already been made. In that sense a word corresponds to an inconsistent cut if it is not a history. In this section we investigate the relation between histories and cuts. In another sense a history may be said to correspond to an inconsistent cut if we consider the announcement $\phi$ as the message $B_a \phi$ sent by agent $a$, as in distributed computing. That will be investigated in Section~\ref{sec.dc}.

Given $n \in \Naturals$, a \emph{$n$-cut} is a function $f: A \imp [0,n]$.  A history $\alpha$  determines a pair $(\alpha\proj_!,f^\alpha)$, where $f^\alpha$ is the $|\alpha|_!$-cut $f$ such that for all $a \in A$, $f(a) = |\alpha|_a$. This notion of cut was proposed in \cite{KnightMS19} and in the prior SR 2017 version of this work, but it is  merely another example of the standard notion of cut in distributed computing \cite{kshemkalyanietal:2008}. It will be clear that $f^\alpha= f^\beta$ iff $|\alpha|_! = |\beta|_!$ and for all $a \in A$, $|\alpha|_a= |\beta|_a$.

Different cuts always correspond to different histories, but different histories $\alpha,\beta$ may correspond to the same cut. If $\alpha\proj_!=\beta\proj_!$ and $|\alpha|_a=|\beta|_a$ for all $a\in A$, then obviously $(\alpha\proj_!,f^\alpha) = (\beta\proj_!, f^\beta)$: all agents have received the same announcements in the same order. However, it may be the case that $s\bowtie \alpha$ and $s\not\bowtie\beta$ for a certain state $s$ in some given model. As an example, consider the histories $\alpha = p(\neg B_a p)aa$ and $\beta = pa(\neg B_ap)a$. Given a model and a state $s$ where $p$ is true but unknown by $a$, $\alpha$ agrees with $s$ but $\beta$ does not agree with $s$. In fact, $\beta$ cannot agree with any state of any model, as after agent $a$ receives $p$ the announcement $\neg B_a p$ cannot be truthful. Histories are a more refined relation on message sending and receiving than cuts.

This suggests that histories have more `expressive power' (in an informal sense) than cuts. However, to an important extent this is not the case: histories containing the same sequence of announcements and corresponding to the same cut are `indistinguishable' in those states with which they both agree, in the sense that those states make the same formulas true, they have the same theory. A minor lemma precedes the proposition stating this result. Consequently, this more closely relates the cut-based semantics of \cite{KnightMS19} to our semantics, although the epistemic modalities are different.
\begin{lemma} \label{lemma.cut}
Let histories $\alpha,\beta$ be given such that $\alpha\proj_!=\beta\proj_!$ and $f^\alpha = f^\beta$. \begin{enumerate}
\item For all $a \in A$ and histories $\gamma$, $\alpha \tri_a \gamma$ iff $\beta \tri_a \gamma$. \label{cutone}
\item For all $a \in A$, if $|\alpha|_! < |\alpha|_a$ then $f^{\alpha{a}} = f^{\beta{a}}$.  \label{cuttwo}
\item For all $\phi \in \Laa$, $f^{\alpha\phi} = f^{\beta\phi}$. \label{cutthree}\end{enumerate}
\end{lemma}
\begin{proof}
Recall that $f^\alpha = f^\beta$, iff $|\alpha|_!=|\beta|_!$ and for all $b \in A$, $|\alpha|_b=|\beta|_b$. (Note that assumption $\alpha\proj_!=\beta\proj_!$ already implies $|\alpha|_!=|\beta|_!$.)
\begin{enumerate}
\item
From $\alpha\proj_!=\beta\proj_!$ and $|\alpha|_a=|\beta|_a$ we obtain $\alpha\proj_{!a}=\beta\proj_{!a}$. We now use Def.~\ref{definition:view:relation} of the view relation.

\item
Given that $|\alpha|_! < |\alpha|_a$, $\alpha{a}$ and $\beta{a}$ are histories. From $\alpha\proj_!=\beta\proj_!$ we obtain that $\alpha{a}\proj_!=\beta{a}\proj_!$ ($= \alpha\proj_!$). Also, $|\alpha{a}|_a = |\alpha|_a +1 = |\beta|_a+1 = |\beta{a}|_a$, and for $b \neq a$, $|\alpha|_b = |\alpha{a}|_b = |\beta{a}|_b = |\beta|_a$. Therefore $f^{\alpha{a}} = f^{\beta{a}}$.

\item
From $\alpha\proj_!=\beta\proj_!$ we obtain $\alpha\phi\proj_!=\beta\phi\proj_!$, as e.g.\ $\alpha\phi\proj_! = (\alpha\proj_!)\phi$. Also, for all $b \in A$, $|\alpha\phi|_b=|\alpha|_b=|\beta|_b=|\beta\phi|_b$. Therefore $f^{\alpha\phi} = f^{\beta\phi}$.
\end{enumerate}
\end{proof}

\begin{proposition}
Let be given $\phi \in\Laa$, model $M = (W,R,V)$ with $s \in W$, histories $\alpha,\beta$ such that $f^\alpha = f^\beta$, and $s\bowtie\alpha$ and $s\bowtie\beta$. Then $s,\alpha\models\phi$ iff $s,\beta\models\phi$.
\end{proposition}
\begin{proof}
	The proof is by induction on $\phi$. The trivial cases when $\phi = p, \bot, \psi_1\vee\psi_2, \neg\psi$ are left to the reader.

Case belief: $s,\alpha\models B_a \phi$, iff $t,\gamma\models\phi$ for all $(t,\gamma)$ such that $R_ast$, $\alpha \tri_a \gamma$, and $t \bowtie \gamma$, iff (by Lemma \ref{lemma.cut}.\ref{cutone}) $t,\gamma\models\phi$ for all $(t,\gamma)$ such that $R_ast$, $\beta \tri_a \gamma$, and $t \bowtie \gamma$, iff $s,\beta\models B_a \phi$.

Case reception: $s,\alpha\models [a]\psi$, iff $|\alpha|_a < |\alpha|_!$ implies $s,\alpha{a}\models \psi$, iff (Lemma~\ref{lemma.cut}.\ref{cuttwo} and induction) $|\beta|_a < |\beta|_!$ implies $s,\beta{a}\models \psi$.

Case announcement: $s,\alpha\models [\phi]\psi$, iff $s,\alpha \models \phi$ implies $s,\alpha\phi\models \psi$, iff (Lemma~\ref{lemma.cut}.\ref{cutthree} and twice induction) $s,\beta \models \phi$ implies $s,\beta\phi\models \psi$, iff $s,\beta\models [\phi]\psi$.

\end{proof}

\section{Distributed computing and asynchronous announcements} \label{sec.dc}

In this section we give a detailed example of our logic and its semantics that is described in terms of distributed computing. Consider two agents $a$, $b$ each only knowing the (binary) value of their local state, respectively, $p$ and $q$. We can see this as an interpreted or distributed system in terms of \cite{halpernmoses:1990,faginetal:1995} for two processes/agents $a,b$ with possible global states $pq$ (i.e., $(p,q)$), $p\overline{q}$, $\overline{p}q$, $\overline{p}\overline{q}$, and with knowledge of the agents induced by their local state. We can alternatively see this as the model depicted in Figure \ref{fig}(i). Let us assume that the actual values are $p$ for $a$ and $q$ for $b$.

In dynamic epistemic logics, an agent $a$ truthfully sending (broadcasting) a message $\phi$ is simulated by the environment sending the (true) message $B_a \phi$. These broadcast messages are the announcements. This is no different for asynchronous announcement logic.

Agent $a$ now sends her local value $p$ to $b$, and after receiving this message agent $b$ sends an acknowledgement to $a$, who then receives that. These are the two announcements $B_a p$ ($a$ announces $p$, for `the value of my local state is $p$') and $B_b B_a p$ ($b$ announces $B_a p$, for `I now know that the value of your local state is $p$'). We can model this in asynchronous announcement logic in different ways: (i) we only model the asynchronous reception of the messages by $a$ and $b$, (ii) we introduce the environment as an agent $e$ sending the messages, or (iii) we let $a$ and $b$ send and receive messages.

\medskip

\noindent {\bf Two processes a and b and implicit environment} \quad Consider agents/processes $a,b$ and reception events $a_1,a_2, b_1,b_2$ of the two announcements $B_a p$ and $B_b B_a p$ made by the environment, where a `before' relation is a partial order between these events, along the familiar terms described in e.g.\ \cite{lamport:1978, panangadenetal:1992,kshemkalyanietal:2008}. How to model the announcements is left implicit for now and deferred until later. We get:

\begin{center}
\begin{tikzpicture}
\node (a) at (0,1) {$a$};
\node (b) at (0,0) {$b$};
\node (e) at (0,0) {$b$};
\draw (1,1) -- (10,1);
\draw (1,0) -- (10,0);
\node (21) at (2,1) {$\bullet$};
\node (21) at (2,1.3) {$a_1$};
\node (61) at (6,1) {$\bullet$};
\node (61) at (6,1.3) {$a_2$};
\node (51) at (5,0) {$\bullet$};
\node (51) at (5,0.3) {$b_1$};
\node (91) at (9,0) {$\bullet$};
\node (91) at (9,0.3) {$b_2$};
\draw[dashed] (4,-1) -- (4,2.5);
\draw[dashed] (1,-1) -- (10,2.3);
\node (cons) at (1,-1) {inconsistent};
\node (cons) at (4,-1) {consistent};
\end{tikzpicture}
\end{center}

What is the partial order? Given the announcements, apart from the obviously enforced $a_1 < a_2$ and $b_1 < b_2$ the only other constraint is that $b_1 < a_2$: agent $a$ cannot receive the message $B_b B_a p$ before agent $b$ has received message $B_a p$. A `cut' determines how many messages each agent has received. The depicted consistent cut corresponds to the history $(B_a p)a$. The inconsistent cut does not respect that $b_1 < a_2$. It corresponds to the history $(B_a p)a(B_b B_a p)a$ (or to $(B_a p)(B_b B_a p)aa$). In our terms we would say that the state $s$ where $p$ and $q$ are both true does not agree with that history, i.e., $s \not\bowtie (B_a p)a(B_b B_a p)a$. 

\medskip

\noindent {\bf Three processes a, b, and e} \quad We now also model the environment explicitly as a process/agent $e$. We can also add this agent $e$ to the models in which we interpret asynchronous announcements, namely as an agent with the identity accessibility relation on such models, which implies that $B_e \phi \eq \phi$: truthful messages --- $B_e \phi$ --- are also true --- $\phi$. Therefore, announcements made by the environment need not be prefixed by the $B_e$ operator. Let us assume that $s_1$ is the announcement of $B_a p$ and $s_2$ is the announcement of $B_b B_a p$. In dynamic epistemic logics, an announcement made by an agent is also received by this agent (if you say something, you also hear it yourself). This is relevant because the message content may be modal and thus affect the knowledge or belief of the announcing agent (such as $a$ announcing to $b$: ``$p$ is true but you don't know this,'' $K_a (p \et \neg K_b p)$). This is unlike a broadcast in distributed systems. But in the following depiction we therefore must also involve two reception events $e_1$ and $e_2$ for the environment.

\begin{center}
\begin{tikzpicture}
\node (e) at (0,2) {$e$};
\node (a) at (0,1) {$a$};
\node (b) at (0,0) {$b$};
\draw (1,2) -- (10,2);
\draw (1,1) -- (10,1);
\draw (1,0) -- (10,0);
\node (02) at (1,2) {$\bullet$};
\node (02) at (1,2.3) {$s_1$};
\node (32) at (3,2) {$\bullet$};
\node (32) at (3,2.3) {$s_2$};
\node (02r) at (1.5,2) {$\bullet$};
\node (02r) at (1.5,2.3) {$e_1$};
\node (32r) at (3.5,2) {$\bullet$};
\node (32r) at (3.5,2.3) {$e_2$};
\node (21) at (2,1) {$\bullet$};
\node (21) at (2,1.3) {$a_1$};
\node (61) at (6,1) {$\bullet$};
\node (61) at (6,1.3) {$a_2$};
\node (51) at (5,0) {$\bullet$};
\node (51) at (5,0.3) {$b_1$};
\node (91) at (9,0) {$\bullet$};
\node (91) at (9,0.3) {$b_2$};
\draw[dashed] (5.8,-1) -- (2,2.5);
\draw[dashed] (4,-1) -- (4,2.5);
\draw[dashed] (1,-1) -- (10,2.3);
\node (cons) at (1,-1) {inconsistent};
\node (cons) at (4,-1) {inconsistent};
\node (cons) at (6.7,-1) {consistent};
\end{tikzpicture}
\end{center}

The partial relation $a_1 < a_2$, $b_1 < b_2$, $b_1 < a_2$ enforced so far is expanded with $s_1 < a_1,b_1,e_1$ and $s_2 < a_2,b_2,e_2$ (a message is sent before it is received) and the event ordering $s_1 < s_2$ and $e_1 < e_2$ for the environment.

However, the fact that messages sent by the environment are {\em true} when sent (the agreement relation of $AA$ again), enforces  other logical/causal constraints that are of more interest. The environment announcing $B_b B_a p$ stands for the truth of that formula. This enforces that $b_1 < s_2$ (but it is consistent with $a_1 < s_2$ and with $s_2 < a_1$). From $b_1 < s_2$ now follows with the above constraint $s_2 < a_2$ that $b_1 < a_2$.

If we abstract from the reception of the environment, the leftmost cut in the figure corresponds to the histories $(B_a p)(B_bB_ap)aa$ and $(B_a p)a(B_bB_a p)a$. These are inconsistent because $s \not\bowtie(B_a p)(B_bB_ap)aa$ and $s \not\bowtie(B_a p)a(B_bB_ap)a$. The middle cut is also inconsistent (it does not obey that $b_1 < s_2$) as it corresponds to histories $(B_a p)(B_bB_ap)a$ and $(B_a p)a(B_bB_ap)$. The right cut is consistent. It corresponds to history $(B_a p)a$.

\medskip

\noindent {\bf Two processes a and b sending and receiving} \quad We now let agents $a$ and $b$ send the messages themselves. The inconsistent cut corresponds to $(B_a p)aa$. In terms of $AA$ it is inconsistent not for lack of agreement (it contains messages that could not have been truthfully announced), but because it is not a history. Expression $(B_a p)aa$ is a \emph{word} in the language $\{a,b\} \union \Laa$ but not a \emph{history}, as $2 = |(B_a p)aa|_a > |(B_a p)aa|_! = 1$. The depicted consistent cut corresponds to history $(B_a p)a$.

\begin{center}
\begin{tikzpicture}
\node (a) at (0,1) {$a$};
\node (b) at (0,0) {$b$};
\node (e) at (0,0) {$b$};
\draw (1,1) -- (10,1);
\draw (1,0) -- (10,0);
\node (11b) at (1,1) {$\bullet$};
\node (11) at (1,1.3) {$s_1$};
\node (21b) at (2,1) {$\bullet$};
\node (21) at (2,1.3) {$a_1$};
\node (61b) at (6,1) {$\bullet$};
\node (61) at (6,1.3) {$a_2$};
\node (41b) at (6,0) {$\bullet$};
\node (41) at (6,0.3) {$s_2$};
\node (51b) at (5,0) {$\bullet$};
\node (51) at (5,0.3) {$b_1$};
\node (91b) at (9,0) {$\bullet$};
\node (91) at (9,0.3) {$b_2$};
\draw[dashed] (4,-1) -- (4,2.5);
\draw[dashed] (1,-1) -- (10,2.3);
\node (cons) at (1,-1) {inconsistent};
\node (cons) at (4,-1) {consistent};
\end{tikzpicture}
\end{center}

This modelling solution is stretching the use of asynchronous announcement logic. Agent $a$ sending a message $\phi$ in $AA$ is simulated by the announcement $B_a \phi$ by the environment, and therefore $a$ and $b$ both have to receive that. But no order is enforced for this reception: $a_1<b_1$ and $b_1<a_1$ are both causally permitted. In other words, the more intuitive history $(B_a p)ab(B_bB_ap)ab$ and the counterintuitive history $(B_a p)b(B_bB_ap)baa$ are both permitted. It seems quite possible to adjust the semantics of $AA$ to enforce only histories wherein the agent sending the message is always the first to receive it, and thus allowing to abstract from it. Such logics, deferred to future research, are more suitable for settings wherein agents both send and receive messages.

\section{Conclusions and further research}

We presented asynchronous announcement logic $AA$, a logic of epistemic change due to  announcements, with separate modalities for sending and for receiving such messages. Our epistemic modality is one of belief, not one of knowledge. We provided an axiomatization for this logic $AA$ that is a reduction system: every formula is equivalent to a formula without announcement and reception modalities. The logic $AA$ is therefore also decidable. We determined results for special formulas and for special model classes: the positive formulas are preserved after update, and on the model class $\mathcal{S}5$, belief of positive formulas is correct and thus knowledge. The complexity of model checking and of satisfiability of AA is left for further research. We envisage numerous generalizations of our work, such as for subgroups synchronously receiving announcements, for non-public actions, and instead of belief for knowledge, wherein one also reasons about the future.

\paragraph*{Acknowledgements} For their comments on or involvement in prior versions of this manuscript, or helpful interactions during seminars on this topic, we thank: Malvin Gattinger, Davide Grossi, Ramanujam, Armando Casta\~{n}eda, Sergio Rajsbaum.

Hans van Ditmarsch is also affiliated to IMSc, Chennai, India. Since 2008, my IMSc host Ramanujam engaged me in discussions on modelling asynchronous messaging in dynamic epistemic logic. His support has always motivated and encouraged me. Sophia Knight introduced me to the work of Prakash Panangaden that inspired asynchronous announcements, and later Bastien Maubert and Fran\c{c}ois Schwarzentruber became involved in this pursuit, all at the CELLO team in LORIA. Then, our ways parted and their work took the direction of \cite{KnightMS19}. Without this past collaboration the underlying work could not have resulted, and I thank them for all the interaction (and music) that we had together.

\bibliographystyle{plain}
\bibliography{biblio2020}

\section*{Appendix: results for the well-founded order $\ll$} \label{appendix}

The first part of the Appendix contains results for the order $\ll$, and the functions ${\parallel}\cdot{\parallel}$ and $deg(\cdot)$. First, concerning ${\parallel}\cdot{\parallel}$, note that, obviously, for all words $\alpha$ over $A\cup\Laa$ and for all $a\in A$, ${\mid}\alpha{\mid}_{a}\leq{\parallel}\alpha{\parallel}$.
\begin{lemma}
For all words $\alpha$ over $A\cup\Laa$, for all $a\in A$ and for all $\varphi\in\Laa$, ${\parallel}a\alpha{\parallel}={\parallel}\alpha{\parallel}+1$ and ${\parallel}\varphi\alpha{\parallel}={\parallel}\alpha{\parallel}+{\parallel}\varphi{\parallel}$.
\end{lemma}
\begin{proof}
The proof is by $<$-induction on ${\mid}\alpha{\mid}$.
\end{proof}
\begin{lemma}
For all words $\alpha$ over $A\cup\Laa$ and for all $\varphi\in\Laa$, ${\parallel}\lbrack\alpha\rbrack\varphi{\parallel}=2{\parallel}\alpha{\parallel}+{\parallel}\varphi{\parallel}$.
\end{lemma}
\begin{proof}
The proof is by $<$-induction on ${\mid}\alpha{\mid}$.
\end{proof}
\begin{lemma}
For all words $\alpha$ over $A\cup\Laa$, for all $k\in\N$, for all $\psi_{1},\ldots,\psi_{k}\in\Laa$ and for all $\varphi\in\Laa$, if $\alpha{\upharpoonright}_{!}=\psi_{1}\ldots\psi_{k}$ then $\deg(\lbrack\alpha\rbrack\varphi)=\deg(\psi_{1})+\ldots+\deg(\psi_{k})+\deg(\varphi)$.
\end{lemma}
\begin{proof}
The proof is by $<$-induction on ${\mid}\alpha{\mid}$.
\end{proof}
\begin{lemma}\label{lemma:let:be:a:word:be}
Let $\alpha$ be a history.
Let $k$ be a nonnegative integer and $\varphi_{1},\ldots,\varphi_{k}\in\Laa$.
In the single-agent case, if ${\alpha}{\upharpoonright}_!=\varphi_{1}\ldots\varphi_{k}$ then ${\parallel}{\alpha}{\parallel}={\mid}{\alpha}{\mid}_{a}+{\parallel}{\varphi_{1}}{\parallel}+\ldots+{\parallel}{\varphi_{k}}{\parallel}$.
Otherwise, in the multi-agent case, considering an enumeration $(a_{1},\ldots,a_{n})$ of $A$ without repetition, if ${\alpha}{\upharpoonright}_!=\varphi_{1}\ldots\varphi_{k}$ then ${\parallel}{\alpha}{\parallel}={\mid}{\alpha}{\mid}_{a_{1}}+\ldots+{\mid}{\alpha}{\mid}_{a_{n}}+{\parallel}{\varphi_{1}}{\parallel}+\ldots+{\parallel}{\varphi_{k}}{\parallel}$
\end{lemma}
\begin{proof}
The proof is by $<$-induction on ${\mid}{\alpha}{\mid}$.
\end{proof}
\begin{lemma}\label{lemma:for:all:words:if:then:B}
Let $\alpha,\beta$ be histories and $a\in A$.
If ${\alpha}{\triangleright_a}{\beta}$ then ${\parallel}{\beta}{\parallel}\leq |A| \cdot {\parallel}{\alpha}{\parallel}$.
\end{lemma}
\begin{proof}
Suppose ${\alpha}{\triangleright_a}{\beta}$.
Hence, ${\mid}{\beta}{\mid}_{a}={\mid}{\alpha}{\mid}_{a}$.
Moreover, \weg{by Lemma~\ref{lemma:histories:triangle:right:prefix},} ${\beta}{\upharpoonright_{!}}$ is a prefix of ${\alpha}{\upharpoonright_{!}}$ and by Lemma~\ref{lemma:5:bis}, for all $b\in A\setminus\{a\}$, ${\mid}{\beta}{\mid}_{b}\leq{\mid}{\alpha}{\mid}_{a}$.
Thus, by Lemma~\ref{lemma:let:be:a:word:be}, ${\parallel}\beta{\parallel}\leq{\mid}A{\mid}\cdot{\mid}\alpha{\mid}_{a}+{\parallel}\alpha{\parallel}-{\mid}\alpha{\mid}_{a}$.
Consequently, ${\parallel}\beta{\parallel}\leq({\mid}A{\mid}-1)\cdot{\mid}\alpha{\mid}_{a}+{\parallel}\alpha{\parallel}$.
Hence, ${\parallel}\beta{\parallel}\leq{\mid}A{\mid}\cdot{\parallel}\alpha{\parallel}$.
\end{proof}
\begin{lemma}\label{if:triangle:then:greater:or:equal}
Let $\alpha,\beta$ be histories and $a\in A$.
Let $k,l\in\N$ and $\varphi_{1},\ldots,\varphi_{k},\psi_{1},\ldots,\psi_{l}\in\Laa$ be such that $\alpha{\upharpoonright}_{!}=\varphi_{1}\ldots\varphi_{k}$ and $\beta{\upharpoonright}_{!}=\psi_{1}\ldots\psi_{l}$.
If ${\alpha}{\triangleright_a}{\beta}$ then $\deg(\varphi_{1})+\ldots+\deg(\varphi_{k})\geq\deg(\psi_{1})+\ldots+\deg(\psi_{l})$.
\end{lemma}
\begin{proof}
Suppose $\alpha\triangleright_{a}\beta$.
Since $\alpha{\upharpoonright}_{!}=\varphi_{1}\ldots\varphi_{k}$ and $\beta{\upharpoonright}_{!}=\psi_{1}\ldots\psi_{l}$, \weg{we obtain by Lemma~\ref{lemma:histories:triangle:right:prefix} that} $\psi_{1}\ldots\psi_{l}$ is a prefix of $\varphi_{1}\ldots\varphi_{k}$.
Thus, $\deg(\varphi_{1})+\ldots+\deg(\varphi_{k})\geq\deg(\psi_{1})+\ldots+\deg(\psi_{l})$.
\end{proof}
\begin{lemma}\label{lemma:let:be:the:well:founded} \ {} \vspace{-.2cm}
\begin{enumerate}
\item $(\alpha,\varphi)\ll(\alpha a,\varphi)$,
\item $(\alpha,\psi)\ll(\alpha\psi,\varphi)$ and $(\alpha,\bot)\ll(\alpha\varphi,\bot)$,
\item $(\alpha,\varphi)\ll(\alpha,\neg\varphi)$ and $(\alpha,\varphi)\ll(\alpha\varphi,\bot)$,
\item $(\alpha,\varphi)\ll(\alpha,\varphi\vee\psi)$ and $(\alpha,\psi)\ll(\alpha,\varphi\vee\psi)$,
\item if ${\alpha}{\triangleright_a}{\beta}$ then $(\beta,\varphi)\ll(\alpha,B_a\varphi)$,
\item $(\alpha,\varphi)\ll(\alpha,\lbrack\varphi\rbrack\psi)$ and $(\alpha\varphi,\psi)\ll(\alpha,\lbrack\varphi\rbrack\psi)$,
\item $(\alpha a,\varphi)\ll(\alpha,\lbrack a\rbrack\varphi)$,
\item $(\alpha,\bot)\ll(\alpha,p)$,
\item $(\alpha,\bot)\ll(\alpha,B_{a}\varphi)$,
\item $(\alpha,\bot)\ll(\alpha,\neg\varphi)$.
\end{enumerate}
\end{lemma}
\begin{proof} \ {} \vspace{-.4cm}

\begin{enumerate}
\item Remark that $\deg(\alpha,\varphi)=\deg(\alpha a,\varphi)$.
Moreover, since ${\parallel}{\alpha}{\parallel}+{\parallel}{\varphi}{\parallel}<{\parallel}{\alpha}{\parallel}+1+{\parallel}{\varphi}{\parallel}$,  $(\alpha,\varphi)\ll(\alpha a,\varphi)$.
\item Remark that $\deg(\alpha,\psi)\leq\deg(\alpha\psi,\varphi)$.
Moreover, since ${\parallel}{\alpha}{\parallel}+{\parallel}{\psi}{\parallel}<{\parallel}{\alpha}{\parallel}+{\parallel}{\psi}{\parallel}+{\parallel}{\varphi}{\parallel}$,  $(\alpha,\psi)\ll(\alpha\psi,\varphi)$.
In other respect, $\deg(\alpha,\bot)\leq\deg(\alpha\varphi,\bot)$.
Moreover, since ${\parallel}{\alpha}{\parallel}+1<{\parallel}{\alpha}{\parallel}+{\parallel}{\varphi}{\parallel}+1$,  $(\alpha,\bot)\ll(\alpha\varphi,\bot)$.
\item Remark that $\deg(\alpha,\varphi)=\deg(\alpha,\neg\varphi)$.
Moreover, since ${\parallel}{\alpha}{\parallel}+{\parallel}{\varphi}{\parallel}<{\parallel}{\alpha}{\parallel}+{\parallel}{\varphi}{\parallel}+1$,  $(\alpha,\varphi)\ll(\alpha,\neg\varphi)$.
In other respect, remark that $\deg(\alpha,\varphi)=\deg(\alpha\varphi,\bot)$.
Moreover, since ${\parallel}\alpha{\parallel}+{\parallel}\varphi{\parallel}<{\parallel}\alpha{\parallel}+{\parallel}\varphi{\parallel}+1$,  we obtain that $(\alpha,\varphi)\ll(\alpha\varphi,\bot)$.
\item Remark that $\deg(\alpha,\varphi)\leq\deg(\alpha,\varphi\vee\psi)$ and $\deg(\alpha,\psi)\leq\deg(\alpha,\varphi\vee\psi)$.
Moreover, since ${\parallel}{\alpha}{\parallel}+{\parallel}{\varphi}{\parallel}<{\parallel}{\alpha}{\parallel}+{\parallel}{\varphi}{\parallel}+{\parallel}{\psi}{\parallel}$ and ${\parallel}{\alpha}{\parallel}+{\parallel}{\psi}{\parallel}<{\parallel}{\alpha}{\parallel}+{\parallel}{\varphi}{\parallel}+{\parallel}{\psi}{\parallel}$, we obtain that $(\alpha,\varphi)\ll(\alpha,\varphi\vee\psi)$ and $(\alpha,\psi)\ll(\alpha,\varphi\vee\psi)$.
\item Suppose ${\alpha}{\triangleright_a}{\beta}$.
Let $k,l\in\N$ and $\varphi_{1},\ldots,\varphi_{k},\psi_{1},\ldots,\psi_{l}$ be formulas such that $\alpha{\upharpoonright}_{!}=\varphi_{1}\ldots\varphi_{k}$ and $\beta{\upharpoonright}_{!}=\psi_{1}\ldots\psi_{l}$.
Hence, by Lemma~\ref{if:triangle:then:greater:or:equal}, $\deg(\varphi_{1})+\ldots+\deg(\varphi_{k})\geq\deg(\psi_{1})+\ldots+\deg(\psi_{l})$.
Thus, $\deg(\beta,\varphi)<\deg(\alpha,B_a\varphi)$.
Consequently, $(\beta,\varphi)\ll(\alpha,B_a\varphi)$.
\item Remark that $\deg(\alpha,\varphi)\leq\deg(\alpha,\lbrack\varphi\rbrack\psi)$ and $\deg(\alpha\varphi,\psi)=\deg(\alpha,\lbrack\varphi\rbrack\psi)$.
Moreover, since ${\parallel}{\alpha}{\parallel}+{\parallel}{\varphi}{\parallel}<{\parallel}{\alpha}{\parallel}+2{\parallel}{\varphi}{\parallel}+{\parallel}{\psi}{\parallel}$ and ${\parallel}{\alpha}{\parallel}+{\parallel}{\varphi}{\parallel}+{\parallel}{\psi}{\parallel}<{\parallel}{\alpha}{\parallel}+2{\parallel}{\varphi}{\parallel}+{\parallel}{\psi}{\parallel}$, we obtain that $(\alpha,\varphi)\ll(\alpha,\lbrack\varphi\rbrack\psi)$ and $(\alpha\varphi,\psi)\ll(\alpha,\lbrack\varphi\rbrack\psi)$.
\item Remark that $\deg(\alpha a,\varphi)=\deg(\alpha,\lbrack a\rbrack\varphi)$.
Moreover, since ${\parallel}{\alpha}{\parallel}+1+{\parallel}{\varphi}{\parallel}<{\parallel}{\alpha}{\parallel}+{\parallel}{\varphi}{\parallel}+2$, we obtain that $(\alpha a,\varphi)\ll(\alpha,\lbrack a\rbrack\varphi)$.
\item Remark that $\deg(\alpha,\bot)=\deg(\alpha,p)$.
Moreover, since ${\parallel}{\alpha}{\parallel}+1<{\parallel}{\alpha}{\parallel}+2$, $(\alpha,\bot)\ll(\alpha,p)$.
\item Let $k\in\N$ and $\varphi_{1},\ldots,\varphi_{k}$ be formulas such that $\alpha{\upharpoonright}_{!}=\varphi_{1}\ldots\varphi_{k}$.
Remark that $\deg(\alpha,\bot)=\deg(\varphi_{1})+\ldots+\deg(\varphi_{k})$ and $\deg(\alpha,B_{a}\varphi)=\deg(\varphi_{1})+\ldots+\deg(\varphi_{k})+\deg(\varphi)+1$.
Hence, $\deg(\alpha,\bot)<\deg(\alpha,B_{a}\varphi)$ and $(\alpha,\bot)\ll(\alpha,B_{a}\varphi)$.
\item Remark that $\deg(\alpha,\bot)\leq\deg(\alpha,\neg\varphi)$.
Moreover, since ${\parallel}{\alpha}{\parallel}+1<{\parallel}{\alpha}{\parallel}+1+{\parallel}{\varphi}{\parallel}$, we obtain that $(\alpha,\bot)\ll(\alpha,\neg\varphi)$.
\end{enumerate}
 \vspace{-.6cm}
\end{proof}

\section*{Appendix: proof details}

The second part of the Appendix contains longer proofs. We repeat the exact formulation of the lemmas and propositions in question.

\bigskip

\noindent {\bf Lemma \ref{lemma:equivalent:formulas::psi:eta:words:become:equivalent:too}}
Let $\phi,\psi$ be formulas such that $\models^*\phi\leftrightarrow\psi$.
Let $M = (W,R,V)$ be a model.
Let $\alpha$ be a history.
Let $\beta$ be a word such that $\alpha\phi\beta$ and $\alpha\psi\beta$ are histories and let $\chi$ be a formula.
For all states $s \in W$,  $s\bowtie \alpha\phi\beta$ iff $s\bowtie \alpha\psi\beta$, and $s,\alpha\phi\beta\models\chi$ iff $s,\alpha\psi\beta\models\chi$.

\begin{proof}
By $\ll$-induction on $(\beta,\chi)$.
\\
Case ``$( \beta,\chi)=( \epsilon,\bot)$''.
Left to the reader.
\\
Case ``$( \beta,\chi)=(\beta'a,\bot)$''.
We have: $s\bowtie \alpha\phi\beta'a$ iff $s\bowtie \alpha\phi\beta'$ and $|\alpha\phi\beta'|_a<|\alpha\phi\beta'|_!$ iff, by induction hypothesis and using the fact that $(\beta^{\prime},\bot)\ll(\beta^{\prime}a,\bot)$, $s\bowtie \alpha\psi\beta'$ and $|\alpha\psi\beta'|_a<|\alpha\psi\beta'|_!$, iff $s\bowtie \alpha\psi\beta'a$.
Moreover, neither $s,\alpha\phi\beta'a\models\bot$, nor $s,\alpha\psi\beta'a\models\bot$.
\\
Case ``$(\beta,\chi)=(\beta'\chi^{\prime},\bot)$''.
We have: $s\bowtie \alpha\phi\beta'\chi^{\prime}$ iff $s\bowtie \alpha\phi\beta'$ and $s \alpha\phi\beta'\models \chi^{\prime}$ iff, by induction hypothesis and using the fact that $(\beta^{\prime},\bot)\ll(\beta^{\prime}\chi^{\prime},\bot)$, $s\bowtie \alpha\psi\beta'$ and $s \alpha\psi\beta'\models \chi^{\prime}$, iff $s\bowtie \alpha\psi\beta'\chi^{\prime}$.
Again, neither $s, \alpha\phi\beta'\chi\models\bot$, nor  $s, \alpha\psi\beta'\chi\models\bot$.
\\
Case ``$(\beta,\chi)=(\beta,p)$''.
We have: $s\bowtie \alpha\phi\beta$ iff, by induction hypothesis and using the fact that $(\beta,\bot)\ll(\beta,p)$, $s\bowtie \alpha\psi\beta$.
Moreover, we have: $s,\alpha\phi\beta\models p$ iff $s\in V(p)$ iff $s,\alpha\psi\beta\models p$.
\\
Case ``$(\beta,\chi)=(\beta,\neg\chi^{\prime})$''.
We have: $s\bowtie \alpha\phi\beta$ iff, by induction hypothesis and the fact that $(\beta,\chi^{\prime})\ll(\beta,\neg\chi^{\prime})$, $s\bowtie \alpha\psi\beta$.
Morever, $s,\alpha\phi\beta\models\neg\chi^{\prime}$ iff $s,\alpha\phi\beta\not\models\chi^{\prime}$ iff, by induction hypothesis and the fact that $(\beta,\chi^{\prime})\ll(\beta,\neg\chi^{\prime})$, $s,\alpha\psi\beta\not\models\chi^{\prime}$ iff $s,\alpha\psi\beta\models\neg\chi^{\prime}$.
\\
Case ``$(\beta,\chi)=(\beta,\chi_1\vee \chi_2)$''.
We have: $s\bowtie \alpha\phi\beta$ iff, by induction hypothesis and the facts that $(\beta,\chi_{1})\ll(\beta,\chi_{1}\vee\chi_{2})$ and $(\beta,\chi_{2})\ll(\beta,\chi_{1}\vee\chi_{2})$, $s\bowtie \alpha\psi\beta$.
Moreover, $s,\alpha\phi\beta\models\chi_1\vee\chi_2$ iff ($s,\alpha\phi\beta\models\chi_1$ or $s,\alpha\phi\beta\models\chi_2$) iff, by induction hypothesis and the facts that $(\beta,\chi_{1})\ll(\beta,\chi_{1}\vee\chi_{2})$ and $(\beta,\chi_{2})\ll(\beta,\chi_{1}\vee\chi_{2})$, $s,\alpha\psi\beta\models\chi_1$ or $s,\alpha\psi\beta\models\chi_2$ iff $s,\alpha\psi\beta\models\chi_1\vee\chi_2$.
\\
Case ``$(\beta,\chi)=(\beta,[a]\chi^{\prime})$''.
We have: $s\bowtie \alpha\phi\beta$ iff, by induction hypothesis and the fact that $(\beta,\chi^{\prime})\ll(\beta,\lbrack a\rbrack\chi^{\prime})$, $s\bowtie \alpha\psi\beta$.
Morever, $s,\alpha\phi\beta\models\lbrack a\rbrack\chi^{\prime}$ iff ${\mid}{\alpha\phi\beta}{\mid}_{a}<{\mid}{\alpha\phi\beta}{\mid}_{!}$ implies $s,\alpha\phi\beta\models\chi^{\prime}$ iff, by induction hypothesis and the fact that $(\beta,\chi^{\prime})\ll(\beta,\lbrack a\rbrack\chi^{\prime})$, ${\mid}{\alpha\psi\beta}{\mid}_{a}<{\mid}{\alpha\psi\beta}{\mid}_{!}$ implies $s,\alpha\psi\beta\models\chi^{\prime}$ iff $s,\alpha\psi\beta\models\lbrack a\rbrack\chi^{\prime}$.
\\
Case ``$(\beta,\chi)=(\beta,[\eta]\chi)$''.
We have: $s\bowtie \alpha\phi\beta$ iff, by induction hypothesis and the fact that $(\beta,\chi^{\prime})\ll(\beta,\lbrack\eta\rbrack\chi^{\prime})$, $s\bowtie \alpha\psi\beta$.
Morever, $s,\alpha\phi\beta\models\lbrack\eta\rbrack\chi^{\prime}$ iff $s,\alpha\phi\beta\models\eta$ implies $s,\alpha\phi\beta\eta\models\chi^{\prime}$ iff, by induction hypothesis and the fact that $(\beta,\eta)\ll(\beta,\lbrack\eta\rbrack\chi^{\prime})$ and $(\beta\eta,\chi^{\prime})\ll(\beta,\lbrack\eta\rbrack\chi^{\prime})$, $s,\alpha\psi\beta\models\eta$ implies $s,\alpha\psi\beta\eta\models\chi^{\prime}$ iff $s,\alpha\psi\beta\models\lbrack\eta\rbrack\chi^{\prime}$.
\\
Case ``$(\beta,\chi)=(\beta,B_a\chi^{\prime})$''.
We have: $s\bowtie \alpha\phi\beta$ iff, by induction hypothesis and the fact that $(\beta,\chi^{\prime})\ll(\beta,B_{a}\chi^{\prime})$, $s\bowtie \alpha\psi\beta$.
Moreover, suppose $s,\alpha\phi\beta\not\models B_{a}\chi^{\prime}$.
Let $t$ be a state and $\delta$ be a history such that $sR_{a}t$, $\alpha\phi\beta\tri_{a}\delta$, $t\bowtie\delta$ and $t,\delta\not\models\chi^{\prime}$.
Let $\delta\lbrack\varphi/\psi\rbrack$ be the history obtained from $\delta$ by eventually replacing an occurrence of $\varphi$ by $\psi$.
Since $\alpha\phi\beta\tri_{a}\delta$,  $\alpha\psi\beta\tri_{a}\delta\lbrack\varphi/\psi\rbrack$.
Now, since $t\bowtie\delta$ and $t,\delta\not\models\chi^{\prime}$, by induction hypothesis, $t\bowtie\delta\lbrack\varphi/\psi\rbrack$ and $t,\delta\lbrack\varphi/\psi\rbrack\not\models\chi^{\prime}$.
Consequently, $s,\alpha\psi\beta\not\models B_{a}\chi^{\prime}$.
%
%
%
%
%
%
\end{proof}
%
%


\noindent {\bf Lemma~\ref{lemma:soundness:of:the:function:tr}}
Let $(W,R,V)$ be a model and $s\in W$.
For all words $\alpha$ over $A\cup\Laa$ and for all formulas $\varphi$, $s,\epsilon\models tr(\alpha,\varphi)$ iff $s,\epsilon\models\lbrack\alpha\rbrack\varphi$.
\begin{proof}
The proof is done by $\ll$-induction on $(\alpha,\varphi)$. In all cases we only show the proof direction from left to right, as the other direction can be shown in a similar way.
\\
{\bf Case $\alpha=\epsilon$ and $\varphi=\bot$.}
Obviously, $s,\epsilon\not\models tr(\epsilon,\bot)$ and $s,\epsilon\not\models\lbrack\epsilon\rbrack\bot$.
\\
{\bf Case $\alpha=\beta a$ and $\varphi=\bot$.}
Suppose $s,\epsilon\models tr(\beta a,\bot)$.
Hence, $s,\epsilon\models tr(\beta,\bot)$ and ${\mid}\beta{\mid}_{a}<{\mid}\beta{\mid}_{!}$, or ${\mid}\beta{\mid}_{a}\geq{\mid}\beta{\mid}_{!}$.
In the former case, by induction hypothesis, $s,\epsilon\models\lbrack\beta\rbrack\bot$.
Thus, by Lemma~\ref{lemma:let:be:a:model:and}, $s\not\bowtie\beta$.
Consequently, $s\not\bowtie\beta a$.
Hence, by Lemma~\ref{lemma:let:be:a:model:and} again, $s,\epsilon\models\lbrack\beta a\rbrack\bot$.
In the latter case, $s\not\bowtie\beta a$.
Thus, by Lemma~\ref{lemma:let:be:a:model:and}, $s,\epsilon\models\lbrack\beta a\rbrack\bot$.
%
%
\\
{\bf Case $\alpha=\beta\psi$ and $\varphi=\bot$.}
Suppose $s,\epsilon\models tr(\beta\psi,\bot)$.
Hence, $s,\epsilon\not\models tr(\beta,\psi)$ or $s,\epsilon\models tr(\beta,\bot)$.
In the former case, by induction hypothesis, $s,\epsilon\not\models\lbrack\beta\rbrack\psi$.
Thus, by Lemma~\ref{lemma:let:be:a:model:and}, $s\bowtie\beta$ and $s,\beta\not\models\psi$.
Consequently, $s\not\bowtie\beta\psi$.
Hence, by Lemma~\ref{lemma:let:be:a:model:and} again, $s,\epsilon\models\lbrack\beta\psi\rbrack\bot$.
In the latter case, by induction hypothesis, $s,\epsilon\models\lbrack\beta\rbrack\bot$.
Thus, by Lemma~\ref{lemma:let:be:a:model:and}, $s\not\bowtie\beta$.
Consequently, $s\not\bowtie\beta\psi$.
Hence, by Lemma~\ref{lemma:let:be:a:model:and} again, $s,\epsilon\models\lbrack\beta\psi\rbrack\bot$.
%
%
\\
{\bf Case $\varphi=p$.}
Suppose $s,\epsilon\models tr(\alpha,p)$.
Hence, $s,\epsilon\models tr(\alpha,\bot)$, or $s,\epsilon\not\models tr(\alpha,\bot)$ and $s,\epsilon\models p$.
In the former case, by induction hypothesis, $s,\epsilon\models\lbrack\alpha\rbrack\bot$.
Thus, by Lemma~\ref{lemma:let:be:a:model:and}, $s\not\bowtie\alpha$.
Consequently, by Lemma~\ref{lemma:let:be:a:model:and} again, $s,\epsilon\models\lbrack\alpha\rbrack p$.
In the latter case, by induction hypothesis, $s,\epsilon\not\models\lbrack\alpha\rbrack\bot$.
Moreover, $s\in V(p)$.
Hence, by Lemma~\ref{lemma:let:be:a:model:and}, $s\bowtie\alpha$.
Moreover, $s,\alpha\models p$.
Thus, by Lemma~\ref{lemma:let:be:a:model:and} again, $s,\epsilon\models\lbrack\alpha\rbrack p$.
%
%
\\
{\bf Case $\varphi=\neg\psi$.}
Suppose $s,\epsilon\models tr(\alpha,\neg\psi)$.
Hence, $s,\epsilon\not\models tr(\alpha,\psi)$ or $s,\epsilon\models tr(\alpha,\bot)$
In the former case, by induction hypothesis, $s,\epsilon\not\models\lbrack\alpha\rbrack\psi$.
Thus, by Lemma~\ref{lemma:let:be:a:model:and}, $s\bowtie\alpha$ and $s,\alpha\not\models\psi$.
Consequently, by Lemma~\ref{lemma:let:be:a:model:and} again, $s,\epsilon\models\lbrack\alpha\rbrack\neg\psi$.
In the latter case, by induction hypothesis, $s,\epsilon\models\lbrack\alpha\rbrack\bot$.
Hence, by Lemma~\ref{lemma:let:be:a:model:and}, $s\not\bowtie\alpha$.
Thus, by Lemma~\ref{lemma:let:be:a:model:and} again, $s,\epsilon\models\lbrack\alpha\rbrack\neg\psi$.
%
%
\\
{\bf Case $\varphi=\psi\vee\chi$.}
Suppose $s,\epsilon\models tr(\alpha,\psi\vee\chi)$.
Hence, $s,\epsilon\models tr(\alpha,\psi)$ or $s,\epsilon\models tr(\alpha,\chi)$.
In the former case, by induction hypothesis, $s,\epsilon\models\lbrack\alpha\rbrack\psi$.
Thus, by Lemma~\ref{lemma:let:be:a:model:and}, $s\not\bowtie\alpha$ or $s,\alpha\models\psi$.
Consequently, $s\not\bowtie\alpha$ or $s,\alpha\models\psi\vee\chi$.
Hence, by Lemma~\ref{lemma:let:be:a:model:and} again, $s,\epsilon\models\lbrack\alpha\rbrack(\psi\vee\chi)$.
The latter case is similarly treated.
%
%
\\
{\bf Case $\varphi=B_{a}\psi$.}
Suppose $s,\epsilon\models tr(\alpha,B_{a}\psi)$.
Hence, $s,\epsilon\models tr(\alpha,\bot)$ or $s,\epsilon\models\bigwedge\{B_{a} tr(\beta,\psi) \mid \alpha\triangleright_{a}\beta\}$.
In the former case, by induction hypothesis, $s,\epsilon\models\lbrack\alpha\rbrack\bot$.
Thus, by Lemma~\ref{lemma:let:be:a:model:and}, $s\not\bowtie\alpha$.
Consequently, by Lemma~\ref{lemma:let:be:a:model:and} again, $s,\epsilon\models\lbrack\alpha\rbrack B_{a}\psi$.
In the latter case, for the sake of the contradiction, suppose $s,\epsilon\not\models\lbrack\alpha\rbrack B_{a}\psi$.
Thus, by Lemma~\ref{lemma:let:be:a:model:and}, $s\bowtie\alpha$ and $s,\alpha\not\models B_{a}\psi$.
Let $t\in W$ and $\gamma$ be a history such that $sR_{a}t$, $\alpha\triangleright_{a}\gamma$, $t\bowtie\gamma$ and $t,\gamma\not\models\psi$.
Since $s,\epsilon\models\bigwedge\{B_{a} tr(\beta,\psi) \mid \alpha\triangleright_{a}\beta\}$, we obtain that $s,\epsilon\models B_{a} tr(\gamma,\psi)$.
We recall the reader that $sR_{a}t$.
Moreover, obviously, $\epsilon\triangleright_{a}\epsilon$ and $t\bowtie\epsilon$.
Consequently, $t,\epsilon\models tr(\gamma,\psi)$.
Hence, by induction hypothesis, $t,\epsilon\models\lbrack\gamma\rbrack\psi$.
Thus, by Lemma~\ref{lemma:let:be:a:model:and}, $t\not\bowtie\gamma$ or $t,\gamma\models\psi$: a contradiction.
Consequently, $s,\epsilon\models\lbrack\alpha\rbrack B_{a}\psi$.
%
%
\\
{\bf Case $\varphi=\lbrack a\rbrack\psi$.}
Suppose $s,\epsilon\models tr(\alpha,\lbrack a\rbrack\psi)$.
Hence, $s,\epsilon\models tr(\alpha a,\psi)$.
Thus, by induction hypothesis, $s,\epsilon\models\lbrack\alpha a\rbrack\psi$.
Consequently, $s,\epsilon\models\lbrack\alpha\rbrack\lbrack a\rbrack\psi$.
%
%
\\
{\bf Case $\varphi=\lbrack\psi\rbrack\chi$.}
Suppose $s,\epsilon\models tr(\alpha,\lbrack\psi\rbrack\chi)$.
Hence, $s,\epsilon\models tr(\alpha\psi,\chi)$.
Thus, by induction hypothesis, $s,\epsilon\models\lbrack\alpha\psi\rbrack\chi$.
Consequently, $s,\epsilon\models\lbrack\alpha\rbrack\lbrack\psi\rbrack\chi$.
%
%
\end{proof}


\noindent {\bf Proposition~\ref{prop.annk}}
Let $\phi,\psi \in \Laa$. Then \
\begin{enumerate}
\item $\wordmodels [\phi] \bot \eq \neg\phi$
\item $\wordmodels [\phi] p \eq (\phi \imp p)$
\item $\wordmodels [\phi] \neg\psi \eq (\phi \imp \neg [\phi] \psi)$
\item $\wordmodels [\phi] (\psi\vee\chi) \eq ([\phi] \psi \vee [\phi] \chi)$
\item $\wordmodels [\phi]B_a\psi \eq (\phi \imp B_a\psi)$
\end{enumerate}
\begin{proof}
Let $(W,R,V)$, $s\in W$, and $\alpha$ with $s \bowtie \alpha$ be given.
We will implicitly use Corollary~\ref{corollary:models:histories:bowtie:aa}.
\begin{enumerate}
\item $s,\alpha \models [\phi] \bot \\ \Eq \\ s, \alpha \models \phi$ implies $s,\alpha\phi \models \bot \\ \Eq \\ s, \alpha\not \models \phi \\ \Eq \\ s,\alpha \models \neg \phi$.
\item $s,\alpha \models [\phi] p \\ \Eq \\ s, \alpha \models \phi$ implies $s,\alpha\phi \models p \\ \Eq (\sharp) \\ s, \alpha \models \phi$ implies $s,\alpha \models p \\ \Eq \\ s, \alpha \models \phi \imp p$.

The equivalence $(\sharp)$ holds because the value of propositional variables is invariant for history shortening and history extension.
\item $s,\alpha \models [\phi] \neg\psi \\ \Eq \\ s,\alpha \models \phi$ implies $s,\alpha\phi\not\models\psi \\ \Eq \\ s,\alpha \models \phi \imp \neg [\phi] \psi$.
\item The proof is elementary.
\item $
s,\alpha \models [\phi]B_a \psi \\
\Eq \\
s,\alpha \models \phi \text{ implies } s,\alpha\phi \models B_a \psi \\
\Eq (*) \\
s,\alpha \models \phi \text{ implies } s,\alpha \models B_a \psi \\
\Eq \\
s,\alpha \models \phi \imp B_a \psi. \\
$

\noindent $(*)$: This holds because $s,\alpha\models\phi$ implies $s \bowtie \alpha\phi$, and because $\alpha\phi\triangleright_a\beta$ iff $\alpha\triangleright_a\beta$, which is true because $|\alpha\phi|_a = |\alpha|_a$: the interpretation of knowledge is not affected by adding announcement $\phi$ to the history $\alpha$.
\end{enumerate} \vspace{-.6cm}
\end{proof}

\noindent {\bf Proposition~\ref{proprecep}}
$\wordmodels [a] [b] \phi \eq [b] [a] \phi$

\begin{proof}
We first show the following, by induction on $\phi$:
\begin{quote}
Let $\phi\in\Laa$, $\alpha,\gamma \in (\Laa \union A)^*$, $s\bowtie\alpha$, and $\beta,\beta'\in A^*$ such that for all $a \in A$, $|\beta|_a = |\beta'|_a$. If $s\bowtie\alpha\beta\gamma$ and $s\bowtie\alpha\beta^{\prime}\gamma$, then $s,\alpha\beta\gamma \models \phi$ iff $s,\alpha\beta'\gamma \models \phi$.
\end{quote}
Note that $s\bowtie\alpha\beta$ iff $s\bowtie\alpha\beta'$, as for each agent $a$ the number of occurrences in $\beta$ and $\beta'$ is the same (their order with respect to other agents does not matter).
The non-trivial cases of the induction are knowledge, announcement, and reception:

{\bf Case $B_a \phi$:} Suppose $s,\alpha\beta\gamma \models B_a \phi$.
Thus, $t,\delta \models \phi$ for all $t, \delta$ such that $sR_at$, $\alpha\beta\gamma \triangleright_a \delta$, and $t \bowtie \delta$. Observe that $\alpha\beta\gamma \triangleright_a \delta$ iff $\alpha\beta'\gamma \triangleright_a \delta$, as from $|\beta|_a = |\beta'|_a$ it follows that $|\alpha\beta\gamma|_a = |\alpha\beta'\gamma|_a$, and we also have $\alpha\beta\gamma\proj_{!a} = \alpha\beta'\gamma\proj_{!a}$, as $\beta$ and $\beta'$ do not contain formulas. Therefore, $t,\delta \models \phi$ for all $t, \delta$ such that $sR_at$, $\alpha\beta'\gamma \triangleright_a \delta$, and $t \bowtie \delta$, i.e., $s,\alpha\beta'\gamma \models B_a \phi$. (No inductive hypothesis is needed.)

{\bf Case $[a] \phi$:} Suppose $s,\alpha\beta\gamma \models [a] \phi$.
Thus, $|\alpha\beta\gamma|_a < |\alpha\beta\gamma|_!$ implies $s,\alpha\beta\gamma{a} \models \phi$. We now use the inductive hypothesis for $\phi$, applied to $\gamma' = \gamma{a}$, to obtain that ${\mid}\alpha\beta^{\prime}\gamma{\mid}_{a}<{\mid}\alpha\beta\gamma{\mid}_{!}$ implies $s,\alpha\beta'\gamma{a} \models \phi$. Thus, $s,\alpha\beta'\gamma \models [a] \phi$.

{\bf Case $[\psi] \phi$:} Suppose $s,\alpha\beta\gamma \models [\psi] \phi$.
Thus, ($s,\alpha\beta\gamma \models \psi$ implies $s,\alpha\beta\gamma\psi \models \phi$). The antecedent of the implication is equivalent to $s,\alpha\beta'\gamma \models \psi$, by induction. The consequent of the implication is equivalent to $s,\alpha\beta'\gamma\psi \models \phi$, also by induction.
Therefore, $s,\alpha\beta'\gamma \models \psi$ implies $s,\alpha\beta'\gamma\psi \models \phi$.
Hence, $s,\alpha\beta'\gamma \models [\psi] \phi$.

This completes the inductive proof.
A corollary is:
\begin{quote} Let $\beta,\beta'\in A^*$ such that for all $a \in A$, $|\beta|_a = |\beta'|_a$. Then $\wordmodels [\beta]\phi \eq [\beta']\phi$. \end{quote} The proposition to be proved is the special case for $\beta = ab$ and $\beta' = ba$.
\end{proof}
%
%

\noindent {\bf Proposition~\ref{prop.nextcomm}}
Let $\phi \in \Laa$ and such that $B_a$ and $[a]$ do not occur in $\phi$. Then $\wordmodels \neg [a]\bot \imp ([a]\phi\eq\phi)$.
\begin{proof}
This follows directly from the following slightly stronger proposition (namely from the case for $\beta =\epsilon$). By ``$a$ does not occur in $\phi$'' we mean ``$\phi$ is a formula in the language $\Laa$ with respect to the set of agents $A\setminus\{a\}$,'' i.e., $B_a$ and $[a]$ do not occur in $\phi$. By ``$a$ does not occur in history $\alpha$'' we mean ``$a$ does not occur as an agent (as a letter) in $\alpha$ and $a$ does not occur in any formula $\psi$ that occurs (as a letter) in $\alpha$.''
\begin{quote}
Given are $\phi\in\Laa$, model $(W,R,V)$ with $s \in W$, and histories $\alpha$ and $\beta$, such that $a$ does not occur in $\beta$ or $\phi$, and $|\alpha|_! > |\alpha|_a$. Then $s,\alpha{a}\beta \models \phi$ iff $s,\alpha\beta\models\phi$.
\end{quote}
The proof is by induction on the structure of $\phi$. The cases $\bot$, $p$, $\phi\et\psi$, and $\neg\phi$ are elementary. We note that the induction in the cases $[\phi]\psi$ and $[b]\psi$ below is justified, as it applies to $\psi$ for {\bf any} $\alpha$ and $\beta$ (where $a$ does not occur in $\beta$), therefore including the $\beta\phi$ and $\beta{b}$, also in the proof below.

{\bf Case $[\phi]\psi$}:

\noindent $
s,\alpha{a}\beta \models [\phi]\psi \\ \Eq \\
s,\alpha{a}\beta \models \phi \text{ implies } s,\alpha{a}\beta\phi \models \psi \\ \Eq (\text{IH})\\
s,\alpha\beta \models \phi \text{ implies } s,\alpha\beta\phi \models \psi \\ \Eq \\
s,\alpha\beta \models [\phi]\psi
$

{\bf Case $[b]\psi$ ($b\neq a$)}:

\noindent $s,\alpha{a}\beta \models [b]\psi \\ \Eq \\ |\alpha{a}\beta|_b < |\alpha{a}\beta|_!$ implies $s,\alpha{a}\beta{b} \models \psi \\ \Eq (\text{IH}) \\ |\alpha{a}\beta|_b < |\alpha{a}\beta|_!$ implies $s,\alpha\beta{b} \models \psi \\ \Eq \\ |\alpha\beta|_b < |\alpha\beta|_!$ implies $s,\alpha\beta{b} \models \psi \\ \Eq \\ s,\alpha\beta \models [b]\psi$

{\bf Case $B_b\psi$}:

\noindent By definition: $
s,\alpha{a}\beta \models B_b\psi$ iff $t,\delta \models \psi \text{ for any } t,\delta \text{ such that } sR_at, \alpha{a}\beta\tri_b \delta, \text{ and } t \bowtie \delta$. It then suffices to prove that $\alpha{a}\beta\tri_b \delta$ iff $\alpha\beta\tri_b \delta$. But this follows directly from the fact that $|\alpha a \beta|_b=|\alpha\beta|_b$ and $|\alpha a \beta|_!=|\alpha\beta|_!$, and thus $\alpha a\beta\proj_{!b}=\alpha\beta\proj_{!b}$, which entails that $$\alpha a \beta\tri_b \delta \text{ iff } \delta\proj_{!b}=\delta\proj_!=\alpha a\beta\proj_{!b} \text{ iff } \delta\proj_{!b}=\delta\proj_!=\alpha \beta\proj_{!b} \text{ iff } \alpha  \beta\tri_b \delta. \vspace{-.9cm} $$
\end{proof}

\noindent {\bf Proposition~\ref{prop.historiesforever}}
Let $\phi\in\Laa$ be given. For all $M$, and $s$ and $\alpha$ with $s \bowtie \alpha$, $s, \alpha \models \phi$ iff $M^{\alpha\proj_!}, s,\alpha \models \phi$.

\begin{proof}
Instead of the above, we prove the slightly stronger proposition:
\begin{quote} Let $\phi\in\Laa$ be given. For all $M$, and $s$ and $\alpha,\beta$ with $s \bowtie \alpha$ and $\alpha\proj_! \subseteq \beta\proj_!$, $s, \alpha \models \phi$ iff $M^{\beta\proj_!}, s,\alpha \models \phi$. \end{quote} From this, the required follows for $\alpha = \beta$.\footnote{As a matter of interest the stronger proposition also proves that truth in asynchronous models is preserved under history extension: let $\gamma\proj_! \subseteq \alpha\proj_! \subseteq \beta\proj_!$, then $M^{\alpha\proj_!},s,\gamma \models \phi$ implies $M^{\beta\proj_!},s,\gamma \models \phi$.} The proof is by induction on $\phi$. All cases are elementary except those involving modalities. Let $M$, and $s$ and $\alpha$ with $s \bowtie \alpha$ be given.

{\bf Case $B_a \phi$:} \ \

\noindent $
s,\alpha \models B_a \phi \\
\Eq \\
t,\gamma \models \phi \ \text{for all}\ (t,\gamma) \ \text{such that}\ sR_at, \alpha \tri_a \gamma, \text{and}\ t \bowtie \gamma \\
\Eq \hfill \text{induction, use that}\ \gamma\proj_! = \alpha\proj_{!a} \subseteq \alpha\proj_! \subseteq \beta\proj_! \\
M^{\beta\proj_!}, t,\gamma \models \phi \ \text{for all}\ (t,\gamma) \ \text{such that}\ sR_at, \alpha \tri_a \gamma, \text{and}\ t \bowtie \gamma \\
\Eq \hfill \text{by definition,}\ (s, \alpha) R'_a (t,\gamma) \ \text{iff}\ sR_at \ \text{and} \  \alpha \tri_a \gamma, \text{and} \ t \bowtie \gamma \ \text{as}\ (t,\gamma) \in W' \\
M^{\beta\proj_!}, t,\gamma \models \phi \text{ for all } (t,\gamma) \ \text{such that}\ (s, \alpha) R'_a (t,\gamma) \ \text{and}\ (s, \alpha) R'_{\tri_a} (t,\gamma) \\
\Eq \\
M^{\beta\proj_!}, s,\alpha \models B_a \phi
$

{\bf Case $[\phi]\psi$:} \ \ In this case we need to distinguish subcase $\alpha\phi\proj_!\subseteq\beta\proj_!$ from subcase $\alpha\phi\proj_!\not\subseteq\beta\proj_!$. Below we show the second subcase, for the first subcase, replace below the occurrences of $M^{\alpha\phi\proj_!}$ by $M^{\beta\proj_!}$ and remove the last equivalence.

\noindent $
s,\alpha \models [\phi]\psi \\
\Eq \\
s,\alpha \models \phi \ \text{implies}\ s,\alpha\phi \models \psi \\
\Eq \hfill \text{induction} \\
M^{\alpha\phi\proj_!}, s,\alpha \models \phi \ \text{implies}\ M^{\alpha\phi\proj_!}, s,\alpha\phi \models \psi \\
\Eq \hfill \text{either $M^{\alpha\phi\proj_!}, s,\alpha \not\models \phi$, or the unique $R'_{\phi}$-successor is $(t,\gamma) = (s,\alpha\phi)$} \\
M^{\alpha\phi\proj_!}, t,\gamma \models \psi \text{ for all } (t,\gamma) \text{ such that } (s,\alpha) R'_\phi (t,\gamma) \\
\Eq \\
M^{\alpha\phi\proj_!}, s,\alpha \models [\phi]\psi \\
\Eq \hfill \text{case `else' of the semantics of $[\phi]\psi$ applies, as}\ \alpha\phi\proj_! \not\subseteq \beta\proj_!  \\
M^{\beta\proj_!}, s,\alpha \models [\phi]\psi
$

{\bf Case $[a]\phi$:} \ \

\noindent $
s,\alpha \models [a]\phi \\
\Eq \\ |\alpha|_a < |\alpha|_! \ \text{implies}\ s,\alpha{a}\models \phi \\
\Eq \hfill \text{induction, also use that}\ \alpha\proj_! = \alpha{a}\proj_! \\
|\alpha|_a < |\alpha|_! \ \text{implies}\ M^{\beta\proj_!}, s,\alpha{a}\models \phi \\
\Eq \hfill \text{either $|\alpha|_a = |\alpha|_!$, or the unique $R'_{[a]}$-successor is $(t,\gamma) = (s,\alpha{a})$} \\
M^{\beta\proj_!}, t,\gamma \models \phi \text{ for all } (t,\gamma) \ \text{such that}\ (s,\alpha) R'_{[a]} (t,\gamma) \\
\Eq \\
M^{\beta\proj_!}, s,\alpha \models [a]\phi
$
\end{proof}

\end{document}